%% file: main.tex
\pdfoutput=1 %
\documentclass[10pt,twocolumn,letterpaper]{article}

\PassOptionsToPackage{table}{xcolor} %
\usepackage[pagenumbers]{cvpr} %

\input{preamble}

\definecolor{cvprblue}{rgb}{0.21,0.49,0.74}
\usepackage[pagebackref,breaklinks,colorlinks,allcolors=cvprblue]{hyperref}
\input{supp/supp_preamble}
\hypersetup{pdftitle={FailPoint-X: A Policy-Matched, Failure-Anchored Benchmark for Offline VLA Post-Training from Human Takeovers},
  pdfauthor={Zirui Li, Yujie Yan, Jingxuan Qian, Yangang Ren, Haizhou Yin, Haoran Xu, Chen Lv}}

\newif\ifshowappendix
\showappendixtrue

\def\confName{CVPR}
\def\confYear{2027}

\title{\fpxfulltitle}

\author{
Zirui Li$^{1,*}$ \quad Yujie Yan$^{1,*}$ \quad Jingxuan Qian$^{1}$ \quad Yangang Ren$^{1}$ \\
Haizhou Yin$^{1}$ \quad Haoran Xu$^{1}$ \quad Chen Lv$^{1,\dagger}$ \\[2pt]
$^{1}$Nanyang Technological University, Singapore \\
{\small $^{*}$Equal contribution \quad $^{\dagger}$Corresponding author}
}

\begin{document}

\twocolumn[{%
\renewcommand\twocolumn[1][]{#1}%
\maketitle
\vspace{-14pt}
\centerline{\small\urlstyle{same}\raisebox{-0.2\height}{\includegraphics[height=1.25em]{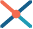}}\hspace{0.7em}%
\faGithub\hspace{0.3em}\href{\fpxcodeurl}{\nolinkurl{github.com/yujieyan81/failpoint-x}}\hspace{1.6em}%
\faGlobe\hspace{0.3em}\href{\fpxwebsiteurl}{\nolinkurl{lzrbit.github.io/FailPoint_X}}}
\vspace{6pt}
\input{figures/fig1_teaser}
}]

\input{sec/0_abstract}

\input{sec/1_intro}
\input{sec/2_related}
\input{sec/4_dataset}
\input{sec/3_protocol}
\input{sec/5_experiments}
\input{sec/6_conclusion}

\input{sec/X_llm_statement}

{
    \small
    \bibliographystyle{ieeenat_fullname}
    \bibliography{refs}
}

\ifshowappendix\input{supp/appendix}\fi

\end{document}

%% file: preamble.tex
\usepackage{multirow}
\usepackage{makecell}
\usepackage{tabularx}
\usepackage{pifont}   %
\usepackage{bm}
\usepackage{soul}   %

\makeatletter
\IfFormatAtLeastTF{2025-06-01}{%
  \@ifpackagelater{lineno}{2025/01/01}{%
    \RemoveFromHook{build/column/before}[lineno]%
    \AddToHook{build/column/before}[fpx/lineno-fix]{%
      \protected@write\@auxout{}{\string\def\string\@LN@column{\if@firstcolumn2\else1\fi}}%
    }%
  }{}%
}{}
\makeatother

\input{macros}
\input{numbers}
\input{results} %
\input{supp/tab_s_configs} %
\input{sec/section5_numbers} %
\input{supp/revision_numbers} %
\input{figures/figure_value_guard} %
\input{soulreg} %
\usepackage{fontawesome5} %

%% file: macros.tex
\newif\iffinal
\finaltrue %
\definecolor{fpxtodo}{HTML}{FFF176}
\DeclareRobustCommand{\todohl}[1]{\iffinal\PackageError{fpx}{Unresolved to-do: #1}{Resolve it or delete it.}\else{\sethlcolor{fpxtodo}\hl{#1}}\fi}
\newcommand{\pending}[1]{\iffinal\PackageError{fpx}{Unresolved pending value: #1}{Insert the frozen number or delete the sentence.}\else{\sethlcolor{fpxtodo}\hl{[PENDING:~#1]}}\fi}
\newcommand{\authorinput}[1]{\iffinal\PackageError{fpx}{Unresolved author input: #1}{Ask the authors.}\else{\sethlcolor{fpxtodo}\hl{[AUTHOR INPUT:~#1]}}\fi}

\definecolor{FPXverA}{HTML}{F26B4F}\definecolor{FPXverB}{HTML}{C0392B}
\definecolor{FPXpetA}{HTML}{1F9BB5}\definecolor{FPXpetB}{HTML}{155E8A}
\newcommand{\fpxlogo}{{\textcolor[HTML]{F26B4F}{F}\textcolor[HTML]{E15A43}{a}\textcolor[HTML]{D14A37}{i}\textcolor[HTML]{C0392B}{l}\textcolor[HTML]{1F9BB5}{P}\textcolor[HTML]{1D91AE}{o}\textcolor[HTML]{1C87A7}{i}\textcolor[HTML]{1A7CA0}{n}\textcolor[HTML]{187298}{t}\textcolor[HTML]{176891}{-}\textcolor[HTML]{155E8A}{X}}}
\newcommand{\fpxfulltitle}{\texorpdfstring{\fpxlogo}{FailPoint-X}: A Policy-Matched, Failure-Anchored Benchmark\\for Offline VLA Post-Training from Human Takeovers}

\newcommand{\fpx}{FailPoint-X\xspace}
\newcommand{\liberox}{LIBERO-X\xspace}
\newcommand{\pizf}{$\pi_{0.5}$\xspace}
\newcommand{\SRrec}{SR$_\text{rec}$\xspace}
\newcommand{\SRnorm}{SR$_\text{norm}$\xspace}
\newcommand{\dSRnorm}{$\Delta$SR$_\text{norm}$\xspace}

\newcommand{\BaseRestart}{\mbox{Base-restart}\xspace}
\newcommand{\DemoCont}{\mbox{Demo-continue}\xspace}
\newcommand{\TakeoverBC}{\mbox{Takeover-BC}\xspace}
\newcommand{\MixBC}{\mbox{Mix-BC}\xspace}
\newcommand{\IQLR}{\mbox{IQL-R2}\xspace}
\newcommand{\WBC}{Weighted-BC\xspace}
\newcommand{\BCstar}{BC$^{\star}$\xspace}
\newcommand{\IQLstar}{IQL$^{\star}$\xspace}

\definecolor{TabHead}{HTML}{E0F1F5}\definecolor{TabGroup}{HTML}{EEF1F4}\definecolor{TabAlt}{HTML}{F7FAFB}
\newcommand{\tabhead}{\rowcolor{TabHead}}
\newcommand{\tabgroup}[2]{\rowcolor{TabGroup}\multicolumn{#1}{@{}l}{\rule{0pt}{2.2ex}\textbf{#2}}\\}
\newcommand{\tabzebra}{\rowcolors{2}{white}{TabAlt}}
\newcommand{\tabstyle}{\setlength{\aboverulesep}{0pt}\setlength{\belowrulesep}{0pt}\setlength{\extrarowheight}{0.35ex}}
\definecolor{cmYes}{HTML}{2E8B57}\definecolor{cmNo}{HTML}{D1495B}\definecolor{cmPart}{HTML}{E59A1C}
\newcommand{\cmark}{\textcolor{cmYes}{\ding{52}}}
\newcommand{\xmark}{\textcolor{cmNo}{\ding{56}}}
\newcommand{\pmark}{\textcolor{cmPart}{\ding{51}}}
\newcommand{\ns}{n/s}

\newcommand{\concurrent}{$^{\dagger}$}

\newcommand{\pendcell}{\iffinal\PackageError{fpx}{Unresolved pending table cell}{Fill from the frozen analysis output.}\else{\colorbox{fpxtodo}{$\bullet$}}\fi}

\newcommand{\fpxcodeurl}{https://github.com/yujieyan81/failpoint-x}
\newcommand{\fpxwebsiteurl}{https://lzrbit.github.io/FailPoint_X/}
\newcommand{\fpxprojectlink}{\href{\fpxwebsiteurl}{{\Urlmuskip=0mu\relax\urlstyle{same}\nolinkurl{lzrbit.github.io/FailPoint_X}}}}
\newcommand{\fpxcodelink}{\href{\fpxcodeurl}{{\Urlmuskip=0mu\relax\urlstyle{same}\nolinkurl{github.com/yujieyan81/failpoint-x}}}}

%% file: numbers.tex
\newcommand{\nFamilies}{10\xspace} %
\newcommand{\nLevels}{five\xspace} %
\newcommand{\nConfigs}{50\xspace} %
\newcommand{\nConfigsComplete}{48\xspace} %
\newcommand{\nFamiliesKitchen}{3\xspace} %
\newcommand{\nFamiliesStudy}{7\xspace} %
\newcommand{\nFailReleased}{629\xspace} %
\newcommand{\nFailTrain}{245\xspace} %
\newcommand{\nFailTest}{384\xspace} %
\newcommand{\nTestConfirm}{240\xspace} %
\newcommand{\nTestDiag}{144\xspace} %
\newcommand{\nShipped}{1,287\xspace} %
\newcommand{\nShippedTrain}{506\xspace} %
\newcommand{\nShippedTest}{781\xspace} %
\newcommand{\nAnnotatedRollouts}{753\xspace} %
\newcommand{\nWithheld}{55\xspace} %
\newcommand{\nUndecided}{603\xspace} %
\newcommand{\nAnnotationRecords}{756\xspace} %
\newcommand{\nAttemptsAll}{2,823\xspace} %
\newcommand{\nAttemptsActive}{2,799\xspace} %
\newcommand{\nAttemptsArchived}{24\xspace} %
\newcommand{\nAccepted}{2,400\xspace} %
\newcommand{\nSuccessAll}{2,471\xspace} %
\newcommand{\nDiscardedSucc}{71\xspace} %
\newcommand{\nStops}{309\xspace} %
\newcommand{\nTimeLimit}{4\xspace} %
\newcommand{\nPhysics}{15\xspace} %
\newcommand{\nUnsuccessful}{328\xspace} %
\newcommand{\pctStopsOfUnsuccessful}{94\xspace} %
\newcommand{\pctDiscardedOfSucc}{2.9\xspace} %
\newcommand{\nSourcesActive}{242\xspace} %
\newcommand{\nSourcesComplete}{240\xspace} %
\newcommand{\nSourcesArchived}{4\xspace} %
\newcommand{\nRemediations}{4\xspace} %
\newcommand{\humanSR}{88.8\xspace} %
\newcommand{\humanSRLOne}{88.5\xspace} %
\newcommand{\humanSRLTwo}{86.3\xspace} %
\newcommand{\humanSRLThree}{89.9\xspace} %
\newcommand{\humanSRLFour}{91.1\xspace} %
\newcommand{\humanSRLFive}{88.4\xspace} %
\newcommand{\humanSRphysFail}{88.3\xspace} %
\newcommand{\humanSRstopVoid}{99.8\xspace} %
\newcommand{\humanSRmin}{86.3\xspace} %
\newcommand{\humanSRmax}{91.1\xspace} %
\newcommand{\nStepsAccepted}{828,827\xspace} %
\newcommand{\hoursAcceptedControl}{11.5\xspace} %
\newcommand{\medianSteps}{317\xspace} %
\newcommand{\medianSec}{15.9\xspace} %
\newcommand{\meanSteps}{345.3\xspace} %
\newcommand{\medSecLOne}{7.0\xspace} %
\newcommand{\pninetySecLOne}{29.7\xspace} %
\newcommand{\nAccLOne}{500\xspace} %
\newcommand{\medSecLTwo}{9.9\xspace} %
\newcommand{\pninetySecLTwo}{28.9\xspace} %
\newcommand{\nAccLTwo}{500\xspace} %
\newcommand{\medSecLThree}{17.2\xspace} %
\newcommand{\pninetySecLThree}{42.8\xspace} %
\newcommand{\nAccLThree}{500\xspace} %
\newcommand{\medSecLFour}{18.1\xspace} %
\newcommand{\pninetySecLFour}{40.3\xspace} %
\newcommand{\nAccLFour}{450\xspace} %
\newcommand{\medSecLFive}{19.2\xspace} %
\newcommand{\pninetySecLFive}{39.0\xspace} %
\newcommand{\nAccLFive}{450\xspace} %
\newcommand{\pctWithinSixHundred}{83.1\xspace} %
\newcommand{\pctWithinTwelveHundred}{99.5\xspace} %
\newcommand{\nOverTwelveHundred}{12\xspace} %
\newcommand{\attPerSourceMean}{11.65\xspace} %
\newcommand{\attPerSourceMedian}{11\xspace} %
\newcommand{\attPerSourceMax}{24\xspace} %
\newcommand{\nExactlyTen}{70\xspace} %
\newcommand{\nTrainSrcLThree}{150\xspace} %
\newcommand{\nTrainAccLThree}{1,500\xspace} %
\newcommand{\nTrainStepsLThree}{441,210\xspace} %
\newcommand{\nTrainSrcPost}{232\xspace} %
\newcommand{\nTrainAccPost}{2,320\xspace} %
\newcommand{\nTrainStepsPost}{797,385\xspace} %
\newcommand{\ftWrongPlacement}{195\xspace} %
\newcommand{\pctftWrongPlacement}{31.0\xspace} %
\newcommand{\ftWrongObject}{190\xspace} %
\newcommand{\pctftWrongObject}{30.2\xspace} %
\newcommand{\ftCollision}{94\xspace} %
\newcommand{\pctftCollision}{14.9\xspace} %
\newcommand{\ftMissedGrasp}{62\xspace} %
\newcommand{\pctftMissedGrasp}{9.9\xspace} %
\newcommand{\ftStuck}{37\xspace} %
\newcommand{\pctftStuck}{5.9\xspace} %
\newcommand{\ftDropped}{37\xspace} %
\newcommand{\pctftDropped}{5.9\xspace} %
\newcommand{\ftDrawerOpen}{14\xspace} %
\newcommand{\pctftDrawerOpen}{2.2\xspace} %
\newcommand{\pctftGrounding}{61.2\xspace} %
\newcommand{\suppWhNoFailureLate}{24\xspace} %
\newcommand{\suppLateTrain}{8\xspace} %
\newcommand{\suppLateTest}{12\xspace} %
\newcommand{\suppLateTestConf}{9\xspace} %
\newcommand{\suppLateRel}{20\xspace} %
\newcommand{\nRolloutsGen}{11{,}190\xspace} %
\newcommand{\nRolloutsGenTrain}{4{,}678\xspace} %
\newcommand{\nRolloutsGenTest}{6{,}512\xspace} %
\newcommand{\nGenInvalidStart}{2{,}559\xspace} %
\newcommand{\nGenPhysFault}{105\xspace} %
\newcommand{\nGenSuccFast}{5{,}492\xspace} %
\newcommand{\nGenSuccSlow}{1{,}017\xspace} %
\newcommand{\nGenFail}{2{,}017\xspace} %
\newcommand{\nGenNearMiss}{166\xspace} %
\newcommand{\nGenStartInvalid}{46\xspace} %
\newcommand{\nCandidates}{1{,}805\xspace} %
\newcommand{\nOddMidpoints}{185\xspace} %
\newcommand{\demoPNNTrans}{0.71--0.83\xspace} %
\newcommand{\demoPNNRot}{0.16--0.24\xspace} %
\newcommand{\demoAboveCapTrans}{19--25\xspace} %
\newcommand{\demoAboveCapRot}{10--13\xspace} %
\newcommand{\nLOneTasks}{600\xspace} %
\newcommand{\nTUnmatched}{20\xspace} %
\newcommand{\TmedianSteps}{342\xspace} %
\newcommand{\nDemoFrames}{889{,}277\xspace} %
\newcommand{\openpiCommit}{215abfb\xspace} %
\newcommand{\liberoxDataRev}{73053111f932\xspace} %
\newcommand{\baseCkptSha}{6a6c6aa1da11176d\allowbreak{}e1c0e16f662fe0f5\allowbreak{}c24eda07dab80cd4\allowbreak{}8ca1e8ac38397c11\xspace} %
\newcommand{\baseSRLOneTwoStep}{76.1\xspace} %
\newcommand{\baseSRFamLOneCap}{71.0\xspace} %
\newcommand{\baseSRFamLOneT}{62.0\xspace} %
\newcommand{\demoDispPerFrame}{0.0105\xspace} %
\newcommand{\simDispPerStep}{0.0116\xspace} %
\newcommand{\nKinRollouts}{229\xspace} %
\newcommand{\demoLenRatioMed}{0.98\xspace} %
\newcommand{\nDemos}{2,520\xspace} %
\newcommand{\baseTrainSteps}{30k\xspace} %
\newcommand{\baseTrainHours}{22.6\xspace} %
\newcommand{\baseSRLOneCap}{77.2\xspace} %
\newcommand{\baseSRLOneT}{62.2\xspace} %
\newcommand{\hoursControl}{14.0\xspace} %
\newcommand{\hoursPreview}{6.9\xspace} %
\newcommand{\hoursAnnotation}{5.4\xspace} %
\newcommand{\controlHz}{20\xspace} %
\newcommand{\budgetSteps}{1,800\xspace} %
\newcommand{\budgetSec}{90\xspace} %
\newcommand{\framesPerFailure}{10\xspace} %
\newcommand{\intervalMin}{10\xspace} %
\newcommand{\intervalMax}{40\xspace} %
\newcommand{\horizonMain}{1,200\xspace} %
\newcommand{\horizonShort}{600\xspace} %
\newcommand{\horizonLong}{1,800\xspace} %
\newcommand{\samplesPerPoint}{3\xspace} %
\newcommand{\nSeeds}{1\xspace} %
\newcommand{\nNominal}{500\xspace} %
\newcommand{\nDevRoots}{8\xspace} %
\newcommand{\nDevStarts}{80\xspace} %
\newcommand{\seedTrain}{2000\xspace} %
\newcommand{\seedTest}{4000\xspace} %
\newcommand{\liberoxCommit}{f528726\xspace} %
\newcommand{\mujocoVersion}{2.3.0\xspace} %
\newcommand{\robosuiteVersion}{1.4.0\xspace} %
\newcommand{\nDeviations}{17\xspace} %
\newcommand{\nDevAfterTest}{0\xspace} %
\newcommand{\effectThreshold}{5\xspace} %
\newcommand{\nFamilyFiveDisc}{39\xspace} %
\newcommand{\nTrainPerConfig}{5\xspace} %
\newcommand{\nTestPerConfig}{8\xspace} %
\newcommand{\timeGateCollect}{1.5\xspace} %
\newcommand{\timeGateScreen}{1.1\xspace} %
\newcommand{\nTestPerConfigShip}{16\xspace} %
\newcommand{\nTrainPerConfigShip}{10\xspace} %
\newcommand{\liberoxPiLOne}{65.2\xspace} %
\newcommand{\liberoxPiLFive}{18.0\xspace} %

%% file: results.tex
\newcommand{\recLoBase}{22.2\xspace} %
\newcommand{\recHiBase}{7.6\xspace} %
\newcommand{\recAllBase}{16.8\xspace} %
\newcommand{\nFaultBase}{65\xspace} %
\newcommand{\normAbsBase}{43.9\xspace} %
\newcommand{\normAbsTBase}{33.6\xspace} %
\newcommand{\vsTBBase}{\ensuremath{-}29.7\xspace} %
\newcommand{\vsTBLoBase}{\ensuremath{-}44.4\xspace} %
\newcommand{\vsTBHiBase}{\ensuremath{-}18.5\xspace} %
\newcommand{\vsTBAbsBase}{29.7\xspace} %
\newcommand{\vsTBAbsLoBase}{18.5\xspace} %
\newcommand{\vsTBAbsHiBase}{44.4\xspace} %
\newcommand{\recLoDemo}{22.5\xspace} %
\newcommand{\recHiDemo}{6.0\xspace} %
\newcommand{\recAllDemo}{16.3\xspace} %
\newcommand{\nFaultDemo}{72\xspace} %
\newcommand{\normAbsDemo}{46.2\xspace} %
\newcommand{\normAbsTDemo}{34.1\xspace} %
\newcommand{\dnormDemo}{+2.3\xspace} %
\newcommand{\dnormLoDemo}{\ensuremath{-}1.1\xspace} %
\newcommand{\dnormHiDemo}{+7.1\xspace} %
\newcommand{\dnormAbsDemo}{2.3\xspace} %
\newcommand{\dnormAbsLoDemo}{\ensuremath{-}1.1\xspace} %
\newcommand{\dnormAbsHiDemo}{7.1\xspace} %
\newcommand{\dnormTDemo}{+0.5\xspace} %
\newcommand{\vsTBDemo}{\ensuremath{-}29.4\xspace} %
\newcommand{\vsTBLoDemo}{\ensuremath{-}44.7\xspace} %
\newcommand{\vsTBHiDemo}{\ensuremath{-}16.9\xspace} %
\newcommand{\vsTBAbsDemo}{29.4\xspace} %
\newcommand{\vsTBAbsLoDemo}{16.9\xspace} %
\newcommand{\vsTBAbsHiDemo}{44.7\xspace} %
\newcommand{\recLoTB}{51.9\xspace} %
\newcommand{\recHiTB}{36.8\xspace} %
\newcommand{\recAllTB}{46.3\xspace} %
\newcommand{\nFaultTB}{45\xspace} %
\newcommand{\normAbsTB}{14.0\xspace} %
\newcommand{\normAbsTTB}{0.5\xspace} %
\newcommand{\dnormTB}{\ensuremath{-}30.0\xspace} %
\newcommand{\dnormLoTB}{\ensuremath{-}37.4\xspace} %
\newcommand{\dnormHiTB}{\ensuremath{-}22.4\xspace} %
\newcommand{\dnormAbsTB}{30.0\xspace} %
\newcommand{\dnormAbsLoTB}{22.4\xspace} %
\newcommand{\dnormAbsHiTB}{37.4\xspace} %
\newcommand{\dnormTTB}{\ensuremath{-}33.2\xspace} %
\newcommand{\recLoFirst}{52.1\xspace} %
\newcommand{\recHiFirst}{36.3\xspace} %
\newcommand{\recAllFirst}{46.2\xspace} %
\newcommand{\nFaultFirst}{39\xspace} %
\newcommand{\normAbsFirst}{14.6\xspace} %
\newcommand{\normAbsTFirst}{0.7\xspace} %
\newcommand{\dnormFirst}{\ensuremath{-}29.3\xspace} %
\newcommand{\dnormLoFirst}{\ensuremath{-}36.8\xspace} %
\newcommand{\dnormHiFirst}{\ensuremath{-}22.1\xspace} %
\newcommand{\dnormAbsFirst}{29.3\xspace} %
\newcommand{\dnormAbsLoFirst}{22.1\xspace} %
\newcommand{\dnormAbsHiFirst}{36.8\xspace} %
\newcommand{\dnormTFirst}{\ensuremath{-}33.0\xspace} %
\newcommand{\vsTBFirst}{+0.1\xspace} %
\newcommand{\vsTBLoFirst}{\ensuremath{-}2.6\xspace} %
\newcommand{\vsTBHiFirst}{+2.2\xspace} %
\newcommand{\vsTBAbsFirst}{\ensuremath{-}0.1\xspace} %
\newcommand{\vsTBAbsLoFirst}{\ensuremath{-}2.2\xspace} %
\newcommand{\vsTBAbsHiFirst}{2.6\xspace} %
\newcommand{\recLoIQL}{53.5\xspace} %
\newcommand{\recHiIQL}{33.6\xspace} %
\newcommand{\recAllIQL}{46.0\xspace} %
\newcommand{\nFaultIQL}{51\xspace} %
\newcommand{\normAbsIQL}{19.7\xspace} %
\newcommand{\normAbsTIQL}{2.3\xspace} %
\newcommand{\dnormIQL}{\ensuremath{-}24.3\xspace} %
\newcommand{\dnormLoIQL}{\ensuremath{-}31.5\xspace} %
\newcommand{\dnormHiIQL}{\ensuremath{-}17.0\xspace} %
\newcommand{\dnormAbsIQL}{24.3\xspace} %
\newcommand{\dnormAbsLoIQL}{17.0\xspace} %
\newcommand{\dnormAbsHiIQL}{31.5\xspace} %
\newcommand{\dnormTIQL}{\ensuremath{-}31.4\xspace} %
\newcommand{\vsTBIQL}{+1.5\xspace} %
\newcommand{\vsTBLoIQL}{\ensuremath{-}2.2\xspace} %
\newcommand{\vsTBHiIQL}{+5.1\xspace} %
\newcommand{\vsTBAbsIQL}{\ensuremath{-}1.5\xspace} %
\newcommand{\vsTBAbsLoIQL}{\ensuremath{-}5.1\xspace} %
\newcommand{\vsTBAbsHiIQL}{2.2\xspace} %
\newcommand{\recLoAWAC}{53.8\xspace} %
\newcommand{\recHiAWAC}{30.6\xspace} %
\newcommand{\recAllAWAC}{45.1\xspace} %
\newcommand{\nFaultAWAC}{48\xspace} %
\newcommand{\normAbsAWAC}{14.4\xspace} %
\newcommand{\normAbsTAWAC}{0.7\xspace} %
\newcommand{\dnormAWAC}{\ensuremath{-}29.5\xspace} %
\newcommand{\dnormLoAWAC}{\ensuremath{-}37.6\xspace} %
\newcommand{\dnormHiAWAC}{\ensuremath{-}20.9\xspace} %
\newcommand{\dnormAbsAWAC}{29.5\xspace} %
\newcommand{\dnormAbsLoAWAC}{20.9\xspace} %
\newcommand{\dnormAbsHiAWAC}{37.6\xspace} %
\newcommand{\dnormTAWAC}{\ensuremath{-}33.0\xspace} %
\newcommand{\vsTBAWAC}{+1.8\xspace} %
\newcommand{\vsTBLoAWAC}{\ensuremath{-}4.4\xspace} %
\newcommand{\vsTBHiAWAC}{+9.3\xspace} %
\newcommand{\vsTBAbsAWAC}{\ensuremath{-}1.8\xspace} %
\newcommand{\vsTBAbsLoAWAC}{\ensuremath{-}9.3\xspace} %
\newcommand{\vsTBAbsHiAWAC}{4.4\xspace} %
\newcommand{\recLoIQLzero}{48.9\xspace} %
\newcommand{\recHiIQLzero}{32.6\xspace} %
\newcommand{\recAllIQLzero}{42.8\xspace} %
\newcommand{\nFaultIQLzero}{47\xspace} %
\newcommand{\normAbsIQLzero}{11.2\xspace} %
\newcommand{\normAbsTIQLzero}{0.5\xspace} %
\newcommand{\dnormIQLzero}{\ensuremath{-}32.7\xspace} %
\newcommand{\dnormLoIQLzero}{\ensuremath{-}42.9\xspace} %
\newcommand{\dnormHiIQLzero}{\ensuremath{-}23.8\xspace} %
\newcommand{\dnormAbsIQLzero}{32.7\xspace} %
\newcommand{\dnormAbsLoIQLzero}{23.8\xspace} %
\newcommand{\dnormAbsHiIQLzero}{42.9\xspace} %
\newcommand{\dnormTIQLzero}{\ensuremath{-}33.2\xspace} %
\newcommand{\vsTBIQLzero}{\ensuremath{-}3.1\xspace} %
\newcommand{\vsTBLoIQLzero}{\ensuremath{-}7.6\xspace} %
\newcommand{\vsTBHiIQLzero}{+1.1\xspace} %
\newcommand{\vsTBAbsIQLzero}{3.1\xspace} %
\newcommand{\vsTBAbsLoIQLzero}{\ensuremath{-}1.1\xspace} %
\newcommand{\vsTBAbsHiIQLzero}{7.6\xspace} %
\newcommand{\recLoMix}{45.3\xspace} %
\newcommand{\recHiMix}{26.2\xspace} %
\newcommand{\recAllMix}{38.1\xspace} %
\newcommand{\nFaultMix}{59\xspace} %
\newcommand{\normAbsMix}{27.5\xspace} %
\newcommand{\normAbsTMix}{2.7\xspace} %
\newcommand{\dnormMix}{\ensuremath{-}16.5\xspace} %
\newcommand{\dnormLoMix}{\ensuremath{-}24.9\xspace} %
\newcommand{\dnormHiMix}{\ensuremath{-}8.4\xspace} %
\newcommand{\dnormAbsMix}{16.5\xspace} %
\newcommand{\dnormAbsLoMix}{8.4\xspace} %
\newcommand{\dnormAbsHiMix}{24.9\xspace} %
\newcommand{\dnormTMix}{\ensuremath{-}30.9\xspace} %
\newcommand{\vsTBMix}{\ensuremath{-}6.7\xspace} %
\newcommand{\vsTBLoMix}{\ensuremath{-}11.4\xspace} %
\newcommand{\vsTBHiMix}{\ensuremath{-}3.3\xspace} %
\newcommand{\vsTBAbsMix}{6.7\xspace} %
\newcommand{\vsTBAbsLoMix}{3.3\xspace} %
\newcommand{\vsTBAbsHiMix}{11.4\xspace} %
\newcommand{\recLoIDQL}{40.7\xspace} %
\newcommand{\recHiIDQL}{17.8\xspace} %
\newcommand{\recAllIDQL}{32.1\xspace} %
\newcommand{\nFaultIDQL}{37\xspace} %
\newcommand{\normAbsIDQL}{2.7\xspace} %
\newcommand{\normAbsTIDQL}{0.0\xspace} %
\newcommand{\dnormIDQL}{\ensuremath{-}41.2\xspace} %
\newcommand{\dnormLoIDQL}{\ensuremath{-}52.9\xspace} %
\newcommand{\dnormHiIDQL}{\ensuremath{-}30.9\xspace} %
\newcommand{\dnormAbsIDQL}{41.2\xspace} %
\newcommand{\dnormAbsLoIDQL}{30.9\xspace} %
\newcommand{\dnormAbsHiIDQL}{52.9\xspace} %
\newcommand{\dnormTIDQL}{\ensuremath{-}33.6\xspace} %
\newcommand{\vsTBIDQL}{\ensuremath{-}11.2\xspace} %
\newcommand{\vsTBLoIDQL}{\ensuremath{-}19.6\xspace} %
\newcommand{\vsTBHiIDQL}{+3.5\xspace} %
\newcommand{\vsTBAbsIDQL}{11.2\xspace} %
\newcommand{\vsTBAbsLoIDQL}{\ensuremath{-}3.5\xspace} %
\newcommand{\vsTBAbsHiIDQL}{19.6\xspace} %
\newcommand{\DeltaOne}{29.4\xspace} %
\newcommand{\DeltaOneLo}{16.9\xspace} %
\newcommand{\DeltaOneHi}{44.7\xspace} %
\newcommand{\DeltaOnePctLo}{16.2\xspace} %
\newcommand{\DeltaOnePctHi}{44.0\xspace} %
\newcommand{\DeltaOneP}{$<10^{-4}$\xspace} %
\newcommand{\DeltaTwo}{1.5\xspace} %
\newcommand{\DeltaTwoLo}{\ensuremath{-}2.2\xspace} %
\newcommand{\DeltaTwoHi}{5.1\xspace} %
\newcommand{\DeltaTwoPctLo}{\ensuremath{-}2.2\xspace} %
\newcommand{\DeltaTwoPctHi}{5.1\xspace} %
\newcommand{\DeltaTwoP}{0.414\xspace} %
\newcommand{\OutcomeOne}{pass (+)\xspace} %
\newcommand{\OutcomeTwo}{inconclusive\xspace} %
\newcommand{\DeltaExtra}{+0.3\xspace} %
\newcommand{\DeltaExtraLo}{\ensuremath{-}1.9\xspace} %
\newcommand{\DeltaExtraHi}{+2.6\xspace} %
\newcommand{\DeltaRepair}{29.7\xspace} %
\newcommand{\DeltaRepairLo}{+18.5\xspace} %
\newcommand{\DeltaRepairHi}{+44.4\xspace} %
\newcommand{\DeltaMixDemo}{+22.8\xspace} %
\newcommand{\DeltaMixDemoLo}{+11.7\xspace} %
\newcommand{\DeltaMixDemoHi}{+39.3\xspace} %
\newcommand{\DeltaIQLzero}{+4.6\xspace} %
\newcommand{\DeltaIQLzeroLo}{+1.0\xspace} %
\newcommand{\DeltaIQLzeroHi}{+11.4\xspace} %
\newcommand{\DeltaOneSix}{+26.7\xspace} %
\newcommand{\DeltaOneSixLo}{+11.5\xspace} %
\newcommand{\DeltaOneSixHi}{+42.5\xspace} %
\newcommand{\DeltaTwoSix}{+2.6\xspace} %
\newcommand{\DeltaTwoSixLo}{\ensuremath{-}2.1\xspace} %
\newcommand{\DeltaTwoSixHi}{+6.4\xspace} %
\newcommand{\DeltaOneEighteen}{+30.7\xspace} %
\newcommand{\DeltaOneEighteenLo}{+18.1\xspace} %
\newcommand{\DeltaOneEighteenHi}{+45.7\xspace} %
\newcommand{\DeltaTwoEighteen}{+0.6\xspace} %
\newcommand{\DeltaTwoEighteenLo}{\ensuremath{-}2.9\xspace} %
\newcommand{\DeltaTwoEighteenHi}{+4.0\xspace} %
\newcommand{\lofoOneMin}{+24.5\xspace} %
\newcommand{\lofoOneMax}{+32.9\xspace} %
\newcommand{\nFamPosOne}{8\xspace} %
\newcommand{\nFamTieOne}{1\xspace} %
\newcommand{\nFamNegOne}{1\xspace} %
\newcommand{\lofoTwoMin}{+0.6\xspace} %
\newcommand{\lofoTwoMax}{+2.5\xspace} %
\newcommand{\nFamPosTwo}{5\xspace} %
\newcommand{\nFamTieTwo}{1\xspace} %
\newcommand{\nFamNegTwo}{4\xspace} %
\newcommand{\nNormValid}{437\xspace} %
\newcommand{\nNormExcl}{51\xspace} %
\newcommand{\nNormSens}{386\xspace} %
\newcommand{\nEvalRows}{10,368\xspace} %
\newcommand{\restoreMaxDiff}{0\xspace} %
\newcommand{\nRecRowsArm}{1,152\xspace} %
\newcommand{\nNormRowsAll}{4,500\xspace} %
\newcommand{\evalClientHours}{314\xspace} %
\newcommand{\nNormSharedTB}{51\xspace} %
\newcommand{\nNormSlowerTB}{48\xspace} %
\newcommand{\medNormSlowerTB}{179\xspace} %
\newcommand{\nFamNormZero}{5\xspace} %
\newcommand{\famNormZero}{0, 5, 7, 9, 15\xspace} %
\newcommand{\faultSensMaxShift}{1.3\xspace} %
\newcommand{\idqlBothN}{274\xspace} %
\newcommand{\idqlBothMedStep}{68\xspace} %
\newcommand{\tbBothMedStep}{138\xspace} %
\newcommand{\idqlFasterPct}{81.0\xspace} %
\newcommand{\idqlLostN}{259\xspace} %
\newcommand{\idqlLostTimeoutN}{241\xspace} %
\newcommand{\idqlLostLateN}{5\xspace} %
\newcommand{\idqlLostFaultN}{13\xspace} %
\newcommand{\idqlTBShortN}{134\xspace} %
\newcommand{\idqlTBShortRate}{79.9\xspace} %
\newcommand{\firstTBShortRate}{99.3\xspace} %
\newcommand{\idqlTBLongN}{147\xspace} %
\newcommand{\idqlTBLongRate}{21.8\xspace} %
\newcommand{\firstTBLongRate}{80.3\xspace} %
\newcommand{\nLofoTwoInside}{3\xspace} %
\newcommand{\lofoTwoInsideFams}{0, 7, 17\xspace} %
\newcommand{\lofoOneLoMin}{+13.3\xspace} %
\newcommand{\DeltaTwoEighteenInside}{yes\xspace} %
\newcommand{\rpQueries}{1,152\xspace} %
\newcommand{\rpSelTrans}{65.7\xspace} %
\newcommand{\rpSelTransLo}{63.2\xspace} %
\newcommand{\rpSelTransHi}{68.2\xspace} %
\newcommand{\rpSelRot}{60.5\xspace} %
\newcommand{\rpSelGrip}{49.0\xspace} %
\newcommand{\rpRhoTrans}{0.21\xspace} %
\newcommand{\rpTransGain}{18.0\xspace} %
\newcommand{\rpGapBelowTwin}{63.4\xspace} %
\newcommand{\rpHeadAgree}{29.1\xspace} %
\newcommand{\rpSelNN}{66.6\xspace} %
\newcommand{\rpQrangeRatio}{3.1\xspace} %
\newcommand{\rpMonoUp}{51.3\xspace} %
\newcommand{\rpMonoDown}{13.7\xspace} %
\newcommand{\rpRhoTransTrain}{0.25\xspace} %
\newcommand{\rpRhoTransPrefix}{0.02\xspace} %
\newcommand{\rpRhoRotPrefix}{0.19\xspace} %
\newcommand{\rpTrainFrames}{993,725\xspace} %
\newcommand{\nBaseNormFaultValid}{58\xspace} %

%% file: supp/tab_s_configs.tex
\newcommand{\suppAnnTrain}{297\xspace} %
\newcommand{\suppWhTrain}{25\xspace} %
\newcommand{\suppUndTrain}{236\xspace} %
\newcommand{\suppShipTrainMin}{9\xspace} %
\newcommand{\suppShipTrainMax}{15\xspace} %
\newcommand{\suppShipNominalCfgTrain}{45\xspace} %
\newcommand{\suppAnnTest}{456\xspace} %
\newcommand{\suppWhTest}{30\xspace} %
\newcommand{\suppUndTest}{367\xspace} %
\newcommand{\suppShipTestMin}{9\xspace} %
\newcommand{\suppShipTestMax}{20\xspace} %
\newcommand{\suppShipNominalCfgTest}{37\xspace} %
\newcommand{\suppCfgOverTestCap}{3\xspace} %
\newcommand{\suppCfgOverTrainNominal}{2\xspace} %
\newcommand{\suppAnnUnd}{69\xspace} %
\newcommand{\suppNotAnn}{534\xspace} %
\newcommand{\suppUndFamNine}{50\xspace} %
\newcommand{\suppWidthFlag}{13\xspace} %
\newcommand{\suppWidthFlagFamFive}{12\xspace} %
\newcommand{\suppWhNoFailure}{26\xspace} %
\newcommand{\suppWhUncertain}{19\xspace} %
\newcommand{\suppWhFailure}{10\xspace} %
\newcommand{\suppArchRel}{3\xspace} %
\newcommand{\suppArchRevoked}{1\xspace} %
\newcommand{\suppArchSucc}{6\xspace} %
\newcommand{\suppArchStop}{15\xspace} %
\newcommand{\suppArchTL}{3\xspace} %
\newcommand{\suppStopsAll}{324\xspace} %
\newcommand{\suppTLAll}{7\xspace} %
\newcommand{\suppPending}{2\xspace} %
\newcommand{\suppOutOfInterval}{3\xspace} %
\newcommand{\suppOverCap}{1\xspace} %
\newcommand{\suppAttemptCap}{20\xspace} %
\newcommand{\suppFamNineLFourRel}{2\xspace} %
\newcommand{\suppFamNineLFourAtt}{2\xspace} %
\newcommand{\suppFamDiag}{9\xspace} %
\newcommand{\suppHSRFamMin}{83.7\xspace} %
\newcommand{\suppHSRFamMax}{98.0\xspace} %
\newcommand{\suppConfigRows}{%
0 (S) & L1 & 10/16 & 5/8 & 0/0 & 5 & 0 & 0 & 58 & 50 & 1 & 7 & 0 & 0 & 87.9 \\
 & L2 & 10/15 & 6/8 & 1/0 & 5 & 2 (11) & 0 & 57 & 50 & 0 & 7 & 0 & 0 & 87.7 \\
 & L3 & 10/16 & 5/8 & 0/0 & 5 & 0 & 0 & 60 & 50 & 0 & 9 & 1 & 0 & 83.3 \\
 & L4 & 10/16 & 5/8 & 1/0 & 5 & 0 & 3 & 64 & 50 & 0 & 10 & 1 & 3 & 82.0 \\
 & L5 & 10/16 & 6/8 & 0/0 & 5 & 1 (11) & 1 & 67 & 50 & 0 & 14 & 0 & 3 & 78.1 \\
 & \textit{all} & 50/79 & 27/40 & 2/0 & 25 & 3 (22) & 4 & 306 & 250 & 1 & 47 & 2 & 6 & 83.7 \\
\midrule
4 (S) & L1 & 9/16 & 5/8 & 0/0 & 5 & 0 & 0 & 55 & 50 & 1 & 4 & 0 & 0 & 92.7 \\
 & L2 & 10/16 & 5/8 & 0/0 & 5 & 0 & 0 & 57 & 50 & 0 & 6 & 0 & 1 & 89.3 \\
 & L3 & 10/16 & 5/8 & 0/0 & 5 & 0 & 0 & 53 & 50 & 0 & 2 & 0 & 1 & 96.2 \\
 & L4 & 10/16 & 5/8 & 0/0 & 5 & 0 & 0 & 55 & 50 & 1 & 4 & 0 & 0 & 92.7 \\
 & L5 & 10/16 & 5/8 & 0/0 & 5 & 0 & 0 & 57 & 50 & 0 & 6 & 0 & 1 & 89.3 \\
 & \textit{all} & 49/80 & 25/40 & 0/0 & 25 & 0 & 0 & 277 & 250 & 2 & 22 & 0 & 3 & 92.0 \\
\midrule
5 (S) & L1 & 10/16 & 5/8 & 1/0 & 5 & 0 & 0 & 61 & 50 & 10 & 1 & 0 & 0 & 98.4 \\
 & L2 & 10/15 & 5/8 & 2/0 & 5 & 0 & 0 & 55 & 50 & 5 & 0 & 0 & 0 & 100.0 \\
 & L3 & 10/14 & 5/8 & 1/2 & 5 & 0 & 0 & 66 & 50 & 11 & 5 & 0 & 0 & 92.4 \\
 & L4 & 10/15 & 5/8 & 0/1 & 5 & 0 & 0 & 55 & 50 & 5 & 0 & 0 & 0 & 100.0 \\
 & L5 & 10/14 & 5/8 & 1/1 & 5 & 0 & 0 & 58 & 50 & 8 & 0 & 0 & 0 & 100.0 \\
 & \textit{all} & 50/74 & 25/40 & 5/4 & 25 & 0 & 0 & 295 & 250 & 39 & 6 & 0 & 0 & 98.0 \\
\midrule
6 (S) & L1 & 10/9 & 5/8 & 0/0 & 5 & 0 & 0 & 57 & 50 & 2 & 5 & 0 & 0 & 91.2 \\
 & L2 & 10/16 & 5/8 & 0/0 & 5 & 0 & 0 & 67 & 50 & 1 & 16 & 0 & 0 & 76.1 \\
 & L3 & 10/16 & 5/8 & 0/0 & 5 & 0 & 0 & 52 & 50 & 1 & 1 & 0 & 0 & 98.1 \\
 & L4 & 10/16 & 5/8 & 0/0 & 5 & 0 & 0 & 55 & 50 & 2 & 3 & 0 & 0 & 94.5 \\
 & L5 & 10/16 & 5/8 & 0/0 & 5 & 0 & 0 & 59 & 50 & 0 & 9 & 0 & 0 & 84.7 \\
 & \textit{all} & 50/73 & 25/40 & 0/0 & 25 & 0 & 0 & 290 & 250 & 6 & 34 & 0 & 0 & 88.3 \\
\midrule
7 (S) & L1 & 10/16 & 5/8 & 0/0 & 5 & 0 & 0 & 50 & 50 & 0 & 0 & 0 & 0 & 100.0 \\
 & L2 & 10/16 & 6/8 & 0/1 & 5 & 1 (2) & 0 & 53 & 50 & 0 & 3 & 0 & 0 & 94.3 \\
 & L3 & 10/13 & 5/8 & 0/3 & 5 & 0 & 0 & 56 & 50 & 0 & 6 & 0 & 0 & 89.3 \\
 & L4 & 10/12 & 5/8 & 0/3 & 5 & 0 & 0 & 58 & 50 & 1 & 7 & 0 & 0 & 87.9 \\
 & L5 & 10/16 & 5/8 & 0/2 & 5 & 0 & 0 & 63 & 50 & 3 & 10 & 0 & 0 & 84.1 \\
 & \textit{all} & 50/73 & 26/40 & 0/9 & 25 & 1 (2) & 0 & 280 & 250 & 4 & 26 & 0 & 0 & 90.7 \\
\midrule
9 (K) & L1 & 10/16 & 5/8 & 0/2 & 5 & 0 & 0 & 66 & 50 & 1 & 15 & 0 & 0 & 77.3 \\
 & L2 & 10/16 & 5/8 & 4/0 & 5 & 0 & 0 & 58 & 50 & 0 & 8 & 0 & 0 & 86.2 \\
 & L3 & 10/16 & 5/8 & 0/0 & 5 & 0 & 0 & 52 & 50 & 1 & 1 & 0 & 0 & 98.1 \\
 & L4$^{a}$ & 10/16 & 2/0 & 0/0 & 2 & 0 & 0 & 2 & 0 & 0 & 0 & 2 & 0 & 0.0 \\
 & L5$^{a}$ & 10/16 & 0/0 & 0/0 & 0 & 0 & 0 & 0 & 0 & 0 & 0 & 0 & 0 & -- \\
 & \textit{all} & 50/80 & 17/24 & 4/2 & 17 & 0 & 0 & 178 & 150 & 2 & 24 & 2 & 0 & 85.4 \\
\midrule
12 (S) & L1 & 10/16 & 5/8 & 0/0 & 5 & 0 & 0 & 67 & 50 & 0 & 17 & 0 & 0 & 74.6 \\
 & L2 & 9/16 & 5/8 & 0/0 & 5 & 0 & 0 & 60 & 50 & 0 & 10 & 0 & 0 & 83.3 \\
 & L3 & 10/16 & 5/8 & 0/0 & 5 & 0 & 0 & 60 & 50 & 2 & 8 & 0 & 0 & 86.7 \\
 & L4 & 10/16 & 5/8 & 0/0 & 5 & 0 & 0 & 57 & 50 & 0 & 4 & 0 & 3 & 92.6 \\
 & L5 & 10/16 & 5/8 & 0/0 & 5 & 0 & 0 & 59 & 50 & 1 & 8 & 0 & 0 & 86.4 \\
 & \textit{all} & 49/80 & 25/40 & 0/0 & 25 & 0 & 0 & 303 & 250 & 3 & 47 & 0 & 3 & 84.3 \\
\midrule
14 (K) & L1 & 10/16 & 5/8 & 0/0 & 5 & 0 & 0 & 55 & 50 & 1 & 4 & 0 & 0 & 92.7 \\
 & L2 & 10/16 & 5/8 & 0/0 & 5 & 0 & 0 & 61 & 50 & 0 & 11 & 0 & 0 & 82.0 \\
 & L3 & 10/16 & 5/8 & 0/0 & 5 & 0 & 0 & 62 & 50 & 0 & 12 & 0 & 0 & 80.6 \\
 & L4 & 10/16 & 5/8 & 0/0 & 5 & 0 & 0 & 52 & 50 & 0 & 2 & 0 & 0 & 96.2 \\
 & L5 & 10/16 & 5/8 & 0/0 & 5 & 0 & 0 & 51 & 50 & 0 & 1 & 0 & 0 & 98.0 \\
 & \textit{all} & 50/80 & 25/40 & 0/0 & 25 & 0 & 0 & 281 & 250 & 1 & 30 & 0 & 0 & 89.3 \\
\midrule
15 (S) & L1 & 10/14 & 5/8 & 0/0 & 5 & 0 & 0 & 57 & 50 & 2 & 5 & 0 & 0 & 91.2 \\
 & L2 & 10/13 & 5/8 & 0/0 & 5 & 0 & 0 & 67 & 50 & 0 & 17 & 0 & 0 & 74.6 \\
 & L3 & 10/16 & 5/8 & 0/0 & 5 & 0 & 0 & 52 & 50 & 0 & 2 & 0 & 0 & 96.2 \\
 & L4 & 10/16 & 5/8 & 0/0 & 5 & 0 & 0 & 57 & 50 & 0 & 6 & 0 & 1 & 89.3 \\
 & L5 & 10/16 & 5/8 & 0/0 & 5 & 0 & 0 & 57 & 50 & 0 & 5 & 0 & 2 & 90.9 \\
 & \textit{all} & 50/75 & 25/40 & 0/0 & 25 & 0 & 0 & 290 & 250 & 2 & 35 & 0 & 3 & 87.8 \\
\midrule
17 (K) & L1 & 14/18 & 5/8 & 7/0 & 5 & 0 & 0 & 63 & 50 & 3 & 10 & 0 & 0 & 84.1 \\
 & L2 & 15/20 & 5/8 & 6/2 & 5 & 0 & 0 & 57 & 50 & 4 & 3 & 0 & 0 & 94.7 \\
 & L3 & 9/17 & 5/8 & 1/9 & 5 & 0 & 0 & 63 & 50 & 2 & 11 & 0 & 0 & 82.5 \\
 & L4 & 10/16 & 5/8 & 0/4 & 5 & 0 & 0 & 56 & 50 & 0 & 6 & 0 & 0 & 89.3 \\
 & L5 & 10/16 & 5/8 & 0/0 & 5 & 0 & 0 & 60 & 50 & 2 & 8 & 0 & 0 & 86.7 \\
 & \textit{all} & 58/87 & 25/40 & 14/15 & 25 & 0 & 0 & 299 & 250 & 11 & 38 & 0 & 0 & 87.3 \\
\midrule
\multicolumn{2}{l}{All} & 506/781 & 245/384 & 25/30 & 242 & 4 (24) & 4 & 2,799 & 2,400 & 71 & 309 & 4 & 15 & 88.8 \\
}
\newcommand{\suppFunnelRows}{%
0 (S) & L1 & 10 & 6 & 5 & 0 & 5 & 16 & 8 & 8 & 0 & 8 & 0 \\
 & L2 & 10 & 8 & 6 & 1 & 3 & 15 & 8 & 8 & 0 & 7 & 0 \\
 & L3 & 10 & 8 & 5 & 0 & 5 & 16 & 9 & 8 & 0 & 8 & 0 \\
 & L4 & 10 & 6 & 5 & 1 & 4 & 16 & 9 & 8 & 0 & 8 & 1 \\
 & L5 & 10 & 6 & 6 & 0 & 4 & 16 & 8 & 8 & 0 & 8 & 0 \\
\midrule
4 (S) & L1 & 9 & 5 & 5 & 0 & 4 & 16 & 8 & 8 & 0 & 8 & 0 \\
 & L2 & 10 & 5 & 5 & 0 & 5 & 16 & 8 & 8 & 0 & 8 & 0 \\
 & L3 & 10 & 5 & 5 & 0 & 5 & 16 & 8 & 8 & 0 & 8 & 0 \\
 & L4 & 10 & 5 & 5 & 0 & 5 & 16 & 8 & 8 & 0 & 8 & 0 \\
 & L5 & 10 & 5 & 5 & 0 & 5 & 16 & 8 & 8 & 0 & 8 & 0 \\
\midrule
5 (S) & L1 & 10 & 6 & 5 & 1 & 4 & 16 & 8 & 8 & 0 & 8 & 1 \\
 & L2 & 10 & 7 & 5 & 2 & 3 & 15 & 8 & 8 & 0 & 7 & 3 \\
 & L3 & 10 & 6 & 5 & 1 & 4 & 14 & 10 & 8 & 2 & 4 & 2 \\
 & L4 & 10 & 5 & 5 & 0 & 5 & 15 & 9 & 8 & 1 & 6 & 3 \\
 & L5 & 10 & 6 & 5 & 1 & 4 & 14 & 9 & 8 & 1 & 5 & 3 \\
\midrule
6 (S) & L1 & 10 & 5 & 5 & 0 & 5 & 9 & 9 & 8 & 0 & 1 & 0 \\
 & L2 & 10 & 5 & 5 & 0 & 5 & 16 & 8 & 8 & 0 & 8 & 0 \\
 & L3 & 10 & 5 & 5 & 0 & 5 & 16 & 8 & 8 & 0 & 8 & 0 \\
 & L4 & 10 & 5 & 5 & 0 & 5 & 16 & 8 & 8 & 0 & 8 & 0 \\
 & L5 & 10 & 5 & 5 & 0 & 5 & 16 & 8 & 8 & 0 & 8 & 0 \\
\midrule
7 (S) & L1 & 10 & 5 & 5 & 0 & 5 & 16 & 8 & 8 & 0 & 8 & 0 \\
 & L2 & 10 & 6 & 6 & 0 & 4 & 16 & 9 & 8 & 1 & 7 & 0 \\
 & L3 & 10 & 5 & 5 & 0 & 5 & 13 & 11 & 8 & 3 & 2 & 0 \\
 & L4 & 10 & 5 & 5 & 0 & 5 & 12 & 11 & 8 & 3 & 1 & 0 \\
 & L5 & 10 & 5 & 5 & 0 & 5 & 16 & 10 & 8 & 2 & 6 & 0 \\
\midrule
9 (K) & L1 & 10 & 10 & 5 & 0 & 5 & 16 & 16 & 8 & 2 & 6 & 0 \\
 & L2 & 10 & 10 & 5 & 4 & 1 & 16 & 16 & 8 & 0 & 8 & 0 \\
 & L3 & 10 & 10 & 5 & 0 & 5 & 16 & 16 & 8 & 0 & 8 & 0 \\
 & L4$^{a}$ & 10 & 2 & 2 & 0 & 8 & 16 & 2 & 0 & 0 & 16 & 0 \\
 & L5$^{a}$ & 10 & 2 & 0 & 0 & 10 & 16 & 0 & 0 & 0 & 16 & 0 \\
\midrule
12 (S) & L1 & 10 & 5 & 5 & 0 & 5 & 16 & 8 & 8 & 0 & 8 & 0 \\
 & L2 & 9 & 5 & 5 & 0 & 4 & 16 & 8 & 8 & 0 & 8 & 0 \\
 & L3 & 10 & 5 & 5 & 0 & 5 & 16 & 8 & 8 & 0 & 8 & 0 \\
 & L4 & 10 & 5 & 5 & 0 & 5 & 16 & 8 & 8 & 0 & 8 & 0 \\
 & L5 & 10 & 5 & 5 & 0 & 5 & 16 & 8 & 8 & 0 & 8 & 0 \\
\midrule
14 (K) & L1 & 10 & 5 & 5 & 0 & 5 & 16 & 8 & 8 & 0 & 8 & 0 \\
 & L2 & 10 & 5 & 5 & 0 & 5 & 16 & 8 & 8 & 0 & 8 & 0 \\
 & L3 & 10 & 5 & 5 & 0 & 5 & 16 & 8 & 8 & 0 & 8 & 0 \\
 & L4 & 10 & 5 & 5 & 0 & 5 & 16 & 8 & 8 & 0 & 8 & 0 \\
 & L5 & 10 & 5 & 5 & 0 & 5 & 16 & 8 & 8 & 0 & 8 & 0 \\
\midrule
15 (S) & L1 & 10 & 5 & 5 & 0 & 5 & 14 & 8 & 8 & 0 & 6 & 0 \\
 & L2 & 10 & 5 & 5 & 0 & 5 & 13 & 8 & 8 & 0 & 5 & 0 \\
 & L3 & 10 & 5 & 5 & 0 & 5 & 16 & 8 & 8 & 0 & 8 & 0 \\
 & L4 & 10 & 5 & 5 & 0 & 5 & 16 & 8 & 8 & 0 & 8 & 0 \\
 & L5 & 10 & 5 & 5 & 0 & 5 & 16 & 8 & 8 & 0 & 8 & 0 \\
\midrule
17 (K) & L1 & 14 & 14 & 5 & 7 & 2 & 18 & 18 & 8 & 0 & 10 & 0 \\
 & L2 & 15 & 15 & 5 & 6 & 4 & 20 & 14 & 8 & 2 & 10 & 0 \\
 & L3 & 9 & 9 & 5 & 1 & 3 & 17 & 17 & 8 & 9 & 0 & 0 \\
 & L4 & 10 & 5 & 5 & 0 & 5 & 16 & 13 & 8 & 4 & 4 & 0 \\
 & L5 & 10 & 5 & 5 & 0 & 5 & 16 & 8 & 8 & 0 & 8 & 0 \\
\midrule
\multicolumn{2}{l}{All} & 506 & 297 & 245 & 25 & 236 & 781 & 456 & 384 & 30 & 367 & 13 \\
}

%% file: sec/section5_numbers.tex
\newcommand{\secFiveNomLoDemo}{\ensuremath{-}2.0\xspace}
\newcommand{\secFiveNomHiDemo}{7.0\xspace}
\newcommand{\secFiveNomLoTB}{\ensuremath{-}37.9\xspace}
\newcommand{\secFiveNomHiTB}{\ensuremath{-}22.6\xspace}
\newcommand{\secFiveNomLoMix}{\ensuremath{-}22.4\xspace}
\newcommand{\secFiveNomHiMix}{\ensuremath{-}10.0\xspace}
\newcommand{\secFiveNomLoIQL}{\ensuremath{-}31.6\xspace}
\newcommand{\secFiveNomHiIQL}{\ensuremath{-}16.3\xspace}

\newcommand{\secFiveMixGainMain}{13.5\xspace}
\newcommand{\secFiveMixGainOriginal}{2.3\xspace}
\newcommand{\secFiveFamMin}{\ensuremath{-}1.4\xspace}
\newcommand{\secFiveFamMax}{+73.6\xspace}
\newcommand{\secFiveRecoveryWeight}{1.54\xspace}
\newcommand{\secFiveFailureWeight}{5.63\xspace}

%% file: supp/revision_numbers.tex
\newcommand{\nRecChunksPost}{776,507\xspace} %
\newcommand{\nRecChunksLo}{409,862\xspace} %
\newcommand{\nRecChunksHi}{366,645\xspace} %
\newcommand{\pctRecChunksLo}{52.8\xspace} %
\newcommand{\nPrefixFramesPost}{70,306\xspace} %
\newcommand{\nFwinChunksPost}{11,926\xspace} %
\newcommand{\nUnsuccEpsPost}{303\xspace} %
\newcommand{\nUnsuccFramesPost}{126,034\xspace} %
\newcommand{\nEligChunksAll}{788,433\xspace} %
\newcommand{\pctEligUnwRec}{98.5\xspace} %
\newcommand{\pctEligUnwFwin}{1.5\xspace} %
\newcommand{\pctDrawRecIQL}{94.7\xspace} %
\newcommand{\pctDrawFwinIQL}{5.3\xspace} %
\newcommand{\essRecIQLPct}{5.4\xspace} %
\newcommand{\wMeanRecAWAC}{1.54\xspace} %
\newcommand{\wMeanFwinAWAC}{1.41\xspace} %
\newcommand{\pctDrawRecAWAC}{98.6\xspace} %
\newcommand{\pctDrawFwinAWAC}{1.4\xspace} %
\newcommand{\clipRecAWACPct}{0.29\xspace} %
\newcommand{\essRecAWACPct}{6.3\xspace} %
\newcommand{\wMeanRecIQLzero}{0.98\xspace} %
\newcommand{\clipRecIQLzeroPct}{0.07\xspace} %
\newcommand{\essRecIQLzeroPct}{7.7\xspace} %
\newcommand{\nNormSharedTBDemo}{53\xspace} %
\newcommand{\nNormSlowerTBDemo}{51\xspace} %
\newcommand{\medNormSlowerTBDemo}{184\xspace} %
\newcommand{\nomLatePctBase}{10.3\xspace} %
\newcommand{\nomLateNBase}{45\xspace} %
\newcommand{\nomFastNBase}{147\xspace} %
\newcommand{\nomSuccNBase}{192\xspace} %
\newcommand{\nomTimeoutPctBase}{42.8\xspace} %
\newcommand{\nomFaultPctBase}{13.3\xspace} %
\newcommand{\nomFaultStableNBase}{7\xspace} %
\newcommand{\nomSensSRBase}{49.7\xspace} %
\newcommand{\nomMedRatioTBase}{1.00\xspace} %
\newcommand{\dnormVsDemoBase}{\ensuremath{-}2.3\xspace} %
\newcommand{\dnormVsDemoAbsBase}{2.3\xspace} %
\newcommand{\dnormVsDemoLoBase}{\ensuremath{-}6.5\xspace} %
\newcommand{\dnormVsDemoHiBase}{+1.8\xspace} %
\newcommand{\dnormTVsDemoBase}{\ensuremath{-}0.5\xspace} %
\newcommand{\nomLatePctDemo}{12.1\xspace} %
\newcommand{\nomLateNDemo}{53\xspace} %
\newcommand{\nomFastNDemo}{149\xspace} %
\newcommand{\nomSuccNDemo}{202\xspace} %
\newcommand{\nomTimeoutPctDemo}{40.5\xspace} %
\newcommand{\nomFaultPctDemo}{13.3\xspace} %
\newcommand{\nomFaultStableNDemo}{7\xspace} %
\newcommand{\nomSensSRDemo}{52.3\xspace} %
\newcommand{\nomMedRatioTDemo}{1.01\xspace} %
\newcommand{\nomLatePctTB}{13.5\xspace} %
\newcommand{\nomLateNTB}{59\xspace} %
\newcommand{\nomFastNTB}{2\xspace} %
\newcommand{\nomSuccNTB}{61\xspace} %
\newcommand{\nomTimeoutPctTB}{73.5\xspace} %
\newcommand{\nomFaultPctTB}{12.6\xspace} %
\newcommand{\nomFaultStableNTB}{4\xspace} %
\newcommand{\nomSensSRTB}{15.8\xspace} %
\newcommand{\nomMedRatioTTB}{1.77\xspace} %
\newcommand{\dnormVsDemoTB}{\ensuremath{-}32.3\xspace} %
\newcommand{\dnormVsDemoAbsTB}{32.3\xspace} %
\newcommand{\dnormVsDemoLoTB}{\ensuremath{-}39.8\xspace} %
\newcommand{\dnormVsDemoHiTB}{\ensuremath{-}25.6\xspace} %
\newcommand{\dnormTVsDemoTB}{\ensuremath{-}33.6\xspace} %
\newcommand{\nomLatePctMix}{24.7\xspace} %
\newcommand{\nomLateNMix}{108\xspace} %
\newcommand{\nomFastNMix}{12\xspace} %
\newcommand{\nomSuccNMix}{120\xspace} %
\newcommand{\nomTimeoutPctMix}{59.7\xspace} %
\newcommand{\nomFaultPctMix}{12.8\xspace} %
\newcommand{\nomFaultStableNMix}{5\xspace} %
\newcommand{\nomSensSRMix}{31.1\xspace} %
\newcommand{\nomMedRatioTMix}{1.55\xspace} %
\newcommand{\dnormVsDemoMix}{\ensuremath{-}18.8\xspace} %
\newcommand{\dnormVsDemoAbsMix}{18.8\xspace} %
\newcommand{\dnormVsDemoLoMix}{\ensuremath{-}25.3\xspace} %
\newcommand{\dnormVsDemoHiMix}{\ensuremath{-}12.3\xspace} %
\newcommand{\dnormTVsDemoMix}{\ensuremath{-}31.4\xspace} %
\newcommand{\nomLatePctIQL}{17.4\xspace} %
\newcommand{\nomLateNIQL}{76\xspace} %
\newcommand{\nomFastNIQL}{10\xspace} %
\newcommand{\nomSuccNIQL}{86\xspace} %
\newcommand{\nomTimeoutPctIQL}{66.4\xspace} %
\newcommand{\nomFaultPctIQL}{14.0\xspace} %
\newcommand{\nomFaultStableNIQL}{10\xspace} %
\newcommand{\nomSensSRIQL}{22.3\xspace} %
\newcommand{\nomMedRatioTIQL}{1.52\xspace} %
\newcommand{\dnormVsDemoIQL}{\ensuremath{-}26.5\xspace} %
\newcommand{\dnormVsDemoAbsIQL}{26.5\xspace} %
\newcommand{\dnormVsDemoLoIQL}{\ensuremath{-}34.5\xspace} %
\newcommand{\dnormVsDemoHiIQL}{\ensuremath{-}19.0\xspace} %
\newcommand{\dnormTVsDemoIQL}{\ensuremath{-}31.8\xspace} %
\newcommand{\nomLatePctIQLzero}{10.8\xspace} %
\newcommand{\nomLateNIQLzero}{47\xspace} %
\newcommand{\nomFastNIQLzero}{2\xspace} %
\newcommand{\nomSuccNIQLzero}{49\xspace} %
\newcommand{\nomTimeoutPctIQLzero}{75.5\xspace} %
\newcommand{\nomFaultPctIQLzero}{13.3\xspace} %
\newcommand{\nomFaultStableNIQLzero}{7\xspace} %
\newcommand{\nomSensSRIQLzero}{12.7\xspace} %
\newcommand{\nomMedRatioTIQLzero}{1.74\xspace} %
\newcommand{\dnormVsDemoIQLzero}{\ensuremath{-}35.0\xspace} %
\newcommand{\dnormVsDemoAbsIQLzero}{35.0\xspace} %
\newcommand{\dnormVsDemoLoIQLzero}{\ensuremath{-}43.8\xspace} %
\newcommand{\dnormVsDemoHiIQLzero}{\ensuremath{-}26.7\xspace} %
\newcommand{\dnormTVsDemoIQLzero}{\ensuremath{-}33.6\xspace} %
\newcommand{\nomLatePctAWAC}{13.7\xspace} %
\newcommand{\nomLateNAWAC}{60\xspace} %
\newcommand{\nomFastNAWAC}{3\xspace} %
\newcommand{\nomSuccNAWAC}{63\xspace} %
\newcommand{\nomTimeoutPctAWAC}{71.9\xspace} %
\newcommand{\nomFaultPctAWAC}{13.7\xspace} %
\newcommand{\nomFaultStableNAWAC}{9\xspace} %
\newcommand{\nomSensSRAWAC}{16.3\xspace} %
\newcommand{\nomMedRatioTAWAC}{1.84\xspace} %
\newcommand{\dnormVsDemoAWAC}{\ensuremath{-}31.8\xspace} %
\newcommand{\dnormVsDemoAbsAWAC}{31.8\xspace} %
\newcommand{\dnormVsDemoLoAWAC}{\ensuremath{-}39.9\xspace} %
\newcommand{\dnormVsDemoHiAWAC}{\ensuremath{-}23.7\xspace} %
\newcommand{\dnormTVsDemoAWAC}{\ensuremath{-}33.4\xspace} %
\newcommand{\nomLatePctFirst}{14.0\xspace} %
\newcommand{\nomLateNFirst}{61\xspace} %
\newcommand{\nomFastNFirst}{3\xspace} %
\newcommand{\nomSuccNFirst}{64\xspace} %
\newcommand{\nomTimeoutPctFirst}{73.5\xspace} %
\newcommand{\nomFaultPctFirst}{11.9\xspace} %
\newcommand{\nomFaultStableNFirst}{1\xspace} %
\newcommand{\nomSensSRFirst}{16.6\xspace} %
\newcommand{\nomMedRatioTFirst}{1.75\xspace} %
\newcommand{\dnormVsDemoFirst}{\ensuremath{-}31.6\xspace} %
\newcommand{\dnormVsDemoAbsFirst}{31.6\xspace} %
\newcommand{\dnormVsDemoLoFirst}{\ensuremath{-}39.2\xspace} %
\newcommand{\dnormVsDemoHiFirst}{\ensuremath{-}24.9\xspace} %
\newcommand{\dnormTVsDemoFirst}{\ensuremath{-}33.4\xspace} %
\newcommand{\nomLatePctIDQL}{2.7\xspace} %
\newcommand{\nomLateNIDQL}{12\xspace} %
\newcommand{\nomFastNIDQL}{0\xspace} %
\newcommand{\nomSuccNIDQL}{12\xspace} %
\newcommand{\nomTimeoutPctIDQL}{82.8\xspace} %
\newcommand{\nomFaultPctIDQL}{14.4\xspace} %
\newcommand{\nomFaultStableNIDQL}{12\xspace} %
\newcommand{\nomSensSRIDQL}{3.1\xspace} %
\newcommand{\nomMedRatioTIDQL}{1.75\xspace} %
\newcommand{\dnormVsDemoIDQL}{\ensuremath{-}43.5\xspace} %
\newcommand{\dnormVsDemoAbsIDQL}{43.5\xspace} %
\newcommand{\dnormVsDemoLoIDQL}{\ensuremath{-}53.2\xspace} %
\newcommand{\dnormVsDemoHiIDQL}{\ensuremath{-}33.1\xspace} %
\newcommand{\dnormTVsDemoIDQL}{\ensuremath{-}34.1\xspace} %
\newcommand{\demoTransNormMed}{0.28\xspace} %
\newcommand{\demoTransNormPninety}{0.63\xspace} %
\newcommand{\demoTransNormPnn}{1.10\xspace} %
\newcommand{\tkTransNormMed}{0.21\xspace} %
\newcommand{\tkTransNormPninety}{0.29\xspace} %
\newcommand{\tkTransNormPnn}{0.35\xspace} %
\newcommand{\tbPredStepsAll}{368,640\xspace} %
\newcommand{\tbPredPNNTrans}{0.22--0.27\xspace} %
\newcommand{\tbPredPNNRot}{0.06--0.07\xspace} %
\newcommand{\tbPredAboveTrans}{0.2--3.7\xspace} %
\newcommand{\tbPredAboveRot}{0.2--0.5\xspace} %
\newcommand{\tbPredTransNormMed}{0.17\xspace} %
\newcommand{\tbPredTransNormPninety}{0.26\xspace} %
\newcommand{\tbPredTransNormPnn}{0.30\xspace} %
\newcommand{\nFlagWidthLo}{6\xspace} %
\newcommand{\nFlagUnionLo}{17\xspace} %
\newcommand{\nRetainedUnionLo}{223\xspace} %
\newcommand{\qOneUnionEst}{+31.4\xspace} %
\newcommand{\qOneUnionLo}{+18.0\xspace} %
\newcommand{\qOneUnionHi}{+46.4\xspace} %
\newcommand{\qTwoUnionEst}{+2.7\xspace} %
\newcommand{\qTwoUnionLo}{\ensuremath{-}1.3\xspace} %
\newcommand{\qTwoUnionHi}{+6.3\xspace} %
\newcommand{\nBaseZeroLoPts}{154\xspace} %
\newcommand{\nBasePartLoPts}{62\xspace} %
\newcommand{\nBaseAllLoPts}{24\xspace} %
\newcommand{\srZeroLoDemo}{3.7\xspace} %
\newcommand{\srZeroLoTB}{46.8\xspace} %
\newcommand{\srZeroLoIQL}{47.4\xspace} %
\newcommand{\nBaseZeroAllPts}{276\xspace} %
\newcommand{\srZeroAllTB}{40.5\xspace} %

%% file: soulreg.tex
\soulregister{\cite}{7}\soulregister{\citep}{7}\soulregister{\citet}{7}
\soulregister{\cref}{7}\soulregister{\Cref}{7}\soulregister{\ref}{7}\soulregister{\eqref}{7}
\soulregister{\emph}{1}\soulregister{\textbf}{1}\soulregister{\textit}{1}\soulregister{\texttt}{1}\soulregister{\textsc}{1}
\soulregister{\BCstar}{0}
\soulregister{\BaseRestart}{0}
\soulregister{\DeltaExtra}{0}
\soulregister{\DeltaExtraHi}{0}
\soulregister{\DeltaExtraLo}{0}
\soulregister{\DeltaIQLzero}{0}
\soulregister{\DeltaIQLzeroHi}{0}
\soulregister{\DeltaIQLzeroLo}{0}
\soulregister{\DeltaMixDemo}{0}
\soulregister{\DeltaMixDemoHi}{0}
\soulregister{\DeltaMixDemoLo}{0}
\soulregister{\DeltaOne}{0}
\soulregister{\DeltaOneEighteen}{0}
\soulregister{\DeltaOneEighteenHi}{0}
\soulregister{\DeltaOneEighteenLo}{0}
\soulregister{\DeltaOneHi}{0}
\soulregister{\DeltaOneLo}{0}
\soulregister{\DeltaOneP}{0}
\soulregister{\DeltaOnePctHi}{0}
\soulregister{\DeltaOnePctLo}{0}
\soulregister{\DeltaOneSix}{0}
\soulregister{\DeltaOneSixHi}{0}
\soulregister{\DeltaOneSixLo}{0}
\soulregister{\DeltaRepair}{0}
\soulregister{\DeltaRepairHi}{0}
\soulregister{\DeltaRepairLo}{0}
\soulregister{\DeltaTwo}{0}
\soulregister{\DeltaTwoEighteen}{0}
\soulregister{\DeltaTwoEighteenHi}{0}
\soulregister{\DeltaTwoEighteenInside}{0}
\soulregister{\DeltaTwoEighteenLo}{0}
\soulregister{\DeltaTwoHi}{0}
\soulregister{\DeltaTwoLo}{0}
\soulregister{\DeltaTwoP}{0}
\soulregister{\DeltaTwoPctHi}{0}
\soulregister{\DeltaTwoPctLo}{0}
\soulregister{\DeltaTwoSix}{0}
\soulregister{\DeltaTwoSixHi}{0}
\soulregister{\DeltaTwoSixLo}{0}
\soulregister{\DemoCont}{0}
\soulregister{\IQLR}{0}
\soulregister{\IQLstar}{0}
\soulregister{\MixBC}{0}
\soulregister{\OutcomeOne}{0}
\soulregister{\OutcomeTwo}{0}
\soulregister{\SRnorm}{0}
\soulregister{\SRrec}{0}
\soulregister{\TakeoverBC}{0}
\soulregister{\TmedianSteps}{0}
\soulregister{\WBC}{0}
\soulregister{\attPerSourceMax}{0}
\soulregister{\attPerSourceMean}{0}
\soulregister{\attPerSourceMedian}{0}
\soulregister{\baseCkptSha}{0}
\soulregister{\baseSRFamLOneCap}{0}
\soulregister{\baseSRFamLOneT}{0}
\soulregister{\baseSRLOneCap}{0}
\soulregister{\baseSRLOneT}{0}
\soulregister{\baseSRLOneTwoStep}{0}
\soulregister{\baseTrainHours}{0}
\soulregister{\baseTrainSteps}{0}
\soulregister{\budgetSec}{0}
\soulregister{\budgetSteps}{0}
\soulregister{\clipRecAWACPct}{0}
\soulregister{\clipRecIQLzeroPct}{0}
\soulregister{\cmark}{0}
\soulregister{\concurrent}{0}
\soulregister{\controlHz}{0}
\soulregister{\dSRnorm}{0}
\soulregister{\demoAboveCapRot}{0}
\soulregister{\demoAboveCapTrans}{0}
\soulregister{\demoDispPerFrame}{0}
\soulregister{\demoLenRatioMed}{0}
\soulregister{\demoPNNRot}{0}
\soulregister{\demoPNNTrans}{0}
\soulregister{\demoTransNormMed}{0}
\soulregister{\demoTransNormPninety}{0}
\soulregister{\demoTransNormPnn}{0}
\soulregister{\dnormAWAC}{0}
\soulregister{\dnormAbsAWAC}{0}
\soulregister{\dnormAbsDemo}{0}
\soulregister{\dnormAbsFirst}{0}
\soulregister{\dnormAbsHiAWAC}{0}
\soulregister{\dnormAbsHiDemo}{0}
\soulregister{\dnormAbsHiFirst}{0}
\soulregister{\dnormAbsHiIDQL}{0}
\soulregister{\dnormAbsHiIQL}{0}
\soulregister{\dnormAbsHiIQLzero}{0}
\soulregister{\dnormAbsHiMix}{0}
\soulregister{\dnormAbsHiTB}{0}
\soulregister{\dnormAbsIDQL}{0}
\soulregister{\dnormAbsIQL}{0}
\soulregister{\dnormAbsIQLzero}{0}
\soulregister{\dnormAbsLoAWAC}{0}
\soulregister{\dnormAbsLoDemo}{0}
\soulregister{\dnormAbsLoFirst}{0}
\soulregister{\dnormAbsLoIDQL}{0}
\soulregister{\dnormAbsLoIQL}{0}
\soulregister{\dnormAbsLoIQLzero}{0}
\soulregister{\dnormAbsLoMix}{0}
\soulregister{\dnormAbsLoTB}{0}
\soulregister{\dnormAbsMix}{0}
\soulregister{\dnormAbsTB}{0}
\soulregister{\dnormDemo}{0}
\soulregister{\dnormFirst}{0}
\soulregister{\dnormHiAWAC}{0}
\soulregister{\dnormHiDemo}{0}
\soulregister{\dnormHiFirst}{0}
\soulregister{\dnormHiIDQL}{0}
\soulregister{\dnormHiIQL}{0}
\soulregister{\dnormHiIQLzero}{0}
\soulregister{\dnormHiMix}{0}
\soulregister{\dnormHiTB}{0}
\soulregister{\dnormIDQL}{0}
\soulregister{\dnormIQL}{0}
\soulregister{\dnormIQLzero}{0}
\soulregister{\dnormLoAWAC}{0}
\soulregister{\dnormLoDemo}{0}
\soulregister{\dnormLoFirst}{0}
\soulregister{\dnormLoIDQL}{0}
\soulregister{\dnormLoIQL}{0}
\soulregister{\dnormLoIQLzero}{0}
\soulregister{\dnormLoMix}{0}
\soulregister{\dnormLoTB}{0}
\soulregister{\dnormMix}{0}
\soulregister{\dnormTAWAC}{0}
\soulregister{\dnormTB}{0}
\soulregister{\dnormTDemo}{0}
\soulregister{\dnormTFirst}{0}
\soulregister{\dnormTIDQL}{0}
\soulregister{\dnormTIQL}{0}
\soulregister{\dnormTIQLzero}{0}
\soulregister{\dnormTMix}{0}
\soulregister{\dnormTTB}{0}
\soulregister{\dnormTVsDemoAWAC}{0}
\soulregister{\dnormTVsDemoBase}{0}
\soulregister{\dnormTVsDemoFirst}{0}
\soulregister{\dnormTVsDemoIDQL}{0}
\soulregister{\dnormTVsDemoIQL}{0}
\soulregister{\dnormTVsDemoIQLzero}{0}
\soulregister{\dnormTVsDemoMix}{0}
\soulregister{\dnormTVsDemoTB}{0}
\soulregister{\dnormVsDemoAWAC}{0}
\soulregister{\dnormVsDemoAbsAWAC}{0}
\soulregister{\dnormVsDemoAbsBase}{0}
\soulregister{\dnormVsDemoAbsFirst}{0}
\soulregister{\dnormVsDemoAbsIDQL}{0}
\soulregister{\dnormVsDemoAbsIQL}{0}
\soulregister{\dnormVsDemoAbsIQLzero}{0}
\soulregister{\dnormVsDemoAbsMix}{0}
\soulregister{\dnormVsDemoAbsTB}{0}
\soulregister{\dnormVsDemoBase}{0}
\soulregister{\dnormVsDemoFirst}{0}
\soulregister{\dnormVsDemoHiAWAC}{0}
\soulregister{\dnormVsDemoHiBase}{0}
\soulregister{\dnormVsDemoHiFirst}{0}
\soulregister{\dnormVsDemoHiIDQL}{0}
\soulregister{\dnormVsDemoHiIQL}{0}
\soulregister{\dnormVsDemoHiIQLzero}{0}
\soulregister{\dnormVsDemoHiMix}{0}
\soulregister{\dnormVsDemoHiTB}{0}
\soulregister{\dnormVsDemoIDQL}{0}
\soulregister{\dnormVsDemoIQL}{0}
\soulregister{\dnormVsDemoIQLzero}{0}
\soulregister{\dnormVsDemoLoAWAC}{0}
\soulregister{\dnormVsDemoLoBase}{0}
\soulregister{\dnormVsDemoLoFirst}{0}
\soulregister{\dnormVsDemoLoIDQL}{0}
\soulregister{\dnormVsDemoLoIQL}{0}
\soulregister{\dnormVsDemoLoIQLzero}{0}
\soulregister{\dnormVsDemoLoMix}{0}
\soulregister{\dnormVsDemoLoTB}{0}
\soulregister{\dnormVsDemoMix}{0}
\soulregister{\dnormVsDemoTB}{0}
\soulregister{\effectThreshold}{0}
\soulregister{\essRecAWACPct}{0}
\soulregister{\essRecIQLPct}{0}
\soulregister{\essRecIQLzeroPct}{0}
\soulregister{\evalClientHours}{0}
\soulregister{\famNormZero}{0}
\soulregister{\faultSensMaxShift}{0}
\soulregister{\firstTBLongRate}{0}
\soulregister{\firstTBShortRate}{0}
\soulregister{\fpx}{0}
\soulregister{\fpxfulltitle}{0}
\soulregister{\fpxlogo}{0}
\soulregister{\fpxprojectlink}{0}
\soulregister{\framesPerFailure}{0}
\soulregister{\ftCollision}{0}
\soulregister{\ftDrawerOpen}{0}
\soulregister{\ftDropped}{0}
\soulregister{\ftMissedGrasp}{0}
\soulregister{\ftStuck}{0}
\soulregister{\ftWrongObject}{0}
\soulregister{\ftWrongPlacement}{0}
\soulregister{\horizonLong}{0}
\soulregister{\horizonMain}{0}
\soulregister{\horizonShort}{0}
\soulregister{\hoursAcceptedControl}{0}
\soulregister{\hoursAnnotation}{0}
\soulregister{\hoursControl}{0}
\soulregister{\hoursPreview}{0}
\soulregister{\humanSR}{0}
\soulregister{\humanSRLFive}{0}
\soulregister{\humanSRLFour}{0}
\soulregister{\humanSRLOne}{0}
\soulregister{\humanSRLThree}{0}
\soulregister{\humanSRLTwo}{0}
\soulregister{\humanSRmax}{0}
\soulregister{\humanSRmin}{0}
\soulregister{\humanSRphysFail}{0}
\soulregister{\humanSRstopVoid}{0}
\soulregister{\idqlBothMedStep}{0}
\soulregister{\idqlBothN}{0}
\soulregister{\idqlFasterPct}{0}
\soulregister{\idqlLostFaultN}{0}
\soulregister{\idqlLostLateN}{0}
\soulregister{\idqlLostN}{0}
\soulregister{\idqlLostTimeoutN}{0}
\soulregister{\idqlTBLongN}{0}
\soulregister{\idqlTBLongRate}{0}
\soulregister{\idqlTBShortN}{0}
\soulregister{\idqlTBShortRate}{0}
\soulregister{\intervalMax}{0}
\soulregister{\intervalMin}{0}
\soulregister{\liberox}{0}
\soulregister{\liberoxCommit}{0}
\soulregister{\liberoxDataRev}{0}
\soulregister{\liberoxPiLFive}{0}
\soulregister{\liberoxPiLOne}{0}
\soulregister{\lofoOneLoMin}{0}
\soulregister{\lofoOneMax}{0}
\soulregister{\lofoOneMin}{0}
\soulregister{\lofoTwoInsideFams}{0}
\soulregister{\lofoTwoMax}{0}
\soulregister{\lofoTwoMin}{0}
\soulregister{\meanSteps}{0}
\soulregister{\medNormSlowerTB}{0}
\soulregister{\medNormSlowerTBDemo}{0}
\soulregister{\medSecLFive}{0}
\soulregister{\medSecLFour}{0}
\soulregister{\medSecLOne}{0}
\soulregister{\medSecLThree}{0}
\soulregister{\medSecLTwo}{0}
\soulregister{\medianSec}{0}
\soulregister{\medianSteps}{0}
\soulregister{\mujocoVersion}{0}
\soulregister{\nAccLFive}{0}
\soulregister{\nAccLFour}{0}
\soulregister{\nAccLOne}{0}
\soulregister{\nAccLThree}{0}
\soulregister{\nAccLTwo}{0}
\soulregister{\nAccepted}{0}
\soulregister{\nAnnotatedRollouts}{0}
\soulregister{\nAnnotationRecords}{0}
\soulregister{\nAttemptsActive}{0}
\soulregister{\nAttemptsAll}{0}
\soulregister{\nAttemptsArchived}{0}
\soulregister{\nBaseAllLoPts}{0}
\soulregister{\nBaseNormFaultValid}{0}
\soulregister{\nBasePartLoPts}{0}
\soulregister{\nBaseZeroAllPts}{0}
\soulregister{\nBaseZeroLoPts}{0}
\soulregister{\nCandidates}{0}
\soulregister{\nConfigs}{0}
\soulregister{\nConfigsComplete}{0}
\soulregister{\nDemoFrames}{0}
\soulregister{\nDemos}{0}
\soulregister{\nDevAfterTest}{0}
\soulregister{\nDevRoots}{0}
\soulregister{\nDevStarts}{0}
\soulregister{\nDeviations}{0}
\soulregister{\nDiscardedSucc}{0}
\soulregister{\nEligChunksAll}{0}
\soulregister{\nEvalRows}{0}
\soulregister{\nExactlyTen}{0}
\soulregister{\nFailReleased}{0}
\soulregister{\nFailTest}{0}
\soulregister{\nFailTrain}{0}
\soulregister{\nFamNegOne}{0}
\soulregister{\nFamNegTwo}{0}
\soulregister{\nFamNormZero}{0}
\soulregister{\nFamPosOne}{0}
\soulregister{\nFamPosTwo}{0}
\soulregister{\nFamTieOne}{0}
\soulregister{\nFamTieTwo}{0}
\soulregister{\nFamilies}{0}
\soulregister{\nFamiliesKitchen}{0}
\soulregister{\nFamiliesStudy}{0}
\soulregister{\nFamilyFiveDisc}{0}
\soulregister{\nFaultAWAC}{0}
\soulregister{\nFaultBase}{0}
\soulregister{\nFaultDemo}{0}
\soulregister{\nFaultFirst}{0}
\soulregister{\nFaultIDQL}{0}
\soulregister{\nFaultIQL}{0}
\soulregister{\nFaultIQLzero}{0}
\soulregister{\nFaultMix}{0}
\soulregister{\nFaultTB}{0}
\soulregister{\nFlagUnionLo}{0}
\soulregister{\nFlagWidthLo}{0}
\soulregister{\nFwinChunksPost}{0}
\soulregister{\nGenFail}{0}
\soulregister{\nGenInvalidStart}{0}
\soulregister{\nGenNearMiss}{0}
\soulregister{\nGenPhysFault}{0}
\soulregister{\nGenStartInvalid}{0}
\soulregister{\nGenSuccFast}{0}
\soulregister{\nGenSuccSlow}{0}
\soulregister{\nKinRollouts}{0}
\soulregister{\nLOneTasks}{0}
\soulregister{\nLevels}{0}
\soulregister{\nLofoTwoInside}{0}
\soulregister{\nNominal}{0}
\soulregister{\nNormExcl}{0}
\soulregister{\nNormRowsAll}{0}
\soulregister{\nNormSens}{0}
\soulregister{\nNormSharedTB}{0}
\soulregister{\nNormSharedTBDemo}{0}
\soulregister{\nNormSlowerTB}{0}
\soulregister{\nNormSlowerTBDemo}{0}
\soulregister{\nNormValid}{0}
\soulregister{\nOddMidpoints}{0}
\soulregister{\nOverTwelveHundred}{0}
\soulregister{\nPhysics}{0}
\soulregister{\nPrefixFramesPost}{0}
\soulregister{\nRecChunksHi}{0}
\soulregister{\nRecChunksLo}{0}
\soulregister{\nRecChunksPost}{0}
\soulregister{\nRecRowsArm}{0}
\soulregister{\nRemediations}{0}
\soulregister{\nRetainedUnionLo}{0}
\soulregister{\nRolloutsGen}{0}
\soulregister{\nRolloutsGenTest}{0}
\soulregister{\nRolloutsGenTrain}{0}
\soulregister{\nSeeds}{0}
\soulregister{\nShipped}{0}
\soulregister{\nShippedTest}{0}
\soulregister{\nShippedTrain}{0}
\soulregister{\nSourcesActive}{0}
\soulregister{\nSourcesArchived}{0}
\soulregister{\nSourcesComplete}{0}
\soulregister{\nStepsAccepted}{0}
\soulregister{\nStops}{0}
\soulregister{\nSuccessAll}{0}
\soulregister{\nTUnmatched}{0}
\soulregister{\nTestConfirm}{0}
\soulregister{\nTestDiag}{0}
\soulregister{\nTestPerConfig}{0}
\soulregister{\nTestPerConfigShip}{0}
\soulregister{\nTimeLimit}{0}
\soulregister{\nTrainAccLThree}{0}
\soulregister{\nTrainAccPost}{0}
\soulregister{\nTrainPerConfig}{0}
\soulregister{\nTrainPerConfigShip}{0}
\soulregister{\nTrainSrcLThree}{0}
\soulregister{\nTrainSrcPost}{0}
\soulregister{\nTrainStepsLThree}{0}
\soulregister{\nTrainStepsPost}{0}
\soulregister{\nUndecided}{0}
\soulregister{\nUnsuccEpsPost}{0}
\soulregister{\nUnsuccFramesPost}{0}
\soulregister{\nUnsuccessful}{0}
\soulregister{\nWithheld}{0}
\soulregister{\nomFastNAWAC}{0}
\soulregister{\nomFastNBase}{0}
\soulregister{\nomFastNDemo}{0}
\soulregister{\nomFastNFirst}{0}
\soulregister{\nomFastNIDQL}{0}
\soulregister{\nomFastNIQL}{0}
\soulregister{\nomFastNIQLzero}{0}
\soulregister{\nomFastNMix}{0}
\soulregister{\nomFastNTB}{0}
\soulregister{\nomFaultPctAWAC}{0}
\soulregister{\nomFaultPctBase}{0}
\soulregister{\nomFaultPctDemo}{0}
\soulregister{\nomFaultPctFirst}{0}
\soulregister{\nomFaultPctIDQL}{0}
\soulregister{\nomFaultPctIQL}{0}
\soulregister{\nomFaultPctIQLzero}{0}
\soulregister{\nomFaultPctMix}{0}
\soulregister{\nomFaultPctTB}{0}
\soulregister{\nomFaultStableNAWAC}{0}
\soulregister{\nomFaultStableNBase}{0}
\soulregister{\nomFaultStableNDemo}{0}
\soulregister{\nomFaultStableNFirst}{0}
\soulregister{\nomFaultStableNIDQL}{0}
\soulregister{\nomFaultStableNIQL}{0}
\soulregister{\nomFaultStableNIQLzero}{0}
\soulregister{\nomFaultStableNMix}{0}
\soulregister{\nomFaultStableNTB}{0}
\soulregister{\nomLateNAWAC}{0}
\soulregister{\nomLateNBase}{0}
\soulregister{\nomLateNDemo}{0}
\soulregister{\nomLateNFirst}{0}
\soulregister{\nomLateNIDQL}{0}
\soulregister{\nomLateNIQL}{0}
\soulregister{\nomLateNIQLzero}{0}
\soulregister{\nomLateNMix}{0}
\soulregister{\nomLateNTB}{0}
\soulregister{\nomLatePctAWAC}{0}
\soulregister{\nomLatePctBase}{0}
\soulregister{\nomLatePctDemo}{0}
\soulregister{\nomLatePctFirst}{0}
\soulregister{\nomLatePctIDQL}{0}
\soulregister{\nomLatePctIQL}{0}
\soulregister{\nomLatePctIQLzero}{0}
\soulregister{\nomLatePctMix}{0}
\soulregister{\nomLatePctTB}{0}
\soulregister{\nomMedRatioTAWAC}{0}
\soulregister{\nomMedRatioTBase}{0}
\soulregister{\nomMedRatioTDemo}{0}
\soulregister{\nomMedRatioTFirst}{0}
\soulregister{\nomMedRatioTIDQL}{0}
\soulregister{\nomMedRatioTIQL}{0}
\soulregister{\nomMedRatioTIQLzero}{0}
\soulregister{\nomMedRatioTMix}{0}
\soulregister{\nomMedRatioTTB}{0}
\soulregister{\nomSensSRAWAC}{0}
\soulregister{\nomSensSRBase}{0}
\soulregister{\nomSensSRDemo}{0}
\soulregister{\nomSensSRFirst}{0}
\soulregister{\nomSensSRIDQL}{0}
\soulregister{\nomSensSRIQL}{0}
\soulregister{\nomSensSRIQLzero}{0}
\soulregister{\nomSensSRMix}{0}
\soulregister{\nomSensSRTB}{0}
\soulregister{\nomSuccNAWAC}{0}
\soulregister{\nomSuccNBase}{0}
\soulregister{\nomSuccNDemo}{0}
\soulregister{\nomSuccNFirst}{0}
\soulregister{\nomSuccNIDQL}{0}
\soulregister{\nomSuccNIQL}{0}
\soulregister{\nomSuccNIQLzero}{0}
\soulregister{\nomSuccNMix}{0}
\soulregister{\nomSuccNTB}{0}
\soulregister{\nomTimeoutPctAWAC}{0}
\soulregister{\nomTimeoutPctBase}{0}
\soulregister{\nomTimeoutPctDemo}{0}
\soulregister{\nomTimeoutPctFirst}{0}
\soulregister{\nomTimeoutPctIDQL}{0}
\soulregister{\nomTimeoutPctIQL}{0}
\soulregister{\nomTimeoutPctIQLzero}{0}
\soulregister{\nomTimeoutPctMix}{0}
\soulregister{\nomTimeoutPctTB}{0}
\soulregister{\normAbsAWAC}{0}
\soulregister{\normAbsBase}{0}
\soulregister{\normAbsDemo}{0}
\soulregister{\normAbsFirst}{0}
\soulregister{\normAbsIDQL}{0}
\soulregister{\normAbsIQL}{0}
\soulregister{\normAbsIQLzero}{0}
\soulregister{\normAbsMix}{0}
\soulregister{\normAbsTAWAC}{0}
\soulregister{\normAbsTB}{0}
\soulregister{\normAbsTBase}{0}
\soulregister{\normAbsTDemo}{0}
\soulregister{\normAbsTFirst}{0}
\soulregister{\normAbsTIDQL}{0}
\soulregister{\normAbsTIQL}{0}
\soulregister{\normAbsTIQLzero}{0}
\soulregister{\normAbsTMix}{0}
\soulregister{\normAbsTTB}{0}
\soulregister{\ns}{0}
\soulregister{\openpiCommit}{0}
\soulregister{\pctDiscardedOfSucc}{0}
\soulregister{\pctDrawFwinAWAC}{0}
\soulregister{\pctDrawFwinIQL}{0}
\soulregister{\pctDrawRecAWAC}{0}
\soulregister{\pctDrawRecIQL}{0}
\soulregister{\pctEligUnwFwin}{0}
\soulregister{\pctEligUnwRec}{0}
\soulregister{\pctRecChunksLo}{0}
\soulregister{\pctStopsOfUnsuccessful}{0}
\soulregister{\pctWithinSixHundred}{0}
\soulregister{\pctWithinTwelveHundred}{0}
\soulregister{\pctftCollision}{0}
\soulregister{\pctftDrawerOpen}{0}
\soulregister{\pctftDropped}{0}
\soulregister{\pctftGrounding}{0}
\soulregister{\pctftMissedGrasp}{0}
\soulregister{\pctftStuck}{0}
\soulregister{\pctftWrongObject}{0}
\soulregister{\pctftWrongPlacement}{0}
\soulregister{\pizf}{0}
\soulregister{\pmark}{0}
\soulregister{\pninetySecLFive}{0}
\soulregister{\pninetySecLFour}{0}
\soulregister{\pninetySecLOne}{0}
\soulregister{\pninetySecLThree}{0}
\soulregister{\pninetySecLTwo}{0}
\soulregister{\qOneUnionEst}{0}
\soulregister{\qOneUnionHi}{0}
\soulregister{\qOneUnionLo}{0}
\soulregister{\qTwoUnionEst}{0}
\soulregister{\qTwoUnionHi}{0}
\soulregister{\qTwoUnionLo}{0}
\soulregister{\recAllAWAC}{0}
\soulregister{\recAllBase}{0}
\soulregister{\recAllDemo}{0}
\soulregister{\recAllFirst}{0}
\soulregister{\recAllIDQL}{0}
\soulregister{\recAllIQL}{0}
\soulregister{\recAllIQLzero}{0}
\soulregister{\recAllMix}{0}
\soulregister{\recAllTB}{0}
\soulregister{\recHiAWAC}{0}
\soulregister{\recHiBase}{0}
\soulregister{\recHiDemo}{0}
\soulregister{\recHiFirst}{0}
\soulregister{\recHiIDQL}{0}
\soulregister{\recHiIQL}{0}
\soulregister{\recHiIQLzero}{0}
\soulregister{\recHiMix}{0}
\soulregister{\recHiTB}{0}
\soulregister{\recLoAWAC}{0}
\soulregister{\recLoBase}{0}
\soulregister{\recLoDemo}{0}
\soulregister{\recLoFirst}{0}
\soulregister{\recLoIDQL}{0}
\soulregister{\recLoIQL}{0}
\soulregister{\recLoIQLzero}{0}
\soulregister{\recLoMix}{0}
\soulregister{\recLoTB}{0}
\soulregister{\restoreMaxDiff}{0}
\soulregister{\robosuiteVersion}{0}
\soulregister{\rpGapBelowTwin}{0}
\soulregister{\rpHeadAgree}{0}
\soulregister{\rpMonoDown}{0}
\soulregister{\rpMonoUp}{0}
\soulregister{\rpQrangeRatio}{0}
\soulregister{\rpQueries}{0}
\soulregister{\rpRhoRotPrefix}{0}
\soulregister{\rpRhoTrans}{0}
\soulregister{\rpRhoTransPrefix}{0}
\soulregister{\rpRhoTransTrain}{0}
\soulregister{\rpSelGrip}{0}
\soulregister{\rpSelNN}{0}
\soulregister{\rpSelRot}{0}
\soulregister{\rpSelTrans}{0}
\soulregister{\rpSelTransHi}{0}
\soulregister{\rpSelTransLo}{0}
\soulregister{\rpTrainFrames}{0}
\soulregister{\rpTransGain}{0}
\soulregister{\samplesPerPoint}{0}
\soulregister{\secFiveFailureWeight}{0}
\soulregister{\secFiveFamMax}{0}
\soulregister{\secFiveFamMin}{0}
\soulregister{\secFiveMixGainMain}{0}
\soulregister{\secFiveMixGainOriginal}{0}
\soulregister{\secFiveNomHiDemo}{0}
\soulregister{\secFiveNomHiIQL}{0}
\soulregister{\secFiveNomHiMix}{0}
\soulregister{\secFiveNomHiTB}{0}
\soulregister{\secFiveNomLoDemo}{0}
\soulregister{\secFiveNomLoIQL}{0}
\soulregister{\secFiveNomLoMix}{0}
\soulregister{\secFiveNomLoTB}{0}
\soulregister{\secFiveRecoveryWeight}{0}
\soulregister{\seedTest}{0}
\soulregister{\seedTrain}{0}
\soulregister{\simDispPerStep}{0}
\soulregister{\srZeroAllTB}{0}
\soulregister{\srZeroLoDemo}{0}
\soulregister{\srZeroLoIQL}{0}
\soulregister{\srZeroLoTB}{0}
\soulregister{\suppAnnTest}{0}
\soulregister{\suppAnnTrain}{0}
\soulregister{\suppAnnUnd}{0}
\soulregister{\suppArchRel}{0}
\soulregister{\suppArchRevoked}{0}
\soulregister{\suppArchStop}{0}
\soulregister{\suppArchSucc}{0}
\soulregister{\suppArchTL}{0}
\soulregister{\suppAttemptCap}{0}
\soulregister{\suppCfgOverTestCap}{0}
\soulregister{\suppCfgOverTrainNominal}{0}
\soulregister{\suppConfigRows}{0}
\soulregister{\suppFamDiag}{0}
\soulregister{\suppFamNineLFourAtt}{0}
\soulregister{\suppFamNineLFourRel}{0}
\soulregister{\suppFunnelRows}{0}
\soulregister{\suppHSRFamMax}{0}
\soulregister{\suppHSRFamMin}{0}
\soulregister{\suppLateRel}{0}
\soulregister{\suppLateTest}{0}
\soulregister{\suppLateTestConf}{0}
\soulregister{\suppLateTrain}{0}
\soulregister{\suppNotAnn}{0}
\soulregister{\suppOutOfInterval}{0}
\soulregister{\suppOverCap}{0}
\soulregister{\suppPending}{0}
\soulregister{\suppShipNominalCfgTest}{0}
\soulregister{\suppShipNominalCfgTrain}{0}
\soulregister{\suppShipTestMax}{0}
\soulregister{\suppShipTestMin}{0}
\soulregister{\suppShipTrainMax}{0}
\soulregister{\suppShipTrainMin}{0}
\soulregister{\suppStopsAll}{0}
\soulregister{\suppTLAll}{0}
\soulregister{\suppUndFamNine}{0}
\soulregister{\suppUndTest}{0}
\soulregister{\suppUndTrain}{0}
\soulregister{\suppWhFailure}{0}
\soulregister{\suppWhNoFailure}{0}
\soulregister{\suppWhNoFailureLate}{0}
\soulregister{\suppWhTest}{0}
\soulregister{\suppWhTrain}{0}
\soulregister{\suppWhUncertain}{0}
\soulregister{\suppWidthFlag}{0}
\soulregister{\suppWidthFlagFamFive}{0}
\soulregister{\tabhead}{0}
\soulregister{\tabstyle}{0}
\soulregister{\tabzebra}{0}
\soulregister{\tbBothMedStep}{0}
\soulregister{\tbPredAboveRot}{0}
\soulregister{\tbPredAboveTrans}{0}
\soulregister{\tbPredPNNRot}{0}
\soulregister{\tbPredPNNTrans}{0}
\soulregister{\tbPredStepsAll}{0}
\soulregister{\tbPredTransNormMed}{0}
\soulregister{\tbPredTransNormPninety}{0}
\soulregister{\tbPredTransNormPnn}{0}
\soulregister{\timeGateCollect}{0}
\soulregister{\timeGateScreen}{0}
\soulregister{\tkTransNormMed}{0}
\soulregister{\tkTransNormPninety}{0}
\soulregister{\tkTransNormPnn}{0}
\soulregister{\vsTBAWAC}{0}
\soulregister{\vsTBAbsAWAC}{0}
\soulregister{\vsTBAbsBase}{0}
\soulregister{\vsTBAbsDemo}{0}
\soulregister{\vsTBAbsFirst}{0}
\soulregister{\vsTBAbsHiAWAC}{0}
\soulregister{\vsTBAbsHiBase}{0}
\soulregister{\vsTBAbsHiDemo}{0}
\soulregister{\vsTBAbsHiFirst}{0}
\soulregister{\vsTBAbsHiIDQL}{0}
\soulregister{\vsTBAbsHiIQL}{0}
\soulregister{\vsTBAbsHiIQLzero}{0}
\soulregister{\vsTBAbsHiMix}{0}
\soulregister{\vsTBAbsIDQL}{0}
\soulregister{\vsTBAbsIQL}{0}
\soulregister{\vsTBAbsIQLzero}{0}
\soulregister{\vsTBAbsLoAWAC}{0}
\soulregister{\vsTBAbsLoBase}{0}
\soulregister{\vsTBAbsLoDemo}{0}
\soulregister{\vsTBAbsLoFirst}{0}
\soulregister{\vsTBAbsLoIDQL}{0}
\soulregister{\vsTBAbsLoIQL}{0}
\soulregister{\vsTBAbsLoIQLzero}{0}
\soulregister{\vsTBAbsLoMix}{0}
\soulregister{\vsTBAbsMix}{0}
\soulregister{\vsTBBase}{0}
\soulregister{\vsTBDemo}{0}
\soulregister{\vsTBFirst}{0}
\soulregister{\vsTBHiAWAC}{0}
\soulregister{\vsTBHiBase}{0}
\soulregister{\vsTBHiDemo}{0}
\soulregister{\vsTBHiFirst}{0}
\soulregister{\vsTBHiIDQL}{0}
\soulregister{\vsTBHiIQL}{0}
\soulregister{\vsTBHiIQLzero}{0}
\soulregister{\vsTBHiMix}{0}
\soulregister{\vsTBIDQL}{0}
\soulregister{\vsTBIQL}{0}
\soulregister{\vsTBIQLzero}{0}
\soulregister{\vsTBLoAWAC}{0}
\soulregister{\vsTBLoBase}{0}
\soulregister{\vsTBLoDemo}{0}
\soulregister{\vsTBLoFirst}{0}
\soulregister{\vsTBLoIDQL}{0}
\soulregister{\vsTBLoIQL}{0}
\soulregister{\vsTBLoIQLzero}{0}
\soulregister{\vsTBLoMix}{0}
\soulregister{\vsTBMix}{0}
\soulregister{\wMeanFwinAWAC}{0}
\soulregister{\wMeanRecAWAC}{0}
\soulregister{\wMeanRecIQLzero}{0}
\soulregister{\xmark}{0}

%% file: supp/supp_preamble.tex
\ifdefined\pdfinfoomitdate \pdfinfoomitdate=1 \fi
\ifdefined\pdftrailerid \pdftrailerid{} \fi
\ifdefined\pdfsuppressptexinfo \pdfsuppressptexinfo=-1 \fi
\makeatletter
\def\cvprsect#1{\cvprsection[#1]{\texorpdfstring{\hskip -1em.~}{}#1}}
\def\cvprsubsect#1{\cvprsubsection[#1]{\texorpdfstring{\hskip -1em.~}{}#1}}
\makeatother
\usepackage{titletoc}
\titlecontents{ssection}[1.7em]{\addvspace{0.3em}\small\bfseries\filright\hyphenpenalty=10000\let\\\space}
  {\contentslabel{1.7em}}{\hspace*{-1.7em}}{\hfill\contentspage}
\titlecontents{ssubsection}[4.0em]{\small\filright\hyphenpenalty=10000\let\\\space}
  {\contentslabel{2.3em}}{\hspace*{-2.3em}}{\titlerule*[0.55pc]{.}\contentspage}
\contentsmargin{2.3em}
\newcommand{\supptoc}{%
  \startcontents[supp]%
  \begingroup\let\\\space
  \printcontents[supp]{s}{1}[2]{\section*{Contents}}%
  \endgroup\vspace{0.5em}}
\let\kitmaketitlesupp\maketitlesupplementary
\renewcommand{\maketitlesupplementary}{\kitmaketitlesupp\supptoc}

\makeatletter
\newcommand{\fpxsuppheadings}{%
  \def\cvprsection{\@startsection{section}{1}{\z@}{-10pt plus -2pt minus -2pt}{7pt}{\large\bf\raggedright\hyphenpenalty=10000}}%
  \def\cvprsubsection{\@startsection{subsection}{2}{\z@}{-8pt plus -2pt minus -2pt}{5pt}{\elvbf\raggedright\hyphenpenalty=10000}}}
\makeatother

\newcommand{\fpxsuppstart}{%
  \clearpage
  \setcounter{page}{1}\renewcommand{\thepage}{S\arabic{page}}%
  \appendix
  \fpxsuppheadings
  \raggedbottom\clubpenalty=10000\relax\csname @clubpenalty\endcsname=10000\relax\widowpenalty=10000\relax
  \setcounter{table}{0}\setcounter{figure}{0}\setcounter{equation}{0}%
  \counterwithin*{table}{section}\counterwithin*{figure}{section}\counterwithin*{equation}{section}%
  \renewcommand{\thetable}{\thesection\arabic{table}}%
  \renewcommand{\thefigure}{\thesection\arabic{figure}}%
  \renewcommand{\theequation}{\thesection\arabic{equation}}%
  \renewcommand{\theHsection}{supp.\Alph{section}}%
  \renewcommand{\theHsubsection}{\theHsection.\arabic{subsection}}%
  \renewcommand{\theHsubsubsection}{\theHsubsection.\arabic{subsubsection}}%
  \renewcommand{\theHtable}{supp.\thesection.\arabic{table}}%
  \renewcommand{\theHfigure}{supp.\thesection.\arabic{figure}}%
  \renewcommand{\theHequation}{supp.\thesection.\arabic{equation}}%
  \crefalias{appendix}{section}\crefalias{subappendix}{section}\crefalias{subsubappendix}{section}%
}

%% file: figures/fig1_teaser.tex
\begin{center}
\includegraphics[width=\linewidth]{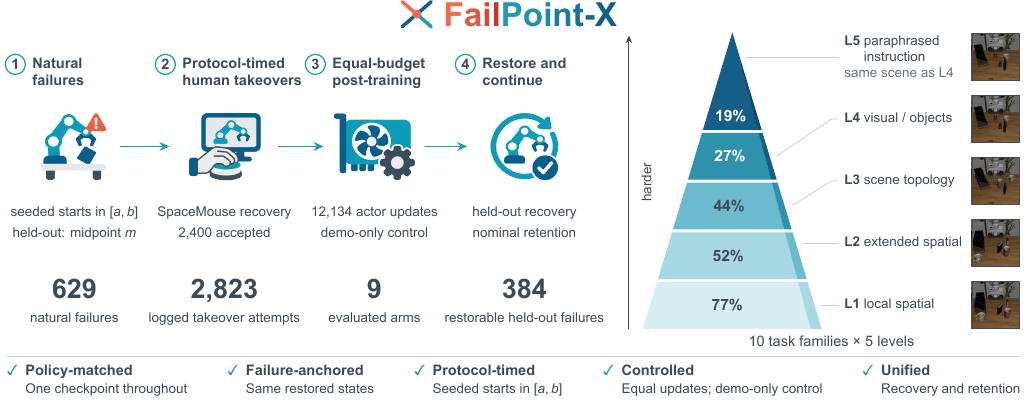}
\vspace{-2mm}
\captionsetup{hypcap=false}\captionof{figure}{\textbf{\fpx{} benchmarks offline VLA post-training on a policy's own failures.} One \pizf{} checkpoint fails in \nFamilies{} \liberox{} task families at five levels (percentages: its success within \horizonMain{} steps); humans take over from seeded starts in annotated training-failure intervals; trained arms share one actor-update budget, and all arms restart from the same held-out failures and nominal starts.}
\fpxFigureGuard{1}\label{fig:teaser}
\end{center}
\vspace{-2mm}

%% file: sec/0_abstract.tex
\begin{abstract}
Human takeovers offer corrective supervision for vision-language-action (VLA) policies, but their contribution to offline post-training must be separated from that of continued optimization. We introduce \fpx, a benchmark for offline VLA post-training with human takeovers. To represent uncertainty in failure onset, an annotator marks an interval for each failure, and training takeovers start at seeded frames sampled within it. From one fixed \pizf checkpoint, which also initializes post-training, we assemble \nFailReleased natural failures across \nFamilies \liberox task families at up to \nLevels difficulty levels. All \nAttemptsAll takeover attempts are logged; \nAccepted are accepted recoveries, of which \nTrainAccPost are used for post-training. Evaluation restarts every method from the fixed midpoints of \nFailTest held-out failure intervals. Under matched actor-update counts, we compare behavior cloning of recoveries, value-based offline learning on recoveries and failed rollouts, and continued training on existing demonstrations. Within \horizonMain steps on the \nTestConfirm L1--L3 held-out failures, cloning improves recovery by \DeltaOne percentage points (pp) over the demonstration-only control; the value-based learner does not establish a further gain (\DeltaTwo pp). In exploratory evaluation from L1--L5 nominal starts at the same horizon, cloning loses \dnormAbsTB pp of success relative to the base policy; mixing in demonstrations reduces the loss to \dnormAbsMix pp, a reduction that shrinks under a stricter, pre-specified horizon. Restorable failure states and a shared evaluator let all methods be compared from identical starts.
\end{abstract}

%% file: sec/1_intro.tex
\section{Introduction}

\label{sec:intro}

Vision-language-action (VLA) policies map visual observations and language instructions to robot control actions. Embodied agents more broadly must decide and act reliably under partial observation, from judging whether a target is present to searching unseen environments for it~\cite{lewis2026safevantage,ji2026sornav,cai2026intentnav}. Reliable manipulation, however, requires handling states induced by the robot's own errors, including missed grasps, misplaced objects, and interrupted task progress. At such states, a human takeover can supply corrective actions that restore progress toward the original goal. Offline post-training seeks to make such recovery autonomous, learning from human corrections or from the robot's own mixed-quality deployment experience~\cite{ren2026mixedquality}; we study it on human recoveries recorded at a fixed policy's own failures. Training uses failed policy executions and human takeovers from their intermediate states; the resulting policy must autonomously complete the original task from held-out failure states of the same task configurations.

The contribution of takeover data is difficult to isolate from overall task success alone. From a nominal start, a higher success rate may reflect either avoiding an error or recovering from one. Moreover, post-training adds optimization steps, so comparing a post-trained policy directly with the base policy conflates the effect of new data with that of continued optimization. LIBERO-RECOVER~\cite{liu2026liberorecover} reports recovery gains from fine-tuning on recoveries mixed with demonstrations, without a demonstration-only control matched in training recipe and optimizer steps. RePO-VLA~\cite{liufu2026repovla} includes a matched-budget ablation of value conditioning on the real-robot Fold Towel task and evaluates recovery under injected perturbations, and SimPT~\cite{wang2026simpt} compares post-training methods at matched policy-update counts on held-out environment seeds; neither reports recovery from the base policy's own held-out failures against a matched demonstration-only control. We therefore ask two questions about using recovery data for offline post-training. \textbf{Q1:} At matched actor updates, does training on human recoveries improve failure recovery more than continued training on existing demonstrations? Beyond the successful corrections, the preceding failed executions provide an additional negative training signal. \textbf{Q2:} Does value-based post-training on both recoveries and failed rollouts improve upon behavior cloning of recoveries alone?

Collecting recovery data also requires deciding when to intervene. Selecting a single frame, whether by a model or a human, commits data collection to a single temporal judgment, yet perceived event boundaries depend on physical changes, interpreted goals, and temporal granularity~\cite{kurby2008segmentation}. We therefore let an annotator bracket the failure onset with an interval $[a,b]$ and start training takeovers from seeded frames sampled within it.

We introduce \fpx, a policy-matched benchmark for offline VLA post-training with human takeovers: one checkpoint both generates the failures and initializes post-training (\cref{fig:teaser}, Table~\ref{tab:related}). Because simulation can restore exactly the same recorded physical state across policies, the benchmark's failure-anchored protocol (\cref{sec:protocol}) restarts every evaluated policy (arm) from the interval midpoints of the same \nFailTest{} held-out failures and, separately, from nominal starts to measure retained task performance. The unchanged base policy (\BaseRestart) provides a reference. \DemoCont{} continues training on existing demonstrations for as many actor updates as the takeover-trained arms: \TakeoverBC{} clones accepted recoveries, \IQLR{} adds failed training rollouts and a learned critic, and \MixBC{} mixes recoveries with demonstrations. Q1 and Q2, paired over task families, are confirmatory on the \nTestConfirm{} held-out L1--L3 failures of all \nFamilies{} families; \MixBC, the \nTestDiag{} L4--L5 failures, and nominal starts are exploratory.

Our collection pipeline (\cref{sec:dataset}, \cref{fig:pipeline}) supplies recovery supervision from restorable failure states and logs its human cost. From rollouts of a fixed \pizf~\cite{black2025pi05} checkpoint, we assemble \nFailReleased{} natural failures across \nFamilies{} \liberox~\cite{wang2026liberox} task families at up to \nLevels{} difficulty levels, split by seed into training and held-out sets. Takeovers are protocol-timed: the protocol, not the operator, draws \framesPerFailure{} seeded start frames per training failure, and unsuccessful attempts are retried at the same frame. We retain all \nAttemptsAll{} attempts, including \nAccepted{} accepted recoveries, of which \nTrainAccPost{} are used for post-training.

Within a \horizonMain-step horizon on L1--L3, \TakeoverBC{} improves recovery by \DeltaOne{} percentage points (pp) over \DemoCont, whereas \IQLR{} does not establish a further gain over \TakeoverBC{} (\DeltaTwo{} pp; \cref{fig:results}a). In exploratory evaluations from L1--L5 nominal starts at the same horizon, \TakeoverBC{} loses \dnormAbsTB{} pp of nominal success relative to the base policy; mixing in demonstrations reduces this loss to \dnormAbsMix{} pp but lowers recovery (\cref{fig:results}b). This smaller loss depends on the evaluation horizon: under the stricter, pre-specified {$1.1T$} horizon ($T$: mean demonstration length), \TakeoverBC{} and \MixBC{} change nominal success by similar amounts (\dnormTTB{} and \dnormTMix~pp; \cref{sec:experiments}). Thus, takeover data improve recovery in this study, while nominal-task evaluation reveals an accompanying retention loss.

Our contributions are:

\begin{enumerate}[leftmargin=*,itemsep=1pt,parsep=0pt,topsep=2pt]

\item \textbf{\fpx, a policy-matched benchmark for offline VLA post-training with human takeovers}, combining restorable failure states from one base checkpoint (reusable across policies, policy-matched when initialized from it), fixed splits, matched actor updates, and joint recovery and nominal-task evaluation (\cref{sec:benchmark,sec:protocol}).

\item \textbf{Interval-based takeover collection and a failure-and-recovery dataset} with human-annotated failure intervals, seeded takeover starts within each training interval, and complete attempt logs of human collection cost (\cref{sec:dataset}).

\item \textbf{A controlled empirical study of recovery benefit and nominal-task loss for a fixed checkpoint}, with pre-specified L1--L3 comparisons of behavior cloning against continued demonstration training (Q1) and value-based learning (Q2), and exploratory analyses of \MixBC, L4--L5, nominal starts, task families, and additional training and inference variants (\cref{sec:experiments}).

\end{enumerate}

%% file: sec/2_related.tex
\input{tables/tab1_related}

\section{Related Work}

\label{sec:related}

\paragraph{Learning from human interventions.}

Human interventions supervise the states a learner actually visits, which independently collected demonstrations may cover poorly. DAgger aggregates expert labels at learner-visited states~\cite{ross2011dagger}, and HG-DAgger lets the human decide when to intervene and return autonomy, yielding contiguous corrective segments~\cite{kelly2019hgdagger}. Sirius weights original demonstrations, autonomous rollouts, human interventions, and pre-intervention segments within an iterative deployment-and-training loop~\cite{liu2023sirius}. Extending human corrections to VLA reinforcement learning, ConRFT combines offline initialization on demonstrations with online intervention-guided learning~\cite{chen2025conrft}. \fpx{} instead fixes one round of protocol-timed human recoveries from a single base policy's failures and compares ways to use them for offline post-training.

\input{figures/fig2_pipeline}
\paragraph{Offline learning from corrections and failures.}

While corrective trajectories supply direct action targets for imitation, autonomous rollouts additionally carry information about action outcomes. Prior work learns from such data with both imitation and value-based objectives: Sirius evaluates behavior cloning (BC) and Implicit Q-Learning (IQL) on deployment logs~\cite{liu2023sirius}, and RECAP integrates demonstrations, autonomous experience, and human corrections via value estimation and advantage-conditioned policy training~\cite{pi2025pistar06}. Focusing on failure recovery, RePO-VLA imitates corrective suffixes and adds progress-aligned value supervision to learn from successful, failed, and corrective trajectories; it includes a non-value control on the same data and budget and an offline RECAP baseline matched in training steps and recovery-data budget~\cite{liufu2026repovla}. Similarly, SimPT benchmarks HG-DAgger, RECAP, and preference-based objectives on fixed datasets of policy rollouts and simulated expert corrections generated from per-task $\pi_{0.5}$ parent policies, matching policy-training steps and evaluating from held-out initial states~\cite{wang2026simpt}.

We compare such objectives under controlled data and evaluation conditions. Our value-based arm adapts the advantage weighting of IQL~\cite{kostrikov2022iql} to the VLA's flow-matching objective; contrasting it with recovery-only behavior cloning evaluates the coupled choice of data composition and learning objective.

\paragraph{Benchmarks for failure recovery.}

Evaluating failure recovery requires specifying the intermediate state from which a policy must resume to complete the original task. \fpx{} builds on the task configurations, demonstrations, and difficulty levels of LIBERO-X~\cite{wang2026liberox}. Among recent benchmarks, LIBERO-RECOVER constructs scenarios from naturally occurring failures across multiple policies, evaluates recovery after restoring and locally perturbing recorded states, and collects human recovery demonstrations; it also reports supervised fine-tuning on mixtures of recoveries and original demonstrations, including nominal-start performance~\cite{liu2026liberorecover}. RoboRecover replays action prefixes to reconstruct deviation states from which diverse policies resume, and studies state reversion with short learned corrections~\cite{li2026roborecover}.

\fpx{} compares post-training choices for a single fixed checkpoint: the failure-generating policy and the post-training initialization are identical, all methods resume from the same held-out failure states, a demonstration-only control matches the number of actor updates, and nominal starts measure task retention (Table~\ref{tab:related}).

\paragraph{Localizing failures and selecting takeover states.}

Recovery starting states can be selected based on task progress, estimated recoverability, or human judgment regarding when a failure began. SimPT tracks progress with a task-stage graph on privileged simulator state and attempts planner-based recovery from saved snapshots~\cite{wang2026simpt}. Kintsugi-VLA estimates a privileged expert's recovery probability along failed trajectories, selecting states within a window bounded by deviation onset and an expert-relative recovery frontier~\cite{snegirev2026kintsugi}. LIBERO-RECOVER instead prompts a vision-language model to pinpoint the earliest task-relevant failure timestamp~\cite{liu2026liberorecover}.

In \fpx, the annotated onset interval expresses uncertainty about when the failure began, not an expert's recoverability window, and seeded sampling within it decouples onset annotation from the selected takeover starts.

%% file: tables/tab1_related.tex
\begin{table*}[t]
\centering
\caption{\textbf{Positioning against related resources.} \cmark/\xmark: present/absent; \pmark: partial; \ns: not stated in the source; --: not applicable; \concurrent{}:~concurrent (earliest public version on or after 2026-09-15). Several works are larger, use more operators or source policies, or include real-robot data.}
\label{tab:related}
\scriptsize
\setlength{\tabcolsep}{2.4pt}
\renewcommand{\arraystretch}{1.08}
\begin{tabular*}{\textwidth}{@{\extracolsep{\fill}}l c c c c c c c c c c c@{}}
\toprule
Resource & Platform & \makecell[b]{Failure source\\(policies)} & \makecell[b]{Same\\ckpt.} & \makecell[b]{Recovery\\supervision\\(operators)} & \makecell[b]{State\\reconstr.} & \makecell[b]{Starts\\per\\failure} & \makecell[b]{Failed\\attempts\\kept} & \makecell[b]{Eval.\ at\\held-out\\failures} & \makecell[b]{Nominal\\check} & \makecell[b]{Extra-\\training\\control} & \makecell[b]{Failures /\\recoveries} \\
\midrule
LIBERO-X~\cite{wang2026liberox} & Sim & None & -- & None & -- & -- & -- & \xmark & -- & -- & -- \\
LIBERO-RECOVER~\cite{liu2026liberorecover} & Sim & Pooled (6) & \xmark & Human (4) & Restore & 1 & \ns & \cmark & \cmark & \xmark & 2,178 / 3,184$^{e}$ \\
RoboRecover\concurrent~\cite{li2026roborecover} & Sim & Mixed (4+4) & \xmark & Other policies & Replay & 1 & \ns & \cmark & \xmark & \xmark & 2,000 / \ns \\
SimPT\concurrent~\cite{wang2026simpt} & Sim+Real & Own (1/task) & \pmark$^{c}$ & Expert+human & Restore & \ns & \ns & \xmark & \cmark & \ns & \ns{} / \ns \\
Kintsugi-VLA\concurrent~\cite{snegirev2026kintsugi} & Sim & Pooled (2) & \xmark & Expert & Restore & 3 & \ns & \cmark & \cmark & \ns & 200 / \ns \\
REBOOT\concurrent~\cite{ofori2026reboot} & Real & Injected & -- & Human (1) & None & 1 & \ns & \xmark & -- & -- & 1,080 / 1,080 \\
RePO-VLA/FRBench~\cite{liufu2026repovla} & Sim+Real & Mixed & \xmark & Expert+human & None & \ns & \ns & \pmark$^{d}$ & \cmark & \ns & \ns{} / 6,392$^{f}$ \\
AgiBot World 2026 RL~\cite{agibot2026worldrl} & Real & Natural & -- & Human & None & \ns & \ns & \xmark & -- & -- & 1,369 / 10,684$^{g}$ \\
robomimic~\cite{mandlekar2021robomimic} & Sim+Real & -- & -- & None & -- & -- & -- & \xmark & -- & -- & -- \\
\midrule
\textbf{\fpx{} (ours)} & Sim & Own (1) & \cmark & Human (1) & Restore$^{a}$ & \framesPerFailure{} & \cmark$^{b}$ & \cmark & \cmark & \cmark & \nFailReleased{} / \nAccepted{} \\
\bottomrule
\end{tabular*}
\par\vspace{3pt}
\begin{minipage}{\textwidth}
\scriptsize
Same ckpt.: training failures, held-out failures, and post-training share one checkpoint. Own/Pooled/Natural: natural failures of the post-trained, several, or an unnamed policy; Injected: injected or perturbed; Mixed: both. Expert: privileged planner or RL teacher. Starts: distinct recovery starts. Nominal check: initial-state success after post-training. Extra-training control: equal update budget, no recovery data. Last column: failures / recovery trajectories as reported (ours: training and held-out failures / accepted recoveries). $^{a}$Exact MuJoCo-state equality verified at every restart from a held-out failure; open-loop replay of recorded takeovers not run. $^{b}$All \nAttemptsAll{} attempts are logged, failed and discarded ones included. $^{c}$Matched for training failures; evaluated from initial states. $^{d}$Errors injected during evaluation. $^{e}$Trajectories cover 413 scenarios. $^{f}$Simulation split. $^{g}$First-release failed rollouts / intervention segments.
\end{minipage}
\end{table*}

%% file: figures/fig2_pipeline.tex
\begin{figure*}[t]
\centering
\includegraphics[width=\linewidth]{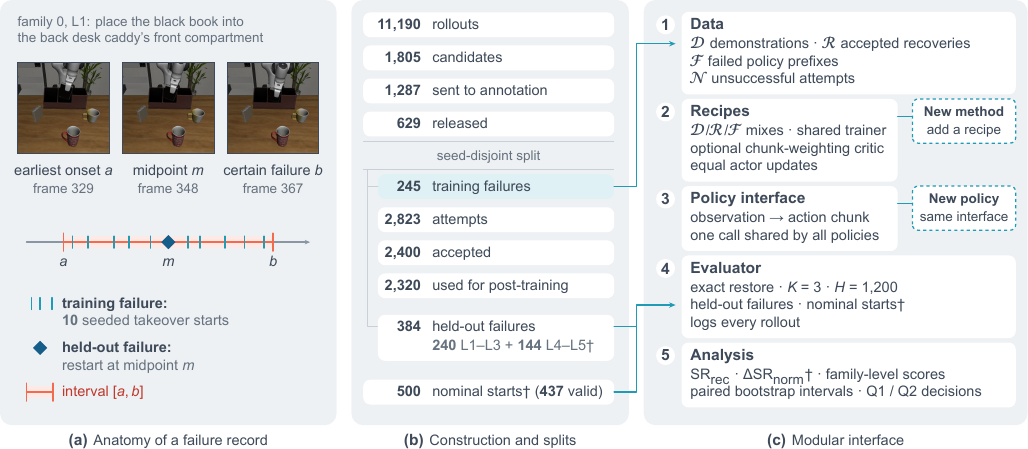}
\vspace{-2mm}
\caption{\textbf{Benchmark construction and interface.} (a)~One recorded failure: the annotated interval $[a,b]$ brackets the failure onset; a training failure receives \framesPerFailure{} seeded takeover starts inside it, and every arm restarts a held-out failure from its midpoint $m$. (b)~Screened natural failures of the fixed \pizf{} checkpoint are released into seed-disjoint splits fixed before collection; takeovers are collected on the training side (\nSourcesComplete{} of \nFailTrain{} complete). (c)~Data, post-training recipes, one policy interface, the restore-and-continue evaluator, and the analysis communicate through fixed formats, so a new method changes only its recipe or the policy served behind the interface ($K$: policy-sampling seeds per failure; $H$: horizon in control steps; $\dagger$:~exploratory).}
\fpxFigureGuard{2}\label{fig:pipeline}
\end{figure*}

%% file: sec/4_dataset.tex
\section{The \fpx{} Benchmark}
\label{sec:benchmark}
\label{sec:dataset}

\fpx{} connects failures generated by a fixed policy, human takeovers started inside annotated failure intervals, and held-out states for autonomous recovery evaluation (\cref{fig:pipeline}).

\subsection{Task Definition and Benchmark Scope}
\label{sec:benchmark_scope}

The task is to improve autonomous recovery through offline post-training. Methods start from one fixed base checkpoint and may use the original demonstrations, training-side failed executions, and logged human takeover attempts; held-out failures, seeds, and annotations are reserved for evaluation. At a restored held-out state, the resulting policy receives images, proprioception, and the original instruction and must complete the task without human assistance. Training and held-out failures come from disjoint rollout seeds of the same task configurations, so the benchmark measures recovery on new executions; task retention is evaluated separately from nominal starts.

One $\pi_{0.5}$ checkpoint~\cite{black2025pi05}, which we fine-tuned with openpi~\cite{openpi} on the \nDemos{} LIBERO-X~\cite{wang2026liberox} demonstrations, both generates the failures and initializes post-training. We use 10 task families (three kitchen, seven study; IDs in \cref{fig:families}). A family groups a task across five levels, and a configuration (50 in total) denotes one family at one level. These levels progressively introduce local spatial perturbations (L1), extended spatial perturbations (L2), scene-topology changes (L3), visual-attribute changes and unseen or confounding objects (L4), and semantically equivalent instruction paraphrases (L5); in family 14, LIBERO-X defines identical L3 and L4 tasks.

\subsection{Natural Failures and Restorable States}
\label{sec:benchmark_states}

To obtain states arising from the policy's own behavior, we execute the fixed checkpoint without human intervention, with rollouts assigned to separate training and held-out seed blocks before collection. Rollouts that have not completed the task by $1.5T$ count as failures, where $T$ is the mean demonstration length for the family's L1 task, counted in 20 Hz control steps and shared across its five levels; failure is thus defined relative to this rollout cutoff. Screening then excludes invalid initial states, physics faults, and near-misses, leaving \nCandidates{} candidates (\cref{fig:pipeline}b). Candidates are sent to annotation largely in seed order, and the per-configuration target is five training and eight held-out failures, so the curated set is a screened, quota-based subset (rules and exceptions in the supplement).

We retain observations, action histories, simulator states, and fixture layouts. Restoration re-applies the recorded layout and simulator state and reconstructs the gripper-controller state from preceding actions, so human takeovers on training failures and all evaluated policies on held-out failures start from exactly the recorded states.

\subsection{Interval-Based Human Takeover Collection}
\label{sec:benchmark_collection}

Failure onset can be ambiguous even when its consequence is clear. To retain this uncertainty, an annotator marks an interval $[a,b]$, targeting 10--40 frames, around a selected main failure event: $a$ is the earliest frame at which the event may have begun, and by frame $b$ it has certainly occurred. A coordinator releases or withholds annotated candidates and finalizes the interval and failure-type label of each released failure. Held-out evaluation uses these finalized intervals. One person performed all annotation and teleoperation, so inter-annotator agreement is not estimated.

For each released training failure selected for collection, the protocol samples 10 frames uniformly without replacement from the finalized $[a,b]$ with a recorded random seed; all training intervals contain at least 10 frames. The operator then starts from the restored state at each sampled frame and supplies corrective actions with a SpaceMouse at 20~Hz, within \budgetSteps{} control steps per attempt.

Collection targets one accepted recovery per sampled start: unsuccessful attempts and operator-discarded successes are retried at the same frame, and acceptance requires a submitted, successful trajectory that passes automatic integrity checks. All attempts, including operator stops, time limits, physics faults, and discarded successes, remain in the log, as do curation decisions such as six start frames redrawn after physics faults (supplement).

\subsection{Dataset Composition and Collection Cost}
\label{sec:benchmark_data}
\input{tables/tab2_stats}

The dataset (Table~\ref{tab:stats}) contains \nFailReleased{} released failures: \nFailTrain{} on the training side and \nFailTest{} held out. Collection is complete for \nSourcesComplete{} training failures in \nConfigsComplete{} configurations (\nAccepted{} accepted recoveries). Five released training failures lack a complete takeover set and are not used: \suppArchRel{} sources archived during collection (each replaced by the next released seed) and the \suppFamNineLFourRel{} family-9 L4 failures, whose collection stopped after \suppFamNineLFourAtt{} attempts at one start reached the time limit (family~9 has no accepted L4--L5 recoveries). Held out are \nTestConfirm{} L1--L3 failures in all 10 families and \nTestDiag{} L4--L5 failures in nine. Seven failure types cover wrong placements, wrong objects, collisions, missed grasps, sustained lack of progress, dropped objects, and unclosed drawers.

Eight complete training failures (the fifth in seed order in eight configurations of two families) form a development set for implementation checks, keeping each source's takeovers together. Their 80 recoveries are excluded from post-training, which uses \nTrainSrcPost{} source failures and \nTrainAccPost{} accepted recoveries from all five levels.

The full log contains \nAttemptsAll{} attempts, including those on archived sources; annotation took \hoursAnnotation{} hours and active takeover control \hoursControl{} hours, and other collection time is not counted. Per-attempt human success varies little across levels, whereas the median accepted recovery grows longer from L1 to L5 (Table~\ref{tab:stats}): takeovers at harder levels are longer rather than less successful.

\subsection{Modular Benchmark Interface}
\label{sec:benchmark_interface}
\fpx{} separates its components behind fixed interfaces (\cref{fig:pipeline}c): datasets with fixed splits; post-training recipes initialized from $\pi_{\mathrm{base}}$ within the shared actor budget (preserving policy matching; policies from other bases reuse the states without policy matching); one shared chunk-inference call; an evaluator that restores held-out states and logs rollouts; and a shared analysis for \SRrec, \dSRnorm{}, Q1, and Q2 (\cref{sec:protocol}). Datasets, states, evaluators, recipes, and checkpoints will be released under CC~BY~4.0 and MIT licenses (supplement).

%% file: tables/tab2_stats.tex
\begin{table}[t]
\centering
\caption{\textbf{Dataset composition by level.} A complete configuration has \nTrainPerConfig{} training sources with \framesPerFailure{} accepted recoveries each and \nTestPerConfig{} held-out failures. Takeover sources: released training failures not archived after release (in parentheses: complete). $^{a}$On the \nSourcesActive{} takeover sources (\nAttemptsAll{} including archived ones). $^{b}$Per attempt; discarded successes count as successes, stops and time limits as failures; physics faults excluded. $^{c}$Nearest-rank.}
\label{tab:stats}
\footnotesize\setlength{\tabcolsep}{3pt}%
\newcolumntype{L}{>{\raggedright\arraybackslash\hangindent=0.5em\hangafter=1}p{95pt}}%
\newcolumntype{N}[1]{>{\centering\arraybackslash\spaceskip=0.2em plus 0.05em\relax}p{#1}}%
\resizebox{\columnwidth}{!}{%
\begin{tabular}{L *{5}{N{24pt}} N{32pt}}
\toprule
Level & L1 & L2 & L3 & L4 & L5 & Total \\
\midrule
Complete configurations & 10 & 10 & 10 & 9 & 9 & 48 \\
Released failures (train / held-out) & 50 / 80 & 52 / 80 & 50 / 80 & 47 / 72 & 46 / 72 & 245 / 384 \\
Takeover sources (complete) & 50 (50) & 50 (50) & 50 (50) & 47 (45) & 45 (45) & 242 (240) \\
Attempts & 589 & 592 & 576 & 511 & 531 & 2,799$^{a}$ \\
\quad accepted & 500 & 500 & 500 & 450 & 450 & 2,400 \\
\quad discarded success & 21 & 10 & 17 & 9 & 14 & 71 \\
\quad operator stop & 68 & 81 & 57 & 42 & 61 & 309 \\
\quad time limit / physics fault & 0 / 0 & 0 / 1 & 1 / 1 & 3 / 7 & 0 / 6 & 4 / 15 \\
Human success (\%)$^{b}$ & 88.5 & 86.3 & 89.9 & 91.1 & 88.4 & 88.8 \\
Accepted length, median [p90] (s)$^{c}$ & 7.0 [29.7] & 9.9 [28.9] & 17.2 [42.8] & 18.1 [40.3] & 19.2 [39.0] & 15.9 [37.0] \\
Accepted control steps (k) & 121.3 & 131.5 & 188.3 & 191.9 & 195.8 & 828.8 \\
\bottomrule
\end{tabular}}
\end{table}

%% file: sec/3_protocol.tex
\section{Controlled Post-Training and Evaluation}
\label{sec:protocol}
\input{tables/tab_report_card}

We compare offline post-training methods initialized from the failure-generating checkpoint, using the training split and development exclusions in Sec.~\ref{sec:dataset}.

\subsection{Post-Training Methods and Controls}
\label{sec:posttrain_methods}

\paragraph{Shared training and controls.}
Table~\ref{tab:reportcard} lists every arm with its data and results. Let $\mathcal D$, $\mathcal R$, $\mathcal F$, and $\mathcal N$ denote the original demonstrations, accepted recoveries, failed policy prefixes, and unsuccessful human attempts (operator stops or time limits) on post-training sources. \TakeoverBC{} clones $\mathcal R$ and \DemoCont{} trains on $\mathcal D$ with the same recipe, comparing the collected recovery data with continued optimization. All trained arms freeze the vision-language backbone and action-expert base weights, updating LoRA adapters and action/time projections. They share normalization statistics, optimizer, global batch size 64, and $U=12{,}134$ actor updates. Actors condition on images and language instructions but not on the available proprioception, and optimize openpi's flow-matching loss~\cite{openpi}. \mbox{Mix-BC} selects demonstrations or recoveries with equal probability, then samples uniformly within that source. Four exploratory variants fit the IQL critic and weight actor chunks on recoveries alone (\mbox{IQL-R0}), change the weighting (AWAC~\cite{nair2020awac}), or select among 32 chunks sampled from \TakeoverBC{} at inference: the first (\mbox{First-of-32}) or the \IQLR{} critic's highest-valued one (\mbox{IDQL-R2}~\cite{hansenestruch2023idql}).

\paragraph{Value-based offline learning.}
\mbox{IQL-R2} fits four-layer, 512-unit MLP critics~\cite{kostrikov2022iql} ($Q_1,Q_2,V$; Adam $3\times10^{-4}$, batch 512, Polyak $0.005$) for 50,000 updates on frozen vision-language features and proprioception, with expectile regression ($\tau=0.7$) and twin-$Q$ TD learning ($\gamma=0.99$ per chunk; AWAC averages over $K=8$ cached \TakeoverBC{} chunks). Recovery endpoints get reward $+1$ and failed-prefix endpoints $-1$; $\mathcal N$ enters the critic only as zero-reward truncations, and physics faults are excluded. Over eligible full 10-action chunks $\mathcal E$ (\nRecChunksPost{} from $\mathcal R$ and \nFwinChunksPost{} from $\mathcal F_{\mathrm{win}}$, frames $\max(0,a-50)$ to $m-10$), with $A_i=\min(Q_1,Q_2)(c_i,\mathbf a_i)-V(c_i)$ and $\sigma_A=\operatorname{std}_{i\in\mathcal E}A_i$, actor sampling uses
\begin{equation}
 w_i=\min\bigl\{e^{A_i/\max(\sigma_A,10^{-3})},100\bigr\},\; p_i=\frac{w_i}{\sum_{j\in\mathcal E}w_j}.
\label{eq:fpx_weights}
\end{equation}
Actor chunks are sampled with replacement by fixed $p_i$; full specifications are in the supplement.

\subsection{Restore-and-Continue Evaluation}
\label{sec:restore_evaluation}

\paragraph{Recovery.}
Under the original instruction, every arm resumes from the same point: the midpoint $m=\lfloor(a+b)/2\rfloor$ of each held-out failure's finalized interval. Restoration verifies exact MuJoCo-state equality, resets cached observations, and empties the action queue. Each point uses $K=3$ policy-sampling seeds shared across arms. Success follows the simulator's task predicates. Rollouts run for up to 1,800 control steps; the primary horizon is $H=1{,}200$, with 600/1,800-step sensitivity analyses. For points $\mathcal P$, let $y^H_{pk}(\pi)$ indicate success within $H$ steps:
\begin{equation}
 \mathrm{SR}_{\mathrm{rec}}^H(\pi;\mathcal P)
 =\frac{1}{K|\mathcal P|}\sum_{p\in\mathcal P}\sum_{k=1}^{K}y^H_{pk}(\pi).
\label{eq:srrec}
\end{equation}
Restore, physics, and execution errors count as failures in the fixed denominator.

\paragraph{Nominal task retention.}
A policy-independent validity check, fixed before nominal evaluation, retains 437 of 500 starts (50 configurations $\times$ 10 seeds) for all arms. We pool success over valid starts and report $\Delta\mathrm{SR}_{\mathrm{norm}}^H(\pi)=\mathrm{SR}_{\mathrm{norm}}^H(\pi)-\mathrm{SR}_{\mathrm{norm}}^H(\pi_{\mathrm{base}})$ at the recovery horizon \mbox{$H=1{,}200$}, adopted for nominal evaluation after nominal results of several arms had been seen, and at the pre-specified, task-dependent {$1.1T$} horizon; negative values indicate retention loss. Exploratory percentile intervals for this difference are paired by nominal start and resample configurations; resampling families instead leaves unchanged which intervals exclude zero (supplement).

\subsection{Research Questions and Inference}
\label{sec:controlled_comparisons}

\paragraph{Primary comparisons.}
Confirmatory scores average rollouts within points, points within family--level configurations, levels within families, and then the 10 families. Let $R_{\mathrm B},R_{\mathrm D},R_{\mathrm T},R_{\mathrm I}$ denote the L1--L3 scores for \mbox{Base-restart}, \mbox{Demo-continue}, \mbox{Takeover-BC}, and \mbox{IQL-R2}. Q1 compares training on the collected recovery dataset with continued training on original demonstrations through $\Delta_1=R_{\mathrm T}-R_{\mathrm D}$ at matched actor updates. Q2 uses $\Delta_2=R_{\mathrm I}-R_{\mathrm T}$ to compare value-weighted learning on recovery and failure data with recovery-only behavior cloning, jointly varying data composition and objective.

\paragraph{Statistical inference.}
We bootstrap paired family differences 10,000 times to obtain 95\% bias-corrected and accelerated (BCa) intervals~\cite{efron1987bca}, with percentile intervals as a sensitivity check and fallback for degenerate BCa calculations. These intervals are conditional on the fixed evaluated checkpoints and exclude training-seed variation. Q1 passes if its interval lies above zero and $\Delta_1\ge5$~pp; Q2 passes if its interval excludes zero and $|\Delta_2|\ge5$~pp. Under pre-specified fixed-sequence testing, Q2 is confirmatory only if Q1 passes.

\paragraph{Exploratory analyses.}
\mbox{Mix-BC}, nominal starts, L4--L5, and the additional value-based and inference-time variants (\mbox{IQL-R0}, AWAC, \mbox{First-of-32}, and \mbox{IDQL-R2}) are exploratory, with intervals unadjusted for multiple comparisons. The supplement provides full implementation specifications and sensitivity analyses, under which both primary decisions remain unchanged.

%% file: tables/tab_report_card.tex
\begin{table*}[t]
\centering
\caption{\textbf{Report card of all evaluated arms.} Recovery (\SRrec) on held-out failures and nominal success on the \nNormValid{} valid starts within \horizonMain{} steps unless noted; one training seed and $U=12{,}134$ actor updates per trained arm (\BaseRestart{} on RTX~5090, all other arms on H200; same-GPU nominal contrasts and the \nNormSens-start stable set are in the supplement). Brackets: 95\% intervals over the 10 families (\SRrec: percentile; $\Delta$: paired BCa) or 50 configurations (\dSRnorm: percentile). $\Delta$: L1--L3 difference arm $-$ \TakeoverBC{} (Q1 is negated \DemoCont, Q2 is \IQLR; unrounded). L4--L5: \nTestDiag{} points in nine families. $\mathcal F_{\mathrm{win}}$: failure-window chunks; $\mathcal N$: unsuccessful human attempts. $\dagger$: exploratory; ref.: reference.}
\label{tab:reportcard}
\scriptsize
\setlength{\tabcolsep}{2pt}
\begin{tabular*}{\textwidth}{@{\extracolsep{\fill}}lllrcrr rrr@{}}
\toprule
 & & & \multicolumn{4}{c}{Recovery, \SRrec{} (\%)} & \multicolumn{3}{c}{Nominal starts$^{\dagger}$} \\
\cmidrule(lr){4-7}\cmidrule(lr){8-10}
Arm & \makecell[bl]{Actor\\data} & Weighting & \multicolumn{1}{c}{\makecell[b]{L1--L3\\{}[95\% CI]}} & \makecell[b]{L1--L3\\@600\,/\,@1,800} & \multicolumn{1}{c}{\makecell[b]{L4--L5$^{\dagger}$}} & \multicolumn{1}{c}{\makecell[b]{$\Delta$ (pp)\\{}[BCa]}} & \multicolumn{1}{c}{\makecell[b]{SR\\(\%)}} & \multicolumn{1}{c}{\makecell[b]{\dSRnorm{} (pp)\\{}[95\% CI]}} & \multicolumn{1}{c@{}}{\makecell[b]{SR @1.1$T$\\(\%)}} \\
\midrule
\BaseRestart & -- & -- & 22.2 [16.1, 28.6] & 19.6 / 23.2 & 7.6 & \ensuremath{-}29.7 [\ensuremath{-}44.4, \ensuremath{-}18.5] & 43.9 & ref. & 33.6 \\
\DemoCont & $\mathcal D$ & -- & 22.5 [17.6, 27.9] & 19.4 / 23.3 & 6.0 & \ensuremath{-}29.4 [\ensuremath{-}44.7, \ensuremath{-}16.9] & 46.2 & +2.3 [\ensuremath{-}2.0, +7.0] & 34.1 \\
\TakeoverBC & $\mathcal R$ & -- & 51.9 [38.2, 64.9] & 46.1 / 54.0 & 36.8 & ref. & 14.0 & \ensuremath{-}30.0 [\ensuremath{-}37.9, \ensuremath{-}22.6] & 0.5 \\
Mix-BC$^{\dagger}$ & $\mathcal D,\mathcal R$ & equal-source & 45.3 [33.3, 57.2] & 38.8 / 46.7 & 26.2 & \ensuremath{-}6.7 [\ensuremath{-}11.4, \ensuremath{-}3.3] & 27.5 & \ensuremath{-}16.5 [\ensuremath{-}22.4, \ensuremath{-}10.0] & 2.7 \\
\IQLR & $\mathcal R,\mathcal F_{\mathrm{win}}$ & IQL ($\mathcal R,\mathcal F,\mathcal N$) & 53.5 [39.2, 67.5] & 48.8 / 54.6 & 33.6 & +1.5 [\ensuremath{-}2.2, +5.1] & 19.7 & \ensuremath{-}24.3 [\ensuremath{-}31.6, \ensuremath{-}16.3] & 2.3 \\
\midrule
IQL-R0$^{\dagger}$ & $\mathcal R$ & IQL ($\mathcal R$) & 48.9 [37.5, 60.0] & 44.2 / 50.3 & 32.6 & \ensuremath{-}3.1 [\ensuremath{-}7.6, +1.1] & 11.2 & \ensuremath{-}32.7 [\ensuremath{-}42.0, \ensuremath{-}24.0] & 0.5 \\
AWAC$^{\dagger}$ & $\mathcal R,\mathcal F_{\mathrm{win}}$ & one-step $Q$ & 53.8 [40.3, 66.3] & 48.8 / 55.8 & 30.6 & +1.8 [\ensuremath{-}4.4, +9.3] & 14.4 & \ensuremath{-}29.5 [\ensuremath{-}38.2, \ensuremath{-}20.9] & 0.7 \\
First-of-32$^{\dagger}$ & \TakeoverBC & first of 32 & 52.1 [39.2, 65.1] & 46.4 / 53.3 & 36.3 & +0.1 [\ensuremath{-}2.6, +2.2] & 14.6 & \ensuremath{-}29.3 [\ensuremath{-}37.1, \ensuremath{-}22.1] & 0.7 \\
IDQL-R2$^{\dagger}$ & \TakeoverBC & best of 32 ($Q$) & 40.7 [25.4, 58.1] & 38.9 / 41.5 & 17.8 & \ensuremath{-}11.2 [\ensuremath{-}19.6, +3.5] & 2.7 & \ensuremath{-}41.2 [\ensuremath{-}51.3, \ensuremath{-}31.2] & 0.0 \\
\bottomrule
\end{tabular*}
\end{table*}

%% file: sec/5_experiments.tex
\section{Experiments and Analysis}
\label{sec:experiments}
\input{figures/fig_results_float}

All arms start from the checkpoint, splits, and recipes of Secs.~\ref{sec:dataset}--\ref{sec:protocol}, with one completed training run per trained arm; \BaseRestart{} was evaluated on an RTX 5090 and the other arms on H200 GPUs. Results use horizon $H$ unless noted; differences are computed before rounding.

\subsection{Main Recovery Results}
\paragraph{Takeover data improve recovery beyond continued training (Q1).}
\TakeoverBC{} raises L1--L3 recovery success to \recLoTB{}\%, compared with \recLoDemo{}\% for \DemoCont{} and \recLoBase{}\% for \BaseRestart. At matched actor updates, the Q1 difference is $+\DeltaOne$~pp, with a 95\% family-paired BCa interval of [+\DeltaOneLo, +\DeltaOneHi{}], so Q1 passes (Fig.~\ref{fig:results}a) across 600--1,800 steps (Fig.~\ref{fig:results}c); on the \nBaseZeroLoPts{} points with $0/3$ \BaseRestart{} successes, \TakeoverBC{} reaches \srZeroLoTB{}\% \vs \srZeroLoDemo{}\%. Q1 evaluates the collected dataset as a whole, without separating its corrective actions from its coverage of failure states. \DemoCont{} minus \BaseRestart{} is \DeltaExtra~pp [\DeltaExtraLo, \DeltaExtraHi{}] across different GPUs.

\paragraph{\IQLR{} does not establish a further gain (Q2).}
\IQLR{} attains \recLoIQL{}\% recovery success, a Q2 difference of $+\DeltaTwo$~pp over \TakeoverBC{} with a 95\% BCa interval of [\DeltaTwoLo, +\DeltaTwoHi{}]. This difference meets neither pre-specified criterion, so it does not establish an advantage over recovery-only cloning for this recipe, which changes both data composition and the learning objective. No exploratory variant improves on \TakeoverBC{} with an interval excluding zero; \MixBC{} is lower by \vsTBAbsMix~pp [\vsTBAbsLoMix, \vsTBAbsHiMix{}] (Table~\ref{tab:reportcard}). The supplement shows rollouts of \BaseRestart, \TakeoverBC, and \IQLR{} from one restored held-out failure, selected after evaluation for a clear contrast.

\subsection{Recovery--Retention Trade-Off}
\paragraph{Recovery gains come with nominal losses (exploratory).}
Every takeover-trained arm loses nominal success relative to both \BaseRestart{} and H200-evaluated \DemoCont{} (\dSRnorm{} includes zero for \DemoCont; Fig.~\ref{fig:results}b, Table~\ref{tab:reportcard}). \TakeoverBC{} succeeds on \normAbsTB{}\% of valid starts \vs \normAbsBase{}\% for \BaseRestart: \dSRnorm{}${}=\dnormTB$~pp [\secFiveNomLoTB, \secFiveNomHiTB{}] (\dnormVsDemoTB~pp [\dnormVsDemoLoTB, \dnormVsDemoHiTB{}] \vs \DemoCont); \IQLR{} changes by \dnormIQL~pp [\secFiveNomLoIQL, \secFiveNomHiIQL{}]. Losses are steepest at L1 (Fig.~\ref{fig:results}e) and reflect slower SpaceMouse-scaled commands (\nomLateNTB/\nomSuccNTB{} successes finish after $1.1T$) alongside \horizonMain-step timeouts (\nomTimeoutPctTB{}\%, supplement), not pure skill forgetting.

\paragraph{\MixBC{} trades recovery for retention (exploratory).}
\MixBC{} retains more nominal success (\normAbsMix{}\%; \dnormMix~pp [\secFiveNomLoMix, \secFiveNomHiMix{}] relative to \BaseRestart) at lower \mbox{L1--L3} recovery (\recLoMix{}\%) than \TakeoverBC. This benefit depends on the evaluation horizon (Sec.~\ref{sec:restore_evaluation}). Under the stricter, pre-specified {$1.1T$} horizon, nominal success is \normAbsTBase{}\% for \BaseRestart{} and changes by \dnormTTB, \dnormTMix, and \dnormTIQL~pp for \TakeoverBC, \MixBC, and \IQLR; \MixBC's advantage over \TakeoverBC{} shrinks from \secFiveMixGainMain{} to \secFiveMixGainOriginal~pp.

\subsection{Heterogeneity and Sensitivity}
\input{figures/fig_families_float}
\paragraph{Gains and losses are uneven across task families and failure types.}
Per-family Q1 differences are positive in \nFamPosOne{} families, tied in \nFamTieOne{}, and negative in \nFamNegOne, ranging from \secFiveFamMin{} to \secFiveFamMax~pp (Fig.~\ref{fig:families}). The nominal change of \TakeoverBC{} is negative in every family but varies widely, and \TakeoverBC{} has no nominal success at any level in \nFamNormZero{} families (\famNormZero). By failure type over all levels, \TakeoverBC{} gains most over \BaseRestart{} and \DemoCont{} on collisions and dropped objects and falls below both only on unclosed drawers (supplement). Because failure types concentrate in a few families, these patterns are descriptive and do not identify a cause.

\paragraph{Decisions hold under horizon and family checks.}
Q1 passes at 600 and 1,800 steps (\DeltaOneSix{} and \DeltaOneEighteen~pp) and Q2 at neither; dropping any family (\lofoOneMin{} to \lofoOneMax~pp) or flagged width, late-success, and revised points (\qOneUnionEst~pp on \nRetainedUnionLo{} points, supplement) keeps both decisions. On exploratory L4--L5 failures, the point-estimate order of \IQLR{} and \TakeoverBC{} reverses (\recHiIQL{}\% \vs \recHiTB{}\%; Table~\ref{tab:reportcard}, Fig.~\ref{fig:results}d); training includes L4--L5, measuring held-out recovery within observed configurations.

\paragraph{The critic weighting fails its pre-specified check.}
The \IQLR{} critic's mean sampling weights are \secFiveRecoveryWeight{} on $\mathcal R$ and \secFiveFailureWeight{} on $\mathcal F_{\mathrm{win}}$, failing the check $\text{Mean}(\mathcal R)>\text{Mean}(\mathcal F_{\mathrm{win}})$, yet $\mathcal R$ still supplies \pctDrawRecIQL{}\% of actor draws (\nRecChunksPost{} \vs \nFwinChunksPost{} chunks; \pctDrawRecAWAC{}\% in AWAC, which passes the check; supplement); this characterizes the weighting without identifying Q2's cause. Exploratory \mbox{IQL-R0} (critic and actor on $\mathcal R$ alone) differs from \TakeoverBC{} by \vsTBIQLzero~pp [\vsTBLoIQLzero, \vsTBHiIQLzero{}], and \IQLR{} minus \mbox{IQL-R0} is \DeltaIQLzero~pp [\DeltaIQLzeroLo, \DeltaIQLzeroHi{}], varying both critic experience and actor data.

%% file: figures/fig_results_float.tex
\begin{figure*}[t]
\centering
\includegraphics[width=\textwidth]{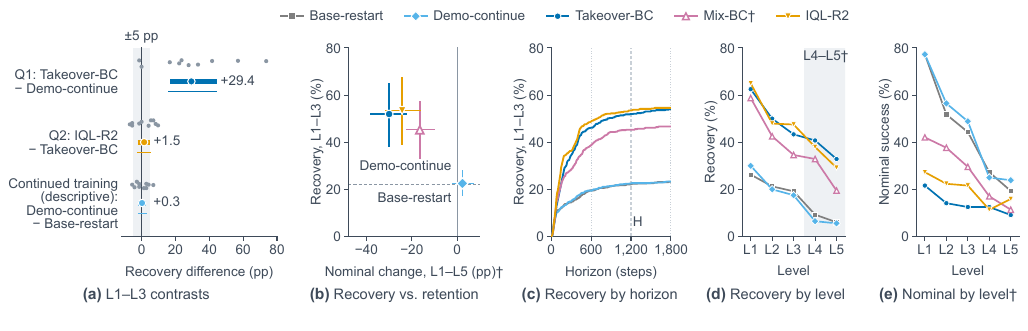}
\vspace{-5mm}
\caption{\textbf{Takeover data raise recovery beyond continued training, at a cost in nominal success.} Horizon $H$ except in (c). (a)~L1--L3 contrasts: 95\% BCa (thick) and percentile (thin) intervals over 10 families; dots: per-family differences; band: $\pm$5\,pp criterion. (b)~L1--L3 recovery \vs L1--L5 nominal change from \BaseRestart{} (gray-line crossing; 95\% intervals over families / configurations). (c)~Cumulative L1--L3 recovery by horizon (dotted: the \horizonShort- and \horizonLong-step sensitivity horizons). (d,\,e)~Recovery and nominal success by level. $\dagger$:~exploratory; continued training in (a) is descriptive.}
\label{fig:results}
\end{figure*}

%% file: figures/fig_families_float.tex
\begin{figure}[t]
\centering
\includegraphics[width=\linewidth]{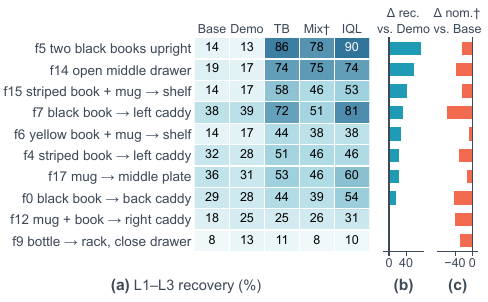}
\vspace{-5mm}
\caption{\textbf{Gains and losses are uneven across task families (descriptive).} (a)~L1--L3 recovery per family and arm (\%, 72 rollouts per cell), sorted by the gain in (b). (b)~Recovery gain of \TakeoverBC{} over \DemoCont{} (pp). (c)~Nominal change of \TakeoverBC{} from \BaseRestart{} over the family's valid L1--L5 starts (pp). TB: \TakeoverBC; IQL: \IQLR; $\dagger$:~exploratory.}
\label{fig:families}
\end{figure}

%% file: sec/6_conclusion.tex
\section{Conclusion and Outlook}
\label{sec:conclusion}
\label{sec:limitations}

\fpx{} measures what human takeovers contribute to offline VLA post-training at restorable failure states. For the evaluated checkpoint, cloning the collected recoveries improves 1,200-step L1--L3 recovery success by \DeltaOne{}~pp over continued demonstration training, whereas the tested \IQLR recipe does not establish a further gain. These recovery gains come with substantial nominal-task losses, an exploratory result that holds under both nominal horizons and argues for evaluating recovery and retention together.

This study is limited to one checkpoint, one operator, one training run per trained arm, and simulation. It does not test the effect of policy matching itself, takeover timing, or gains over new demonstrations of equal human time (supplement). Next steps include restarts from several states along each held-out failure, factorial designs that separate data composition from the learning objective, and skill-preserving objectives. With restorable states and logged attempts, these studies need no new takeovers. New methods plug into the same recipe and policy interfaces.

%% file: sec/X_llm_statement.tex
\section*{LLM Usage Statement}
We used Claude (Anthropic) to edit, restructure
and condense the text of each section, to search for and summarize related work, and to write scripts that lay out and render figures from our experimental outputs. The authors designed the method and experiments, conducted all experiments and analyses, and verified every statement, number, citation and figure in the paper. No LLM was used to construct the benchmark data or to form the research ideas, hypotheses or conclusions. All AI-assisted code was reviewed and tested by the authors. The authors take full responsibility for the accuracy, integrity, and final content of this work.

%% file: supp/appendix.tex
\clearpage
\begingroup
\input{supp/supplementary_setup}

\fpxsuppstart
\maketitlesupplementary
\input{supp/supplementary_content}
\clearpage
\endgroup

%% file: supp/supplementary_setup.tex
\let\fpxPendingOrig\pending
\renewcommand{\pending}[1]{\leavevmode\fpxPendingOrig{#1}}
\let\fpxAuthorInputOrig\authorinput
\renewcommand{\authorinput}[1]{\leavevmode\fpxAuthorInputOrig{#1}}
\let\fpxPendcellOrig\pendcell
\renewcommand{\pendcell}{\leavevmode\fpxPendcellOrig}

\renewcommand{\floatpagefraction}{0.5}
\renewcommand{\dblfloatpagefraction}{0.5}
\makeatletter
\setlength{\@fptop}{0pt}\setlength{\@dblfptop}{0pt}
\renewcommand\paragraph{\@startsection{paragraph}{4}{\z@}{1.0ex \@plus .5ex \@minus .2ex}{-1em}{\normalfont\normalsize\bfseries}}
\makeatother

\newcommand{\Sfinal}{S$_\text{final}$\xspace}
\newcolumntype{L}{>{\raggedright\arraybackslash}X}
\newcolumntype{P}[1]{>{\raggedright\arraybackslash}p{#1}}
\newcolumntype{C}[1]{>{\centering\arraybackslash}p{#1}}
\newcommand{\yesT}{Yes$^{\ddagger}$}
\setlist[description]{font=\normalfont\bfseries, leftmargin=0pt, itemsep=1pt, topsep=2pt}

%% file: supp/supplementary_content.tex
\input{supp/app_a}      %
\input{supp/app_cases}  %
\input{supp/app_b}      %
\input{supp/app_c}      %
\input{supp/app_d}      %
\input{supp/app_e}      %
\input{supp/app_f}      %
\input{supp/app_h}      %

%% file: supp/app_a.tex
\section{Benchmark Construction: Platform, Base Policy, Task Families and Screening}
\label{sec:s-construction}
\label{sec:s-protocol}

This section describes how the benchmark was built around one \pizf checkpoint and \nFamilies \mbox{LIBERO-X} task families: the simulator and base policy (\cref{sec:s-platform-base}), the task families, levels and configurations (\cref{sec:s-families-levels}), the screening of base-policy rollouts into annotation candidates (\cref{sec:s-funnel}) and base-policy success by level (\cref{sec:s-base-success}). Annotation and collection counts refer to the data as frozen at the end of collection.

\subsection{Platform and base policy}
\label{sec:s-platform-base}

\begin{description}
\item[Simulator.] \mbox{LIBERO-X}~\cite{wang2026liberox} (commit \texttt{\liberoxCommit}) with its kitchen and study scenes, built on robosuite~\cite{zhu2020robosuite} (\robosuiteVersion) and MuJoCo~\cite{todorov2012mujoco} (\mujocoVersion, EGL rendering, Python~3.10.16); control runs at \controlHz~Hz.
\item[Base policy.] \pizf~\cite{black2025pi05}, fully fine-tuned from the \pizf base model with openpi~\cite{openpi} (commit \texttt{\openpiCommit}, Python~3.11.17, JAX~0.5.3) on all \nDemos \mbox{LIBERO-X} demonstrations (\nDemoFrames frames; dataset revision \texttt{\liberoxDataRev}) for \baseTrainSteps steps (batch 256, action horizon 10; 4$\times$H200 GPUs, \baseTrainHours~h). The 7-D actions are zero-padded to the model's 32 action dimensions, and the EMA weights of the final step form the base checkpoint (SHA-256 \texttt{6a6c6aa1\allowbreak da11176d\allowbreak e1c0e16f\allowbreak 662fe0f5\allowbreak c24eda07\allowbreak dab80cd4\allowbreak 8ca1e8ac\allowbreak 38397c11}).
\item[Frames and control steps.] A training chunk spans 10 consecutive demonstration frames, and at evaluation all 10 predicted actions are executed, one per control step, before the policy is queried again. One demonstration frame corresponds to one control step at \controlHz~Hz, although \mbox{LIBERO-X} lists its demonstrations at 10\,fps. Per unit of commanded translation, the end effector moves \demoDispPerFrame{}\,m per demonstration frame (median over the \nDemos episodes) and \simDispPerStep{}\,m per control step (\nKinRollouts of our rollouts); a frame spanning two steps would roughly double the per-frame value. Successful L1 rollouts also take a median of \demoLenRatioMed times the mean demonstration length in steps. A chunk therefore spans 0.5\,s.
\end{description}

\subsection{Task families, levels and configurations}
\label{sec:s-families-levels}

\begin{table*}[tp]
\centering
\footnotesize
\tabstyle
\caption[Task families]{\textbf{Task families.} Each family (Fam.) is one \mbox{LIBERO-X} task in a kitchen or study scene (\mbox{LIBERO-X} scene number), posed at five levels (\cref{sec:s-families-levels}). Task: shortened instruction shared by L1--L3; L1 and L2 apply local and extended spatial perturbations, and L3 reconfigures the scene topology. Objects at L4 and L5: objects that replace the task objects at L4, in order of mention; L5 poses the L4 task with paraphrased instructions. $^{a}$The target also changes (family~12: back to left compartment; family~17: middle to nearest plate). $^{b}$L4 repeats the L3 task (identical \mbox{LIBERO-X} definitions). $^{c}$The L4 instruction omits ``upright''.}
\label{tab:s-bddl}
\tabzebra
\begin{tabularx}{\textwidth}{r l L l}
\toprule
\tabhead Fam. & Scene & Task at L1--L3 (shortened instruction) & Objects at L4 and L5 \\
\midrule
0 & Study 22 & Black book into the back desk caddy's front compartment & navy speckle book \\
4 & Study 10 & Left striped book into the left desk caddy's right compartment & pink wavy book \\
5 & Study 43 & Right and left black books upright on top of the shelf & navy speckle books \\
6 & Study 46 & Yellow book upright and red mug on top of the shelf & ketchup, butter$^{c}$ \\
7 & Study 3 & Right black book into the left desk caddy's front compartment & white speckle book \\
9 & Kitchen 46 & Wine bottle from the drawer onto the wine rack's top layer; close the drawer & cookies \\
12 & Study 11 & Red mug and yellow book into the right desk caddy's right and back compartments & BBQ sauce, ketchup$^{a}$ \\
14 & Kitchen 4 & Open the middle drawer of the white cabinet & unchanged$^{b}$ \\
15 & Study 50 & Right striped book upright and red mug on top of the shelf & chocolate pudding, milk$^{c}$ \\
17 & Kitchen 49 & Porcelain mug from its plate onto the middle plate (relative to the robot) & milk$^{a}$ \\
\bottomrule
\end{tabularx}
\end{table*}

\begin{description}
\item[Families.] \nFamilies families, each one \mbox{LIBERO-X} task across the \nLevels levels (\cref{tab:s-bddl}), with our own IDs 0, 4, 5, 6, 7, 9, 12, 14, 15 and 17: \nFamiliesKitchen kitchen (9, 14, 17) and \nFamiliesStudy study. Family~17 replaces a planned family in which an object's initial region overlapped a fixture.
\item[Task definitions.] Each configuration uses one \mbox{LIBERO-X} task definition (scene layout, objects, goal and instruction). L1 and L2 use the family's L1 and L2 definitions of the same task, L3 the L3 task with the same scene and instruction, L4 the first L4 variant of that task (only family~0 has several), and L5 the L4 task with the five \mbox{LIBERO-X} paraphrases of its instruction; the $i$-th seed of a block uses paraphrase $\lfloor i/2\rfloor \bmod 5+1$, as in the \mbox{LIBERO-X} evaluation protocol. For family~14, \mbox{LIBERO-X} defines identical L3 and L4 tasks, so its L4 configuration repeats the L3 task and its L5 configuration differs from L3 only by the paraphrased instructions.
\item[Levels.] As defined by \mbox{LIBERO-X}: local spatial perturbation (L1), extended spatial perturbation (L2), scene-topology reconfiguration (L3), visual-attribute changes including unseen objects and distractors (L4), and semantically equivalent instruction paraphrases (L5). Levels vary task difficulty, not policy strength.
\item[Configurations.] \nConfigs family--level configurations, of which \nConfigsComplete are complete: they have \nTrainPerConfig{} training sources with \framesPerFailure{} accepted recoveries each and \nTestPerConfig{} held-out failures. Family~9 L4 and L5 are incomplete (\cref{tab:s-configs}).
\item[Mean demonstration length.] $T$ is the mean length, in control steps (one per frame; \cref{sec:s-platform-base}), of the training demonstrations whose instruction equals the task's instruction; the \nTUnmatched of the \nLOneTasks \mbox{LIBERO-X} L1 tasks without a matching demonstration get the median over instructions, \TmedianSteps steps. All ten families have matched demonstrations, and their L2--L5 configurations inherit $T$ from the L1 task. \mbox{LIBERO-X} does not publish per-task limits, so $T$ is our reconstruction; it sets the \timeGateScreen{}$T$ and \timeGateCollect{}$T$ cutoffs (\cref{sec:s-funnel}).
\item[Family selection.] The pre-specified family screening (a fixed order, a capability gate of success within \timeGateScreen{}$T$, and quotas over scene type and single- and multi-step tasks) was not completed; instead, a family was retained if every level yielded \nTrainPerConfig{} qualified training failures.
\end{description}

\subsection{Rollout generation and failure screening}
\label{sec:s-funnel}

This subsection gives the screening rules summarized in Sec.~3.2 of the main paper and the resulting counts.

\begin{description}
\item[Seeds.] Rollouts use pre-assigned seed blocks, starting at \seedTrain for training and at \seedTest for held-out candidates.
\item[Failure.] A rollout that has not succeeded by \timeGateCollect{}$T$; rollouts that succeed later (late successes) remain failures under this definition (\cref{sec:s-flags}).
\item[Exclusions.] Rollouts with an unstable start (the start fails the physics guard's 2\,s hold-still check) or with a physics fault during the episode (the guard fires) count as neither success nor failure. Failures are excluded as near-misses when they fail solely the 3\,cm centering condition, and as invalid starts under the start-validity rule (object penetration above 2\,mm, or objects ejected at the start). The start-validity rule was not applied to families 0 and 9 before annotation.
\item[Candidates for annotation.] In each configuration, nominally the first \nTrainPerConfigShip training and \nTestPerConfigShip held-out candidates in seed order are sent to annotation; actual counts in \cref{tab:s-rollout-funnel,tab:s-funnel}.
\end{description}

\begin{table*}[tp]
\centering
\scriptsize
\setlength{\tabcolsep}{2.6pt}
\tabstyle
\caption[Rollout funnel]{\textbf{Rollout funnel per configuration and seed block.} Fam.: task family; Lv.: level. Gen.: rollouts generated from contiguous seeds starting at \seedTrain{}/\seedTest{}, the sum of the next five columns (rollouts generated only for validation or screening, replaced rollout pools and unused reserve rollouts are not counted). Unst.\ start: the start failed the physics guard's 2\,s hold-still check; Phys.\ fault: the guard fired during the episode; neither counts as success or failure. Succ.: success within \timeGateScreen{}$T$, or after \timeGateScreen{}$T$ but within \timeGateCollect{}$T$ ($T$: mean demonstration length, \cref{sec:s-families-levels}). Fail.: no success by \timeGateCollect{}$T$. Near miss: failures excluded because they fail solely the 3\,cm centering condition. Start inv.: failures excluded by the start-validity rule; an ejected object was identified by a displacement above 10\,cm (family~5) or relative to the object's median displacement in its configuration (families 4, 6, 7, 12, 14, 15 and 17); --: rule not applied before annotation (families~0 and~9). Cand.: candidates, Fail.\ minus Near miss and Start inv. Sent: sent to annotation, nominally the first \nTrainPerConfigShip{}/\nTestPerConfigShip{} candidates in seed order (family~17 L1--L3 also sent later candidates; \cref{sec:s-flags}). All: column totals.}
\label{tab:s-rollout-funnel}
\tabzebra
\begin{tabular}{ll rrrrrrrrrr rrrrrrrrrr}
\toprule
\tabhead & & \multicolumn{10}{c}{Training seed block} & \multicolumn{10}{c}{Held-out seed block} \\
\cmidrule(lr){3-12}\cmidrule(lr){13-22}
\tabhead Fam. & Lv. & Gen. & \makecell{Unst.\\start} & \makecell{Phys.\\fault} & \makecell{Succ.\\${\le}1.1T$} & \makecell{Succ.\\$({1.1},{1.5}]T$} & Fail. & \makecell{Near\\miss} & \makecell{Start\\inv.} & Cand. & Sent & Gen. & \makecell{Unst.\\start} & \makecell{Phys.\\fault} & \makecell{Succ.\\${\le}1.1T$} & \makecell{Succ.\\$({1.1},{1.5}]T$} & Fail. & \makecell{Near\\miss} & \makecell{Start\\inv.} & Cand. & Sent \\
\midrule
0 & L1 & 114 & 33 & 1 & 46 & 15 & 19 & 0 & -- & 19 & 10 & 91 & 20 & 0 & 31 & 20 & 20 & 0 & -- & 20 & 16 \\
 & L2 & 61 & 17 & 0 & 18 & 12 & 14 & 0 & -- & 14 & 10 & 69 & 14 & 0 & 26 & 14 & 15 & 0 & -- & 15 & 15 \\
 & L3 & 33 & 13 & 0 & 3 & 7 & 10 & 0 & -- & 10 & 10 & 42 & 10 & 0 & 4 & 8 & 20 & 0 & -- & 20 & 16 \\
 & L4 & 30 & 13 & 0 & 1 & 1 & 15 & 0 & -- & 15 & 10 & 30 & 6 & 1 & 0 & 2 & 21 & 0 & -- & 21 & 16 \\
 & L5 & 30 & 13 & 1 & 0 & 1 & 15 & 0 & -- & 15 & 10 & 30 & 6 & 2 & 1 & 0 & 21 & 0 & -- & 21 & 16 \\
\midrule
4 & L1 & 272 & 91 & 0 & 167 & 5 & 9 & 0 & 0 & 9 & 9 & 990 & 307 & 1 & 638 & 20 & 24 & 0 & 0 & 24 & 16 \\
 & L2 & 91 & 27 & 1 & 39 & 13 & 11 & 0 & 1 & 10 & 10 & 232 & 75 & 0 & 107 & 23 & 27 & 0 & 0 & 27 & 16 \\
 & L3 & 48 & 17 & 0 & 13 & 0 & 18 & 0 & 0 & 18 & 10 & 38 & 6 & 1 & 11 & 0 & 20 & 0 & 0 & 20 & 16 \\
 & L4 & 45 & 16 & 1 & 14 & 2 & 12 & 0 & 0 & 12 & 10 & 65 & 17 & 2 & 29 & 0 & 17 & 0 & 0 & 17 & 16 \\
 & L5 & 85 & 28 & 3 & 28 & 1 & 25 & 0 & 0 & 25 & 10 & 109 & 33 & 4 & 49 & 1 & 22 & 0 & 0 & 22 & 16 \\
\midrule
5 & L1 & 199 & 70 & 1 & 114 & 1 & 13 & 0 & 0 & 13 & 10 & 254 & 67 & 7 & 152 & 1 & 27 & 0 & 0 & 27 & 16 \\
 & L2 & 30 & 12 & 0 & 8 & 0 & 10 & 0 & 0 & 10 & 10 & 70 & 20 & 5 & 25 & 3 & 17 & 0 & 2 & 15 & 15 \\
 & L3 & 44 & 17 & 1 & 15 & 0 & 11 & 0 & 0 & 11 & 10 & 59 & 23 & 1 & 20 & 0 & 15 & 0 & 1 & 14 & 14 \\
 & L4 & 33 & 13 & 1 & 6 & 0 & 13 & 0 & 0 & 13 & 10 & 42 & 15 & 1 & 9 & 0 & 17 & 0 & 2 & 15 & 15 \\
 & L5 & 44 & 17 & 3 & 6 & 0 & 18 & 0 & 0 & 18 & 10 & 43 & 16 & 1 & 9 & 1 & 16 & 0 & 2 & 14 & 14 \\
\midrule
6 & L1 & 918 & 305 & 4 & 553 & 43 & 13 & 0 & 0 & 13 & 10 & 918 & 282 & 3 & 574 & 50 & 9 & 0 & 0 & 9 & 9 \\
 & L2 & 50 & 16 & 1 & 14 & 8 & 11 & 0 & 0 & 11 & 10 & 127 & 42 & 2 & 32 & 19 & 32 & 0 & 2 & 30 & 16 \\
 & L3 & 30 & 7 & 2 & 6 & 0 & 15 & 0 & 0 & 15 & 10 & 48 & 15 & 1 & 12 & 2 & 18 & 0 & 2 & 16 & 16 \\
 & L4 & 30 & 6 & 0 & 0 & 0 & 24 & 0 & 0 & 24 & 10 & 30 & 9 & 2 & 0 & 0 & 19 & 0 & 0 & 19 & 16 \\
 & L5 & 30 & 6 & 4 & 0 & 0 & 20 & 0 & 0 & 20 & 10 & 30 & 9 & 1 & 0 & 0 & 20 & 0 & 0 & 20 & 16 \\
\midrule
7 & L1 & 213 & 78 & 3 & 117 & 5 & 10 & 0 & 0 & 10 & 10 & 289 & 96 & 4 & 164 & 8 & 17 & 0 & 0 & 17 & 16 \\
 & L2 & 129 & 47 & 1 & 66 & 2 & 13 & 0 & 3 & 10 & 10 & 208 & 85 & 2 & 94 & 7 & 20 & 0 & 1 & 19 & 16 \\
 & L3 & 181 & 70 & 3 & 84 & 13 & 11 & 0 & 1 & 10 & 10 & 145 & 48 & 4 & 68 & 8 & 17 & 0 & 4 & 13 & 13 \\
 & L4 & 101 & 33 & 0 & 53 & 3 & 12 & 0 & 1 & 11 & 10 & 129 & 45 & 3 & 62 & 4 & 15 & 0 & 3 & 12 & 12 \\
 & L5 & 77 & 27 & 1 & 32 & 0 & 17 & 0 & 0 & 17 & 10 & 62 & 21 & 1 & 24 & 0 & 16 & 0 & 0 & 16 & 16 \\
\midrule
9 & L1 & 40 & 0 & 0 & 21 & 5 & 14 & 0 & -- & 14 & 10 & 60 & 0 & 0 & 35 & 9 & 16 & 0 & -- & 16 & 16 \\
 & L2 & 30 & 1 & 0 & 9 & 2 & 18 & 0 & -- & 18 & 10 & 30 & 0 & 0 & 7 & 1 & 22 & 0 & -- & 22 & 16 \\
 & L3 & 30 & 1 & 0 & 5 & 2 & 22 & 0 & -- & 22 & 10 & 30 & 1 & 0 & 6 & 6 & 17 & 0 & -- & 17 & 16 \\
 & L4 & 30 & 0 & 0 & 0 & 0 & 30 & 0 & -- & 30 & 10 & 30 & 0 & 0 & 0 & 0 & 30 & 0 & -- & 30 & 16 \\
 & L5 & 30 & 0 & 0 & 0 & 0 & 30 & 0 & -- & 30 & 10 & 30 & 0 & 0 & 0 & 0 & 30 & 0 & -- & 30 & 16 \\
\midrule
12 & L1 & 75 & 19 & 2 & 40 & 3 & 11 & 0 & 0 & 11 & 10 & 66 & 9 & 0 & 38 & 2 & 17 & 0 & 0 & 17 & 16 \\
 & L2 & 77 & 23 & 0 & 41 & 1 & 12 & 0 & 3 & 9 & 9 & 74 & 9 & 1 & 40 & 4 & 20 & 0 & 4 & 16 & 16 \\
 & L3 & 30 & 6 & 0 & 7 & 2 & 15 & 0 & 0 & 15 & 10 & 61 & 18 & 1 & 16 & 7 & 19 & 0 & 0 & 19 & 16 \\
 & L4 & 30 & 6 & 3 & 0 & 0 & 21 & 0 & 0 & 21 & 10 & 30 & 3 & 1 & 0 & 0 & 26 & 0 & 0 & 26 & 16 \\
 & L5 & 30 & 6 & 2 & 0 & 0 & 22 & 0 & 0 & 22 & 10 & 30 & 3 & 2 & 1 & 0 & 24 & 0 & 0 & 24 & 16 \\
\midrule
14 & L1 & 182 & 0 & 0 & 121 & 50 & 11 & 0 & 0 & 11 & 10 & 446 & 0 & 0 & 280 & 144 & 22 & 0 & 0 & 22 & 16 \\
 & L2 & 82 & 0 & 1 & 37 & 20 & 24 & 0 & 3 & 21 & 10 & 66 & 0 & 0 & 26 & 19 & 21 & 0 & 5 & 16 & 16 \\
 & L3 & 110 & 1 & 0 & 66 & 23 & 20 & 0 & 1 & 19 & 10 & 88 & 0 & 0 & 49 & 21 & 18 & 0 & 0 & 18 & 16 \\
 & L4 & 78 & 0 & 0 & 53 & 15 & 10 & 0 & 0 & 10 & 10 & 101 & 0 & 0 & 58 & 24 & 19 & 0 & 0 & 19 & 16 \\
 & L5 & 30 & 0 & 0 & 15 & 5 & 10 & 0 & 0 & 10 & 10 & 54 & 0 & 0 & 22 & 13 & 19 & 0 & 0 & 19 & 16 \\
\midrule
15 & L1 & 113 & 35 & 2 & 64 & 1 & 11 & 0 & 0 & 11 & 10 & 180 & 57 & 4 & 103 & 2 & 14 & 0 & 0 & 14 & 14 \\
 & L2 & 33 & 14 & 0 & 8 & 1 & 10 & 0 & 0 & 10 & 10 & 43 & 15 & 1 & 12 & 0 & 15 & 0 & 2 & 13 & 13 \\
 & L3 & 30 & 11 & 2 & 0 & 1 & 16 & 0 & 0 & 16 & 10 & 43 & 18 & 0 & 3 & 4 & 18 & 0 & 0 & 18 & 16 \\
 & L4 & 30 & 5 & 0 & 0 & 0 & 25 & 0 & 1 & 24 & 10 & 30 & 7 & 1 & 0 & 0 & 22 & 0 & 0 & 22 & 16 \\
 & L5 & 30 & 5 & 0 & 0 & 0 & 25 & 0 & 1 & 24 & 10 & 30 & 7 & 1 & 0 & 0 & 22 & 0 & 0 & 22 & 16 \\
\midrule
17 & L1 & 428 & 0 & 0 & 255 & 85 & 88 & 63 & 0 & 25 & 14 & 609 & 0 & 0 & 363 & 145 & 101 & 69 & 0 & 32 & 18 \\
 & L2 & 90 & 0 & 0 & 51 & 15 & 24 & 5 & 0 & 19 & 15 & 76 & 0 & 0 & 34 & 14 & 28 & 4 & 1 & 23 & 20 \\
 & L3 & 62 & 0 & 0 & 29 & 9 & 24 & 12 & 0 & 12 & 9 & 95 & 0 & 0 & 33 & 22 & 40 & 13 & 0 & 27 & 17 \\
 & L4 & 30 & 0 & 0 & 0 & 6 & 24 & 0 & 0 & 24 & 10 & 30 & 0 & 0 & 0 & 5 & 25 & 0 & 0 & 25 & 16 \\
 & L5 & 30 & 0 & 0 & 0 & 3 & 27 & 0 & 0 & 27 & 10 & 30 & 0 & 0 & 0 & 3 & 27 & 0 & 0 & 27 & 16 \\
\midrule
\multicolumn{2}{l}{All} & 4,678 & 1,125 & 44 & 2,225 & 381 & 903 & 80 & 15 & 808 & 506 & 6,512 & 1,434 & 61 & 3,267 & 636 & 1,114 & 86 & 31 & 997 & 781 \\
\bottomrule
\end{tabular}
\end{table*}

\begin{table*}[tp]
\centering
\scriptsize
\setlength{\tabcolsep}{4.2pt}
\tabstyle
\caption[Annotation funnel]{\textbf{Annotation funnel per configuration.} Fam.: task family and scene (K: kitchen, S: study); Lv.: level. Sent: sent to annotation; Ann.: annotated at least once; Rel.: released into the dataset; Wh.: withheld; Und.: no decision; Sent = Rel.\ + Wh.\ + Und. Width: released held-out intervals flagged by the interval-width check (\cref{sec:s-flags}). All: column totals. $^{a}$Family~9 L4/L5: collection stopped before the quotas were met (note to \cref{tab:s-configs}).}
\label{tab:s-funnel}
\tabzebra
\begin{tabular}{ll rrrrr rrrrrr}
\toprule
\tabhead & & \multicolumn{5}{c}{Training candidates} & \multicolumn{6}{c}{Held-out candidates} \\
\cmidrule(lr){3-7}\cmidrule(lr){8-13}
\tabhead Fam. & Lv. & Sent & Ann. & Rel. & Wh. & Und. & Sent & Ann. & Rel. & Wh. & Und. & Width \\
\midrule
\suppFunnelRows
\bottomrule
\end{tabular}
\end{table*}

\paragraph{Rollout funnel.} The training and held-out seed blocks of the ten families produced \nRolloutsGen rollouts (\nRolloutsGenTrain training, \nRolloutsGenTest held-out; \cref{tab:s-rollout-funnel}). Of these, \nGenInvalidStart had an unstable start and \nGenPhysFault a physics fault during the episode, \nGenSuccFast succeeded within \timeGateScreen{}$T$ and \nGenSuccSlow between \timeGateScreen{}$T$ and \timeGateCollect{}$T$. Of the remaining \nGenFail failures, \nGenNearMiss near-misses and \nGenStartInvalid invalid starts were excluded, leaving \nCandidates candidates. Nominally, the first \nTrainPerConfigShip training and \nTestPerConfigShip held-out candidates in seed order (all of them where fewer existed) were sent to annotation, \nShipped in total; for family~17 L1--L3, later candidates that never succeeded within the episode were also sent.

\paragraph{Annotation funnel.} Of the \nShipped candidates sent to annotation, \nAnnotatedRollouts were annotated (\suppAnnTrain training, \suppAnnTest held-out), with \nAnnotationRecords annotation records in total. The coordinator released \nFailReleased (\nFailTrain training, \nFailTest held-out) and withheld \nWithheld (\suppWhTrain training, \suppWhTest held-out). By the annotator's judgment, \suppWhNoFailure withheld candidates showed no failure (\suppWhNoFailureLate of them late successes, all in family~17), \suppWhUncertain were judged uncertain (families 5, 7 and 17) and \suppWhFailure were judged failures. The remaining \nUndecided candidates received no decision: \suppAnnUnd were annotated and \suppNotAnn were never annotated. Apart from the \suppUndFamNine in family~9 L4/L5, where collection stopped, all undecided candidates belong to configurations whose quotas were filled. \Cref{tab:s-funnel} gives the counts per configuration, and \cref{sec:s-flags} lists the curation decisions behind them.

\subsection{Base-policy success by level}
\label{sec:s-base-success}

On all \nLOneTasks \mbox{LIBERO-X} L1 tasks, each from its 10 official initial states and evaluated as in \mbox{LIBERO-X} (10 no-op steps, then up to \horizonMain policy steps), the base policy succeeds in \baseSRLOneCap{}\% of episodes within \horizonMain steps and in \baseSRLOneT{}\% within \timeGateScreen{}$T$; counting a demonstration frame as two control steps, which doubles $T$, would give \baseSRLOneTwoStep{}\% within \timeGateScreen{}$T$. On the L1 tasks of the ten families, the same evaluation gives \baseSRFamLOneCap{}\% and \baseSRFamLOneT{}\%, respectively. \Cref{tab:s-basesr} reports base success by level on the ten families from the nominal evaluation of the base checkpoint, which draws its starts from fixed seeds rather than from the official initial states (\cref{sec:s-evaluation}).

\begin{table}[t]
\centering\footnotesize
\tabstyle
\caption[Base-policy success by level]{\textbf{Base-policy success (\%) by level on the ten families.} Nominal evaluation of the base checkpoint with 10 seeded starts per configuration. Valid starts: starts that pass the policy-independent validity check, the only ones scored (\cref{sec:s-evaluation}). Success within \horizonMain policy steps and within \timeGateScreen{}$T$ ($T$: mean demonstration length, \cref{sec:s-families-levels}); physics-guard faults, \nBaseNormFaultValid of the \nNormValid valid starts, count as failures. All: pooled over levels.}
\label{tab:s-basesr}
\tabzebra
\input{supp/tab_s_basesr}
\end{table}

%% file: supp/tab_s_basesr.tex
\begin{tabular}{lrrr}
\toprule
\tabhead Level & Valid starts & $\le$\,\horizonMain steps & $\le$\,\timeGateScreen{}$T$ \\
\midrule
L1 & 88 & 77.3 & 63.6 \\
L2 & 85 & 51.8 & 44.7 \\
L3 & 88 & 44.3 & 26.1 \\
L4 & 88 & 27.3 & 19.3 \\
L5 & 88 & 19.3 & 14.8 \\
\midrule
All & \nNormValid & \normAbsBase & \normAbsTBase \\
\bottomrule
\end{tabular}

%% file: supp/app_cases.tex
\section{Case Studies}
\label{sec:s-case-studies}

This section illustrates the benchmark on recorded examples: one training failure of each observed failure type (\cref{sec:s-gallery}) and the rollouts of \BaseRestart, \TakeoverBC and \IQLR from one held-out failure (\cref{sec:s-cases}).

\begin{figure*}[t]
\centering
\includegraphics[width=\textwidth]{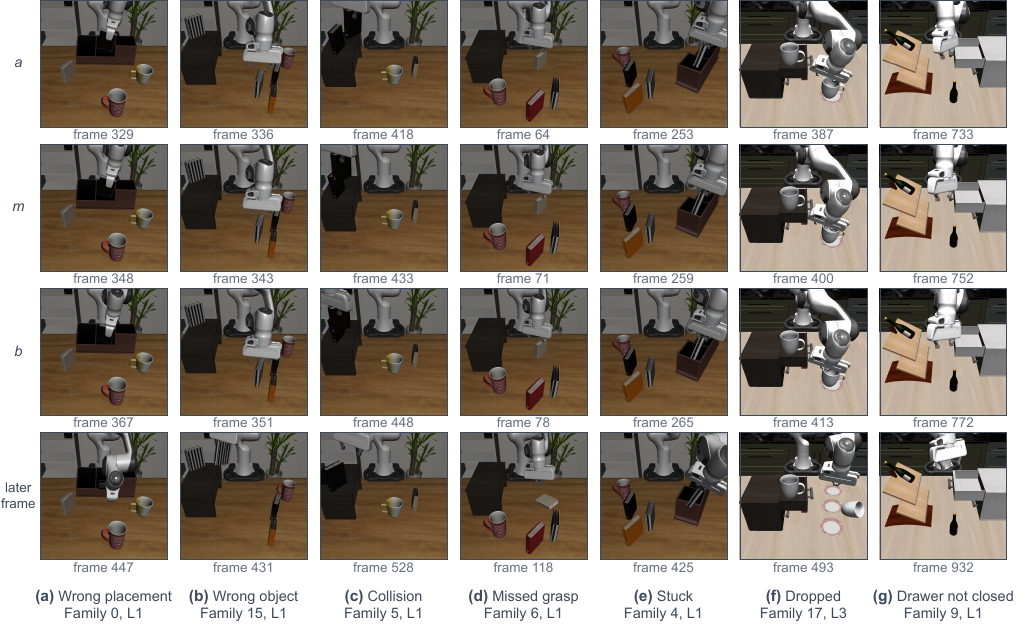}
\caption{\textbf{Recorded examples of the seven observed failure types.} Each column, (a)--(g), follows one training failure of the base policy in the fixed third-person view: the frames at the start $a$, the midpoint $m=\lfloor(a+b)/2\rfloor$ and the end $b$ of the annotated interval, and one later frame outside it. Gray numbers: frame indices of the recorded rollout. Below each column: failure type, task family and level; the task of each family is given in \cref{tab:s-bddl}. Frames are unedited simulator renders; the examples were chosen for visibility.}
\label{fig:s-gallery}
\end{figure*}

\begin{figure*}[t]
\centering
\includegraphics[width=\textwidth]{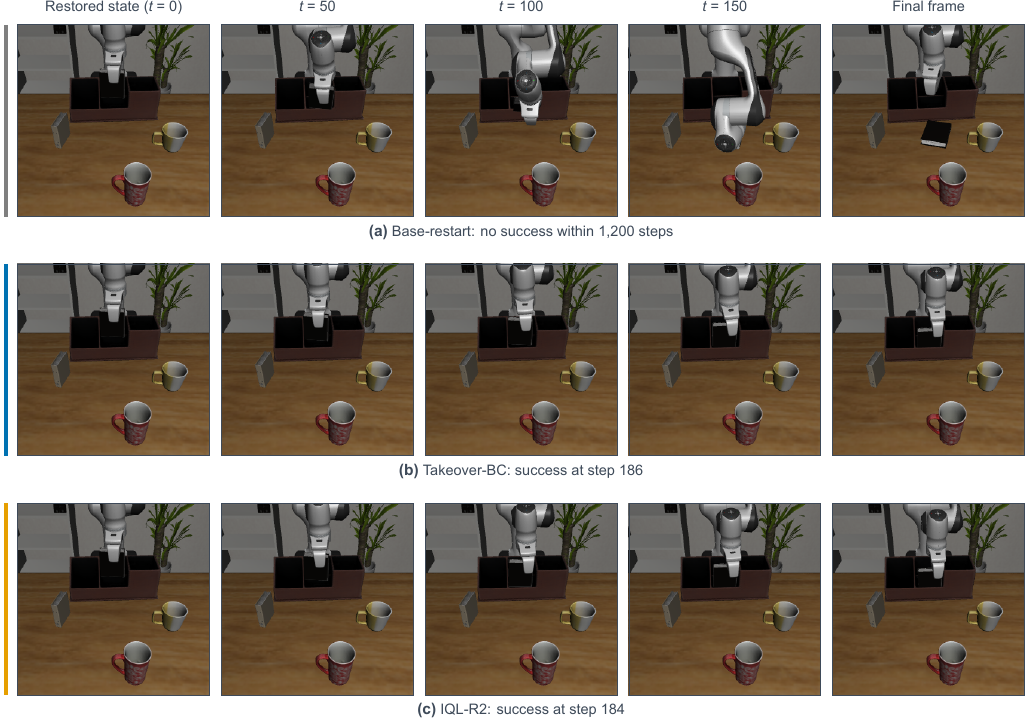}
\caption{\textbf{Recovery rollouts of three arms from one held-out failure (qualitative).} Task (family~0, L1): ``place the black book into the back desk caddy's front compartment''; the base policy's rollout is annotated as a wrong placement in $[266,297]$. (a)--(c)~\BaseRestart, \TakeoverBC and \IQLR, each restarted from the restored state at the midpoint $m=281$ (first column) with the same policy-sampling seed. Columns show the fixed third-person view $t$ steps after the restart; the last column is the final frame of each rollout, at the success step or at the \horizonMain-step horizon. Colored bars: arm colors used throughout the paper. The point was selected after evaluation for a clear contrast and carries no statistical weight; the frames are re-runs of the evaluated checkpoints (\cref{sec:s-cases}).}
\label{fig:s-cases}
\end{figure*}

\subsection{Recorded examples of each failure type}
\label{sec:s-gallery}

\Cref{fig:s-gallery} shows one training failure for each of the seven failure types observed in the annotations (definitions in \cref{sec:s-ftypes}). Each column follows the rollout through its annotated interval $[a,b]$, which brackets the onset of the failure event: the start $a$, the midpoint $m$ (where held-out failures are restored) and the end $b$, followed by one later frame outside the interval. The examples were chosen for visibility in the third-person view and do not reflect label frequencies, which \cref{tab:s-ftype-level} reports.

\subsection{Recovery rollouts from one held-out failure}
\label{sec:s-cases}

\Cref{fig:s-cases} follows \BaseRestart, \TakeoverBC and \IQLR from the same held-out failure, a wrong placement of the black book in family~0 at L1 (annotated interval $[266,297]$). All three arms restart from the restored state at the midpoint $m=281$ with the same policy-sampling seed. When the \horizonMain-step horizon ends, \BaseRestart has not completed the task and the book lies on the table in front of the caddy; \TakeoverBC and \IQLR place the book in the target compartment and succeed after 186 and 184 steps, respectively.

The point was selected after evaluation by a fixed rule: it is the first L1--L3 held-out point, in identifier order, at which all three evaluated \BaseRestart rollouts fail and all three evaluated rollouts of both \TakeoverBC and \IQLR succeed within \horizonMain steps. It shows a clear contrast, not typical behavior; aggregate results are in \cref{sec:s-results}. The frames are re-runs of the evaluated checkpoints on an RTX~5090 GPU (the same GPU model as the frozen \BaseRestart{} evaluation) from the restored state with the first of the $K$ policy-sampling seeds (\cref{sec:s-restore}). They reproduce the evaluated outcomes and the success steps of \TakeoverBC and \IQLR; the evaluated \BaseRestart rollout ended in a physics-guard fault at step 371 (a failure under the protocol), whereas its re-run on the same GPU model reaches the horizon without a fault.

%% file: supp/app_b.tex
\section{Failure Annotation and Human Takeover Collection}
\label{sec:s-collection}

This section defines the failure annotation and the takeover protocol, reports the collection effort, and summarizes the recorded data-quality flags and curation decisions.

\subsection{Interval guideline and failure-type definitions}
\label{sec:s-ftypes}

For each candidate, the annotator records an interval $[a,b]$ (target width \intervalMin--\intervalMax frames), one of eight failure-type labels, a no-single-event-frame flag and a judgment. The coordinator then releases the candidate into the dataset (a released failure) or withholds it, under quotas of \nTrainPerConfig training and \nTestPerConfig held-out failures per configuration, and finalizes the interval and label of each released failure. One annotator produced all \nAnnotationRecords annotation records; there is no second annotation or adjudication, so inter-annotator agreement is not estimated.

\paragraph{Label definitions.} Each label has one fixed definition. \emph{Missed grasp}: the gripper does not grasp the intended object. \emph{Dropped}: a grasped object leaves the gripper and falls. \emph{Collision}: another object is knocked over or pushed away. \emph{Wrong placement}: the target object is placed at the wrong location or on the wrong layer. \emph{Stuck}: the robot is stuck, circles in place or makes no progress for a sustained time; the interval marks where this is observed. \emph{Wrong object}: a non-target object is taken (in the wine-bottle task, including the second bottle on the table). \emph{Drawer not closed}: closing the original drawer fails, is blocked or remains incomplete. \emph{Other}: another observable problem, described in a short note; this label was never chosen.

\paragraph{Interval guideline.} The guideline reads: ``Mark that the failure happens between these two frames: start $a$ is the earliest frame at which you think the failure may already have begun; end $b$ is the frame by which the failure has certainly happened, for example when the object leaves the gripper, is knocked over, or the gripper opens after a wrong placement. If unsure, widen the interval a little rather than forcing a single frame.'' Takeovers use only the interval of the main event, which the annotator chooses and which need not be the first or last event; an interval longer than 2\,s is recorded as it is, is not treated as a point-like failure and receives no takeovers.

\paragraph{Flag and judgment.} The no-single-event-frame flag is set when the robot is stuck or circling, with the interval placed near the start of a sustained (${\geq}\,2$\,s) lack of progress. Judgments are ``failure'' (a failure can be confirmed), ``no failure'' (no clear failure observed) and ``uncertain'' (images or evidence insufficient); only a failure carries an interval and a label.

\subsection{Failure-type labels by split and level}

\Cref{tab:s-ftype-level} gives the final labels of the \nFailReleased released failures by split and level. Training counts include the five released sources without a complete takeover set (\cref{sec:s-configs}), and L4 and L5 have no family-9 held-out failures. Wrong placement (\pctftWrongPlacement{}\%) and wrong object (\pctftWrongObject{}\%), the latter an object-selection (grounding) error, together account for \pctftGrounding{}\% of the released failures, followed by collisions (\pctftCollision{}\%; \cref{fig:s-data}a). Wrong placements are the most frequent label at L1--L3, while wrong objects and missed grasps account for most L4--L5 failures; because release follows fixed quotas per configuration, these are counts in the curated set, not failure rates.

\begin{table*}[t]
\centering
\footnotesize
\tabstyle
\caption{\textbf{Failure-type labels by split and level.} Final annotations of the \nFailReleased released failures (labels defined in \cref{sec:s-ftypes}). L1--L5: levels; All: split total; Total: both splits. The label ``other'' was never chosen and is omitted.}
\label{tab:s-ftype-level}
\input{supp/tab_s_ftype_level}
\end{table*}

\subsection{Takeover protocol and device mapping}
\label{sec:s-takeover}

\begin{figure*}[tp]
\centering
\includegraphics[width=\textwidth]{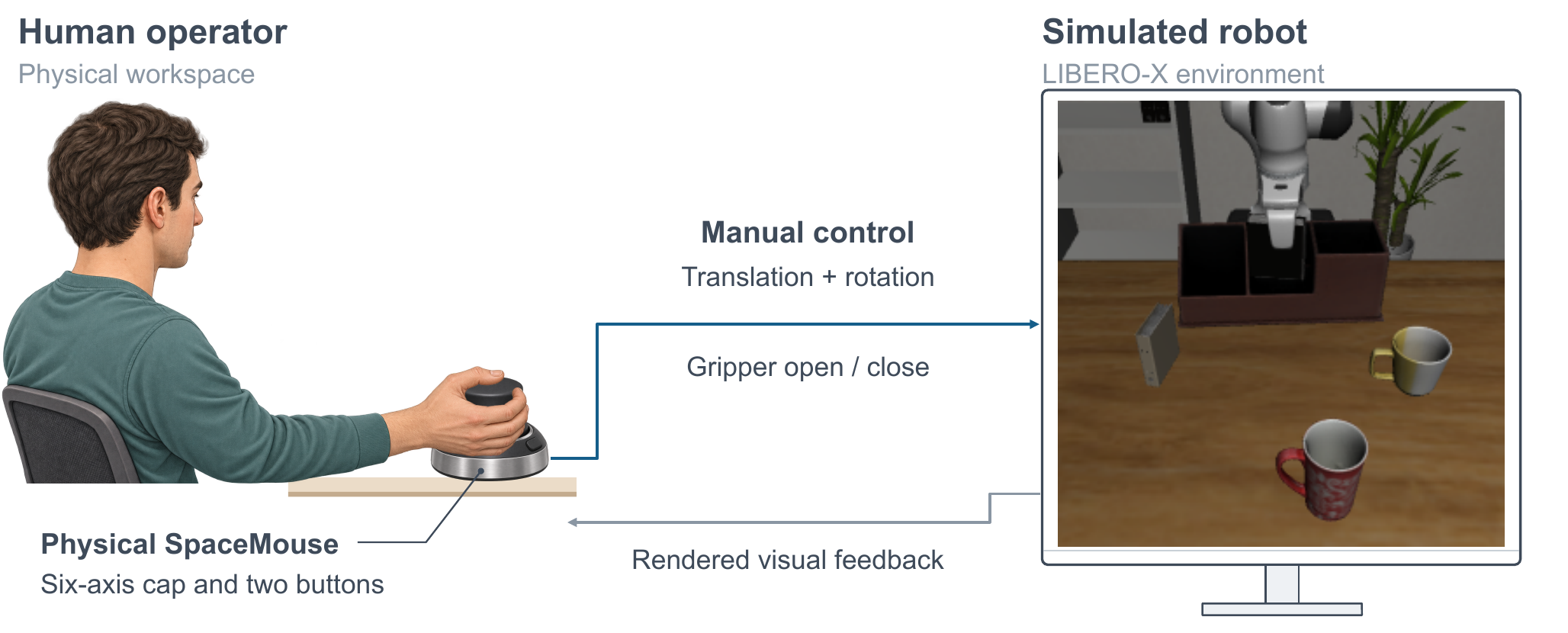}
\caption{\textbf{SpaceMouse teleoperation of the simulated robot.} The operator moves a physical SpaceMouse while watching the rendered simulator view: six-axis cap deflections command end-effector translation and rotation, and the two buttons open and close the gripper (\cref{sec:s-takeover}). The arm and objects are simulated in \liberox; no physical robot is involved. The operator and the device are an AI-generated illustration, not a photograph of the collection; the monitor shows an unmodified recorded simulator frame.}
\label{fig:s-spacemouse}
\end{figure*}

Takeovers are collected on training failures only, under the following protocol; \cref{fig:s-spacemouse} illustrates the control loop from the physical device to the simulation.

\begin{description}
\item[Starts.] Protocol-timed takeovers: \framesPerFailure frames per training failure, sampled uniformly without replacement inside $[a,b]$ with a recorded seed, so the operator does not choose when to take over; each start state is restored from the recorded rollout.
\item[Operator and budget.] One operator with a SpaceMouse at \controlHz~Hz; \budgetSteps steps (\budgetSec~s) per attempt.
\item[Device mapping.] The device commands the action space of the demonstrations, delta end-effector pose control in \liberox. At every control step, the bias-corrected six-axis deflection (dead zone 0.05, no smoothing) is mapped to translations along and rotations about $x$, $y$, $z$ (forward, left, up) with fixed gains of 0.25 (translation) and 0.08 (rotation). Commands therefore stay within $\pm0.25$ and $\pm0.08$ of the normalized range $[-1,1]$, i.e.\ at most 1.25\,cm and 0.04\,rad per step under the controller's scaling of 0.05\,m and 0.5\,rad. The two buttons open ($-1$) and close ($+1$) the gripper, and each takeover starts with the gripper command in force at the restored state. Executed commands are stored unchanged, and all \nAttemptsAll attempts use the same mapping.
\item[Command scale.] The demonstrations, collected by VR teleoperation, use larger commands: the 99th percentile of $|a|$ is \demoPNNTrans per translation axis and \demoPNNRot per rotation axis, against 0.25 and 0.08 for the accepted recoveries; \demoAboveCapTrans{}\% of demonstration steps exceed 0.25 on each translation axis (\demoAboveCapRot{}\% exceed 0.08 on each rotation axis).
\item[Redo rule.] Attempts ending in an operator stop, a time limit or a physics fault are redone at the same frame until success, as are successes that the operator discards; frame replacements are listed in \cref{tab:s-flags}.
\item[Acceptance.] A takeover is accepted if it succeeds by the \liberox success predicate with a normal episode end, is not discarded by the operator, and passes automatic integrity checks: its first recorded state equals the restored start state, its state, action and time records are aligned at \controlHz~Hz, and both camera streams are complete. No takeover was rejected (\cref{sec:s-configs}).
\item[Retention.] Every attempt is kept unmodified, including operator stops, time limits, physics faults and discarded successes (\cref{tab:s-configs}).
\end{description}

\subsection{Collection effort and attempt outcomes}
\label{sec:s-configs}

\input{supp/dataset_characteristics}

\paragraph{Review and curation decisions.} Acceptance relies on the automatic checks of \cref{sec:s-takeover}, which verify task success and data integrity for every takeover; no takeover was replayed by a reviewer. In family~0 (L4 and L5), six sampled start frames whose attempts had ended in physics faults were replaced by frames drawn uniformly from the remaining frames of the same interval (three approved replacement requests; one further request was rejected); the original frames and attempts are kept. The operator also discarded \nDiscardedSucc successful attempts (\nFamilyFiveDisc of them in family~5) at their own discretion, without a predefined criterion, and redid them at the same frame.

\paragraph{Per-configuration counts.} \Cref{tab:s-configs} breaks the collection down by family and level, with the counting conventions of the main paper.
\begin{itemize}
\item \textbf{Candidates.} \nShipped candidates were sent to annotation (\nShippedTrain training, \nShippedTest held-out). The nominal number was \nTrainPerConfigShip training and \nTestPerConfigShip held-out candidates per configuration; \suppShipNominalCfgTrain of the \nConfigs configurations received the nominal training count and \suppShipNominalCfgTest the nominal held-out count, with actual counts from \suppShipTrainMin to \suppShipTrainMax (training) and from \suppShipTestMin to \suppShipTestMax (held-out).
\item \textbf{Training sources.} The \nFailTrain released training failures comprise \nSourcesComplete complete active sources in \nConfigsComplete configurations (\framesPerFailure accepted recoveries each, \nAccepted in total) and five released sources without a complete takeover set, which are not used for post-training: \suppArchRel archived sources (training failures replaced in takeover collection by other released failures; their attempts are kept) and the \suppFamNineLFourRel family-9 L4 sources (note~$a$ of \cref{tab:s-configs}). A further archived source (family~0 L2) was withdrawn from the released set and is counted as withheld.
\item \textbf{Attempts.} Of the \nAttemptsAll attempts, \nAttemptsActive are on active sources and \nAttemptsArchived on archived sources (\suppArchSucc successes, \suppArchStop operator stops, \suppArchTL time limits). \Cref{tab:s-configs} counts active sources; over all attempts there are \suppStopsAll operator stops and \suppTLAll time limits.
\end{itemize}

\begin{table*}[tp]
\centering
\scriptsize
\tabstyle
\setlength{\tabcolsep}{3.6pt}
\renewcommand{\arraystretch}{0.9}%
\caption{\textbf{Collection counts per configuration.} Fam.: family and scene (K: kitchen, S: study); Lv.: level. Sent: sent to annotation. Released training failures include the \suppArchRel archived sources that were not withdrawn. Archived (att.): archived sources (their attempts), not in the attempt columns. Repl.: requests to replace sampled frames, approved or rejected (\cref{tab:s-flags}). Attempt columns count the attempts on active (non-archived) sources by end reason (Disc.\ succ.: discarded success; Stop: operator stop; Phys.\ fault: physics fault); accepted plus discarded successes are the successes. Human SR: successes over attempts minus physics faults. \emph{all}: family subtotal; --: no attempts. $^{a}$Family~9 L4/L5: its \suppFamNineLFourAtt L4 attempts were made at the same sampled start and both reached the \budgetSec{}\,s limit; L4 collection then stopped, and L5 was not collected. The \suppFamNineLFourRel released L4 sources remain in the data, and no family-9 L4/L5 held-out candidate was released.}
\label{tab:s-configs}
\tabzebra
\begin{tabular}{ll ccc rrr rrrrrr r}
\toprule
\tabhead & & \multicolumn{3}{c}{Candidates (training/held-out)} & \multicolumn{3}{c}{Training sources} & \multicolumn{6}{c}{Takeover attempts on active sources} & \\
\cmidrule(lr){3-5}\cmidrule(lr){6-8}\cmidrule(lr){9-14}
\tabhead Fam. & Lv. & Sent & Released & Withheld & Active & \makecell{Archived\\(att.)} & Repl. & All & Accepted & \makecell{Disc.\\succ.} & Stop & \makecell{Time\\limit} & \makecell{Phys.\\fault} & \makecell{Human\\SR (\%)} \\
\midrule
\suppConfigRows
\bottomrule
\end{tabular}
\end{table*}

\begin{table*}[t]
\centering
\footnotesize
\tabstyle
\setlength{\tabcolsep}{4pt}
\renewcommand{\tabgroup}[2]{\rowcolor{TabGroup}\multicolumn{#1}{l}{\rule{0pt}{2.2ex}\textbf{#2}}\\}
\caption{\textbf{Data-quality flags and curation decisions.} Main flags and decisions recorded for the released failures and the takeover attempts; details in \cref{sec:s-configs,sec:s-flags}. Count: records affected. Scope: split, task family (Fam.) and level (L1--L5). Handling: treatment in the data and in the analyses.}
\label{tab:s-flags}
\begin{tabularx}{\textwidth}{P{3.7cm} P{2.9cm} P{3.6cm} L}
\toprule
\tabhead Flag or decision & Count & Scope & Handling \\
\midrule
\tabgroup{4}{Intervals and labels}
Width check failed ($b-a+1<10$ or ${>}\,41$) & \suppWidthFlag & Held-out; \suppWidthFlagFamFive in Fam.~5 & Kept as annotated; midpoints are evaluation starts. \\
Interval revised after attempts & \suppOutOfInterval attempts & One training source, Fam.~7 L5 & Attempts kept; frames redrawn inside the revised interval. \\
Single annotator & \nAnnotationRecords records & All annotations & No second annotation or adjudication; no agreement estimate. \\
\midrule
\tabgroup{4}{Candidate selection}
Candidates above the nominal number & \suppCfgOverTestCap held-out, \suppCfgOverTrainNominal training configurations & Fam.~17 L1--L3 & Later candidates sent only if their rollouts never succeeded; released failures stay at the quota. \\
Late successes, released & \suppLateTrain training, \suppLateTest held-out & \suppLateTestConf of the \nTestConfirm held-out L1--L3 points & Retained as failures; intervals end before the success. \\
Late successes, withheld & \suppWhNoFailureLate & Fam.~17 & Withheld as showing no failure. \\
\midrule
\tabgroup{4}{Takeover attempts}
Acceptance & \nAccepted decisions & All accepted recoveries & Automatic success and integrity checks only; no takeover replayed by a reviewer. \\
Archived-source successes & \suppArchSucc & Archived sources & Kept without review; not used for post-training. \\
Discarded successes & \nDiscardedSucc & Training sources; \nFamilyFiveDisc in Fam.~5 & Discarded at the operator's discretion (no predefined criterion) and redone at the same frame; kept. \\
Frame replacements & 6 frames (3 requests) & Fam.~0 L4, L5 & Redrawn inside the same interval after physics faults; original frames and attempts kept; one further request rejected. \\
Attempt cap exceeded & \suppOverCap source & Fam.~15 L2 & \attPerSourceMax attempts against a cap of \suppAttemptCap; source kept. \\
Attempts without a decision & \suppPending & Fam.~9 L4; archived Fam.~7 L2 source & Not used for training or evaluation. \\
Archived sources & \nSourcesArchived & Fam.~0 L2 and L5; Fam.~7 L2 & \suppArchRel remain in the released set; attempts kept and counted separately; not used for post-training. \\
\bottomrule
\end{tabularx}
\end{table*}

\subsection{Data-quality flags and curation decisions}
\label{sec:s-flags}

\Cref{tab:s-flags} summarizes the main data-quality flags and curation decisions recorded for the released failures and the takeover attempts, with their counts, scope and handling.

\paragraph{Interval width.} A width check flags intervals whose inclusive frame count $b-a+1$ is below 10 or above 41. It flags \suppWidthFlag released held-out intervals (\suppWidthFlagFamFive in family~5; last column of \cref{tab:s-funnel}) and no released training interval, because the coordinator adjusted the family-5 and family-9 training intervals annotated outside this range. Flagged held-out intervals are kept as annotated, and their midpoints remain evaluation starts.

\paragraph{Interval revisions.} One training source (family~7 L5) was relabeled from wrong placement to wrong object with an earlier interval after the operator had stopped \suppOutOfInterval attempts at a frame of the original interval. Its ten takeover frames were redrawn inside the revised interval; the stopped attempts are kept but lie outside the final interval.

\paragraph{Late successes.} Rollouts that succeed after \timeGateCollect{}$T$ are failures under the failure definition (\cref{sec:s-funnel}) and are retained. \suppLateTrain of the \nFailTrain released training failures and \suppLateTest of the \nFailTest held-out failures (\suppLateTestConf of the \nTestConfirm L1--L3 points) reached the goal later in the episode, at 1.1--2.7 times the \timeGateCollect{}$T$ cutoff. All \suppLateTrain training sources are used for post-training, and in all \suppLateRel cases the annotated interval ends before the success. A further \suppWhNoFailureLate late successes in family~17 were withheld as showing no failure.

\paragraph{Candidate selection.} For family~17 (L1--L3), later candidates were also sent to annotation, chosen because their rollouts never succeeded within the episode; candidates that succeeded late were not added. As a result, \suppCfgOverTestCap of these configurations exceed the held-out cap of \nTestPerConfigShip{} candidates and \suppCfgOverTrainNominal exceed the nominal \nTrainPerConfigShip{} training candidates (\cref{tab:s-funnel}); their released failures stay at the quota.

%% file: supp/tab_s_ftype_level.tex
\tabzebra
\begin{tabular}{l rrrrr r rrrrr r r}
\toprule
\tabhead & \multicolumn{6}{c}{Training} & \multicolumn{6}{c}{Held-out} & \\
\cmidrule(lr){2-7}\cmidrule(lr){8-13}
\tabhead Label & L1 & L2 & L3 & L4 & L5 & All & L1 & L2 & L3 & L4 & L5 & All & Total \\
\midrule
wrong placement & 24 & 24 & 18 & 5 & 7 & 78 & 32 & 35 & 31 & 11 & 8 & 117 & 195 \\
wrong object & 6 & 12 & 17 & 25 & 22 & 82 & 10 & 12 & 18 & 34 & 34 & 108 & 190 \\
collision & 9 & 10 & 6 & 4 & 5 & 34 & 14 & 14 & 16 & 7 & 9 & 60 & 94 \\
missed grasp & 1 & 0 & 1 & 13 & 10 & 25 & 1 & 3 & 1 & 16 & 16 & 37 & 62 \\
stuck & 3 & 3 & 4 & 0 & 1 & 11 & 12 & 7 & 3 & 1 & 3 & 26 & 37 \\
dropped & 5 & 2 & 4 & 0 & 1 & 12 & 9 & 5 & 6 & 3 & 2 & 25 & 37 \\
drawer not closed & 2 & 1 & 0 & 0 & 0 & 3 & 2 & 4 & 5 & 0 & 0 & 11 & 14 \\
\midrule
All & 50 & 52 & 50 & 47 & 46 & 245 & 80 & 80 & 80 & 72 & 72 & 384 & 629 \\
\bottomrule
\end{tabular}

%% file: supp/dataset_characteristics.tex
\label{sec:s-data-analysis}

\input{supp/figure_dataset_characteristics}

Without any post-training, the data describe how one \pizf checkpoint fails on \nFamilies \liberox task families and what recovery costs a human operator (\cref{fig:s-data}; per-level totals are in Table~2 of the main paper).

\paragraph{Recovery effort.}
Attempt-level human success on active (non-archived) training failures is \humanSR{}\% with operator stops and time limits as failures and physics faults excluded, \humanSRphysFail{}\% with physics faults as failures, and \humanSRstopVoid{}\% with stops as void; discarded successes count as successes throughout.
Frame-level success on complete training failures is 100\% by construction (same-frame redo), not a capability measure.
Operator stops make up \pctStopsOfUnsuccessful{}\% of the \nUnsuccessful unsuccessful attempts; only \nTimeLimit reached the \budgetSec{}\,s limit.
A complete training failure took \attPerSourceMean attempts on average (median \attPerSourceMedian, maximum \attPerSourceMax{}), and \nExactlyTen of the \nSourcesComplete needed no retry.
The amount of accepted data per failure is fixed; the human effort behind it is not.
Per-family success spans \suppHSRFamMin--\suppHSRFamMax{}\%; the maximum (family~5) depends on counting its \nFamilyFiveDisc discarded successes, over half of all \nDiscardedSucc, as successes.

\paragraph{Durations and horizon.}
The median accepted recovery lasts \medianSteps steps (\medianSec{}\,s) and lengthens with level (\cref{fig:s-data}b); attempt-level human success by level is plotted in \cref{fig:s-data}c.
Of the accepted recoveries, \pctWithinSixHundred{}\% end within \horizonShort steps and all but \nOverTwelveHundred within $H=$~\horizonMain steps, the evaluation horizon of Sec.~4.2 of the main paper.

\paragraph{Cost and curation.}
Annotation took \hoursAnnotation{}\,h of recorded labeling time, and takeovers took \hoursControl{}\,h of active control across all attempts; the accepted recoveries alone contain \hoursAcceptedControl{}\,h of control at \controlHz{}\,Hz.
Human decisions also shape the data through \nDiscardedSucc operator-discarded successes (\pctDiscardedOfSucc{}\% of successes), coordinator-approved frame replacements and \nSourcesArchived archived sources (\cref{tab:s-flags}), and through \nWithheld withheld candidates (\cref{tab:s-funnel}); all are counted per configuration in \cref{tab:s-configs}.

%% file: supp/figure_dataset_characteristics.tex
\begin{figure*}[t]
\centering
\includegraphics[width=\linewidth]{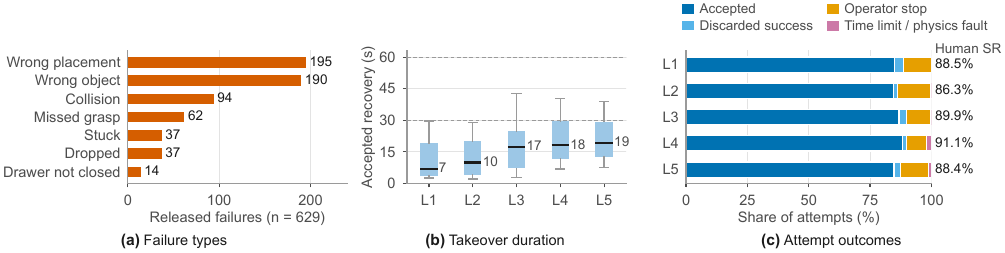}
\caption{\textbf{Failure types and human recovery effort.}
(a)~Failure types: counts of the \nFailReleased released failures by label.
(b)~Takeover duration: the \nAccepted accepted recoveries by level; box: interquartile range; line and label: median (s); whiskers: 10th and 90th percentiles (nearest rank); dashed: \horizonShort- and \horizonMain-step horizons at \controlHz{}\,Hz.
(c)~Attempt outcomes: shares of the \nAttemptsActive attempts on active training failures by level and end reason; right: attempt-level human success (Human SR), successes (discarded ones included) over attempts minus physics faults.
Table~2 of the main paper gives the per-level median durations, attempt outcomes and human success.}
\label{fig:s-data}
\end{figure*}

%% file: supp/app_c.tex
\section{Post-Training Implementation Details}
\label{sec:s-posttrain}
\label{sec:s-posttrain-implementation}

This section details the post-training methods of Sec.~4.1 of the main paper: what a method may use, the actor recipe shared by all trained arms, the critics of the value-based arms, the two inference-time variants and the compute spent on training and evaluation.

\subsection{Shared actor recipe}
\label{sec:s-recipe}

\paragraph{Information access.} The benchmark is instantiated once, for a single base checkpoint $\pi_{\mathrm{base}}$. A method receives $\pi_{\mathrm{base}}$, the \nDemos LIBERO-X demonstrations, the training failures (rollout prefix, final interval and failure-type label), all takeover attempts on them and the actor-update budget $U$; held-out failure states, seeds and labels are hidden. Training failures, takeover start states and held-out failures all come from $\pi_{\mathrm{base}}$, so the data are policy-matched by construction; no arm uses failures or takeovers collected from another policy.

\paragraph{Development set.} Tuning was permitted only on a development set of \nDevRoots complete training failures with \nDevStarts takeover starts: the fifth non-archived training failure in seed order in each configuration of family~0 (L1--L5) and of family~9 (L1--L3). The recipe below was fixed without tuning, so these failures served only for implementation checks. They are excluded from all training, including the actor data, the critic data and the critic's normalization statistics.

\paragraph{Training data.} The training data comprise, per configuration, the \nTrainPerConfig lowest-seed non-archived training failures with \framesPerFailure accepted recoveries each. Without the development failures, post-training uses \nTrainSrcPost source failures from all five levels, with \nTrainAccPost accepted recoveries and \nTrainStepsPost control steps. As in the main paper, $\mathcal D$ denotes the demonstrations, and $\mathcal R$, $\mathcal F$ and $\mathcal N$ denote the accepted recoveries, the failed policy prefixes and the unsuccessful human attempts (ending in operator stops or time limits) of these training sources; physics-fault attempts are excluded.

\paragraph{Flow-matching objective.} All actors keep the flow-matching loss of the base \pizf model as implemented in openpi~\cite{openpi}. As in the base checkpoint, the actor conditions on the camera images and the instruction but not on proprioception, which enters only the critic (\cref{sec:s-value-arms}). For actor context $x$, a normalized chunk $\mathbf a$ of 10 actions (the action horizon of the base model), standard Gaussian noise $\epsilon$ and sampled flow time $\lambda$, write
\begin{equation}
\begin{split}
 z_\lambda &= (1-\lambda)\mathbf a+\lambda\epsilon,\\
 \ell_{\mathrm{FM}}(\theta;x,\mathbf a)
 &=\mathbb E_{\epsilon,\lambda}\!\left[\operatorname{MSE}\!\left(v_\theta(z_\lambda,\lambda;x),\epsilon-\mathbf a\right)\right].
\end{split}
\label{eq:s-fm}
\end{equation}
Here $v_\theta$ predicts the flow velocity, and MSE averages over action dimensions and chunk steps. Noise and flow time are sampled as for the base model.

\paragraph{Optimization.} Every trained arm starts from the base checkpoint, freezes the vision-language backbone and the base weights of the action expert, and trains rank-32 ($\alpha=32$) LoRA~\cite{hu2022lora} adapters on the attention and feed-forward layers of the action expert together with the action and time projections. All trained arms keep the quantile normalization statistics of the base checkpoint and use AdamW ($\beta_1=0.9$, $\beta_2=0.95$, $\epsilon=10^{-8}$, weight decay $10^{-10}$, no parameter EMA) with global-norm gradient clipping at 1.0, a cosine schedule with 500 warm-up steps (peak learning rate $5\times10^{-5}$, final $5\times10^{-6}$), global batch 64 and $U=12{,}134$ actor updates (checkpoint step 12,133). Complete 10-action chunks (starts up to 10 frames before an episode end) are sampled with replacement; each arm is trained with seed~42 and evaluated at its final checkpoint.

\paragraph{Arms and actor data.} \BaseRestart runs the base checkpoint without further training and serves as the untrained reference. \DemoCont trains further on $\mathcal D$ alone and is the continued-training control of Q1; \TakeoverBC trains on $\mathcal R$ and enters both Q1 and Q2; the exploratory \MixBC draws source $\mathcal D$ or $\mathcal R$ with probability 0.5 each and then samples complete chunks uniformly within the chosen source; and \IQLR, the Q2 arm, trains on $\mathcal R$ and on failure-window chunks of $\mathcal F$ with value-based sampling weights (\cref{sec:s-value-arms}). The other four arms are exploratory variants of \IQLR (\mbox{IQL-R0} and AWAC; \cref{sec:s-value-arms}) and of \TakeoverBC (\mbox{First-of-32} and \mbox{IDQL-R2}; \cref{sec:s-inference-variants}).

\input{supp/posttrain_implementation}

\subsection{Training, evaluation compute and frozen artifacts}
\label{sec:s-compute}

\paragraph{Training.} Actor training ran on H200 GPUs (openpi commit \texttt{\openpiCommit} with the locked post-training patch; Python~3.11.17, JAX~0.5.3, Flax~0.10.2, Optax~0.2.4, PyTorch~2.7.1) at global batch 64 for every trained arm, with one to four GPUs per arm: \DemoCont took 6.3~h on one GPU; \TakeoverBC, \IQLR and AWAC about 1.7~h each on four; \MixBC 2.0~h on four; and \mbox{IQL-R0} 3.2~h on two, about 41 GPU-hours in total. Fitting each 50,000-step critic took about 3~min on one H200 GPU. \BaseRestart, \mbox{First-of-32} and \mbox{IDQL-R2} require no actor training.

\paragraph{Evaluation and artifact mapping.} Each arm was evaluated on \nFailTest{}~$\times$~\samplesPerPoint restored (\texttt{test\_\allowbreak manifest.json} and \texttt{fpx\_\allowbreak eval.py}) and \nNominal nominal (\texttt{nominal\_\allowbreak configs.json}, \texttt{nominal\_\allowbreak validity.json} and \texttt{nominal\_\allowbreak extcheck.json} with \texttt{fpx\_\allowbreak nominal\_\allowbreak eval.py}) episodes, for \nEvalRows restored and \nNormRowsAll nominal episodes over the nine arms (\texttt{<arm>/rec.jsonl} and \texttt{<arm>/norm.jsonl}, scored by the frozen script \texttt{fpx\_\allowbreak analyze.py} under CPython~3.9--3.11 to produce Table~3 of the main paper). Every result shard and replay record logs the SHA-256 digests of the evaluator code, policy server and served checkpoint directory (\texttt{ckpt\_\allowbreak digest}, hashed recursively over \texttt{params/} and \texttt{assets/}: \texttt{79a141abcd51...} for \TakeoverBC, \texttt{5d05f18ec5a3...} for \IQLR, and \texttt{22284ad4723c...} for \BaseRestart, corresponding to the base-parameter archive digest \texttt{6a6c6aa1da11...} in \cref{sec:s-platform-base}). Rollout time, the wall-clock duration summed over all episodes, was \evalClientHours~h; episodes ran in parallel. \BaseRestart was evaluated on an RTX~5090 GPU and the other eight arms on H200 GPUs.

%% file: supp/posttrain_implementation.tex
\subsection{Value-based arms:\texorpdfstring{\\}{ }IQL-R2, IQL-R0 and AWAC}
\label{sec:s-value-arms}

\paragraph{Critic fitting.} \IQLR uses implicit Q-learning~\cite{kostrikov2022iql} to derive sampling weights before actor training. Its critic consists of twin $Q$ networks ($Q_1,Q_2$) and a state-value network $V$. Each network is a four-layer MLP with three 512-unit hidden layers (\texttt{Linear} $\rightarrow$ \texttt{LayerNorm} $\rightarrow$ \texttt{ReLU}, no dropout) and a linear scalar output. The state input $c\in\mathbb R^{2056}$ concatenates the z-score normalized 2,048-dimensional mean-pooled prefix embedding of the images and instruction from the frozen vision-language backbone with z-score normalized 8-dimensional proprioception (3-D end-effector position, 3-D axis-angle orientation, 2-D gripper joint positions); $Q_1$ and $Q_2$ additionally receive an 80-dimensional action input concatenating the z-score normalized $10\times7$ action chunk (positions past the episode end zeroed) and its 10-bit validity mask, with per-dimension means and standard deviations ($+10^{-6}$) computed over all critic frames. $V$ is fitted by expectile regression ($\tau=0.7$) toward $\min(Q_{1,\mathrm{tgt}},Q_{2,\mathrm{tgt}})$, and $Q_1,Q_2$ minimize the sum of their squared TD errors to $r+0.99(1-d)V(c')$, where $d$ marks termination and discounting applies once per chunk transition. Both optimizers are plain Adam (constant learning rate $3\times10^{-4}$, $\beta_1=0.9$, $\beta_2=0.999$, zero weight decay, no gradient clipping) with uniform transition batches of size 512, seed~42, Polyak target updates every step ($\tau_{\mathrm{target}}=0.005$) and 50,000 updates, a budget separate from $U$.

\paragraph{Critic data and rewards.} The critic is fitted on $\mathcal R$, $\mathcal F$ and $\mathcal N$ (\rpTrainFrames{} chunk transitions plus \nUnsuccEpsPost{} stored final observations). A chunk that reaches the end of an accepted recovery receives terminal reward $+1$; a failed policy prefix ends at the annotated interval midpoint with terminal reward $-1$. Unsuccessful human attempts are nonterminal truncations with zero reward that bootstrap from the recorded final observation. Intermediate chunks receive zero reward. Chunks that extend past the end of an episode are padded with the last action, zeroed in normalized coordinates beyond the mask and masked in the critic, whereas the actor trains on complete 10-step chunks only.

\paragraph{Actor eligibility and weights.} The eligible actor set $\mathcal E$ (\nEligChunksAll{} complete chunks) comprises all \nRecChunksPost{} complete recovery chunks ($\mathcal R$, \pctEligUnwRec{}\% of $\mathcal E$) and \nFwinChunksPost{} failure-window chunks ($\mathcal F_{\mathrm{win}}$, \pctEligUnwFwin{}\% of $\mathcal E$): failed-prefix chunks starting at frames $t$ with $\max(0,a-50)\leq t\leq m-10$, where $a$ is the start of the failure interval and $m$ its midpoint. The upper bound keeps each 10-step chunk within the recorded failure prefix, and unsuccessful human attempts contribute to critic fitting only. For chunk $i$ with critic input $c_i$, let $A_i=\min(Q_1,Q_2)(c_i,\mathbf a_i)-V(c_i)$, and let $\sigma_A$ be the standard deviation of $A_i$ over $\mathcal E$ (without mean subtraction; $\sigma_A=0.032$ for \IQLR). Following Eq.~(1) of the main paper ($\beta=1$), $A_i$ is divided by $\max(\sigma_A,10^{-3})$ and exponentiated, the weight $w_i$ is capped at 100, and chunk $i$ is drawn with replacement with probability $p_i=w_i/\sum_{j\in\mathcal E}w_j$, giving expected actor loss $\sum_{i\in\mathcal E}p_i\,\ell_{\mathrm{FM}}(\theta;x_i,\mathbf a_i)$ (\cref{eq:s-fm}). Under fixed weights, \pctDrawRecIQL{}\% of expected actor draws come from $\mathcal R$ and \pctDrawFwinIQL{}\% from $\mathcal F_{\mathrm{win}}$; the group-mean weight check is not met (\cref{sec:s-critic}).

\paragraph{Exploratory value-based arms.} \mbox{IQL-R0} fits the same IQL critic architecture and optimizer for 50,000 updates on $\mathcal R$ alone (\nTrainStepsPost{} transitions) and applies its weights to the \nRecChunksPost{} recovery chunks. AWAC~\cite{nair2020awac} samples the same eligible set $\mathcal E$ ($\mathcal R$ and $\mathcal F_{\mathrm{win}}$) as \IQLR{} and fits a one-step twin-$Q$ critic (same 512-wide MLP architecture, inputs, normalization, plain Adam at $3\times10^{-4}$, batch 512, Polyak update 0.005, discount 0.99 and 50,000 updates with seed~42) of the fixed \TakeoverBC{} checkpoint on the \rpTrainFrames{} transitions of $\mathcal R\cup\mathcal F\cup\mathcal N$. At every transition state and stored final observation, $K=8$ action chunks are pre-sampled from \TakeoverBC{} (10 Euler flow-matching steps, seed~42) and encoded with the row's validity mask; both the TD target $r+0.99(1-d)\,\frac{1}{8}\sum_{k=1}^8\min(Q_{1,\mathrm{tgt}},Q_{2,\mathrm{tgt}})(c',\hat{\mathbf a}'_k)$ and the advantage baseline $\frac{1}{8}\sum_{k=1}^8\min(Q_1,Q_2)(c_i,\hat{\mathbf a}_{ik})$ average over these eight chunks, followed by the same $\sigma_A$ standardization ($\sigma_A=0.082$, $\beta=1$) and cap at 100 (\pctDrawRecAWAC{}\% of actor draws from $\mathcal R$ and \pctDrawFwinAWAC{}\% from $\mathcal F_{\mathrm{win}}$). The AWAC recipe was fixed without tuning after held-out evaluation had begun and before any AWAC model was trained or evaluated.

\subsection{Inference-time variants:\texorpdfstring{\\}{ }First-of-32 and IDQL-R2}
\label{sec:s-inference-variants}

Both variants run the \TakeoverBC checkpoint without further training and change only how the executed chunk is chosen at each policy query. \mbox{IDQL-R2}~\cite{hansenestruch2023idql} samples 32 chunks from \TakeoverBC per query and executes the one maximizing $\min(Q_1,Q_2)$ under the \IQLR critic. \mbox{First-of-32} draws the same 32 samples and executes the first, so the two variants differ only in this selection. \Cref{sec:s-analyses} analyzes the reranking.

%% file: supp/app_d.tex
\section{Evaluation Protocol and\texorpdfstring{\\}{ }Statistical Procedures}
\label{sec:s-evaluation}

This section defines how all nine arms are evaluated, from restored held-out failure points (\cref{sec:s-restore}) and from nominal starts (\cref{sec:s-nominal-starts}), and the statistical procedures behind their decisions and intervals (\cref{sec:s-decision,sec:s-nominal-ci}).

\subsection{Restore-and-continue evaluation}
\label{sec:s-restore}

Every arm resumes from the same restored held-out states under the following protocol.
\begin{description}
\item[Held-out list.] \nFailTest points, fixed before the first held-out rollout: \nTestConfirm confirmatory (L1--L3, \nFamilies families) and \nTestDiag exploratory (L4--L5, \suppFamDiag families), with \nTestPerConfig points in every listed configuration.
\item[Failure point.] The restart state is frame $m=\mbox{$\lfloor(a+b)/2\rfloor$}$, the midpoint of the final interval $[a,b]$ rounded down, which differs from the exact midpoint for \nOddMidpoints of the \nFailTest intervals. Interval bounds and $m$ index the recorded simulator states, one per control step at \controlHz~Hz: frame $f$ is the state after $f$ environment steps from reset, counting the 10 zero-motion wait steps, so it follows base-policy step $f-10$; each rendered frame shows exactly that state.
\item[Restore.] Restoring re-applies the recorded fixture layout, sets the recorded MuJoCo state at $m$ and rebuilds the gripper-controller state from the first $m$ recorded actions. Before every rollout, the restored MuJoCo state is checked against the recorded one; all \nEvalRows recovery rollouts pass this check (maximum absolute difference \restoreMaxDiff).
\item[Policy state.] The policy starts with no queued actions and no observation history; the instruction is that of the original rollout, including L5 paraphrases.
\item[Success.] Defined by the LIBERO-X task predicates.
\item[Horizon.] Success counts within \horizonMain control steps (primary), a horizon that covers the length of \pctWithinTwelveHundred{}\% of accepted recoveries, and within \horizonShort and \horizonLong steps (sensitivity analyses); all three are read from one rollout of up to \horizonLong steps.
\item[Samples.] $K=$~\samplesPerPoint policy-sampling seeds per point, shared across arms. Every started rollout is scored, so denominators are fixed: restore errors, physics-guard faults and execution errors count as failures.
\item[Metrics.] \SRrec{} (Eq.~(2) of the main paper) for success rates and the family scores of \cref{eq:s-family-score} for contrasts; every difference between arms, including the improvement over \BaseRestart, is paired over the same restored points.
\item[Completeness.] Results come only from complete evaluation runs; interrupted runs were repeated in full.
\end{description}

\subsection{Nominal starts, validity and horizons}
\label{sec:s-nominal-starts}

Nominal evaluation measures how much task performance each arm retains when it starts from an initial state rather than from a failure; episodes follow the \mbox{LIBERO-X} evaluation procedure with seeded starts.
\begin{description}
\item[Starts.] \nNominal seeded initial states per arm, an equal number in each of the \nConfigs configurations, with seeds shared across arms. After the zero-motion wait steps the policy runs for up to \horizonMain steps; at L5 the instruction cycles through the level's paraphrases. Physics-guard faults count as failures.
\item[Validity.] \nNormValid of the \nNominal starts are valid under a policy-independent check that was fixed before nominal policy evaluation and is shared by all arms; all nominal rates and changes use these \nNormValid starts as their denominator. The check applies the start-validity rule of candidate screening (\cref{sec:s-funnel}) to the settled start after the wait steps, with each object's settling baseline taken over the starts of its configuration; a start is also invalid if the physics guard fires during the wait.
\item[Sensitivity set.] Some valid starts ended in a physics-guard fault within the first policy steps for every arm. A policy-free check that replays each start exactly and appends 20 no-op steps flags \nNormExcl of the \nNormValid valid starts as unstable; removing them gives the \nNormSens-start sensitivity set. No arm succeeds on a flagged start, so every numerator is unchanged on this set; the reported rates keep the \nNormValid-start denominator.
\item[Horizons.] Nominal success is reported within \horizonMain steps, the horizon of the recovery evaluation, and within \timeGateScreen{}$T$, the pre-specified task-dependent criterion, where $T$ is the mean demonstration length (\cref{sec:s-families-levels}). Rates and changes refer to \horizonMain steps unless noted; this horizon was adopted for nominal evaluation after nominal results of several arms had been seen.
\item[Change.] \dSRnorm{} is the nominal success of an arm minus that of \BaseRestart, paired over the same valid starts and pooled over them. All nominal analyses are exploratory; their intervals follow \cref{sec:s-nominal-ci}.
\end{description}

\subsection{Scores, paired bootstrap and decision rules}
\label{sec:s-decision}

Recovery scores average rollouts within points, points within family--level configurations, levels within families, and then families. For arm $\pi$, family $f$ and set of levels $\mathcal L$, with $\mathcal P_{f\ell}$ the held-out points of family $f$ at level $\ell$, and $K$, $H$ and $y^H_{pk}(\pi)$ as in Eq.~(2) of the main paper,
\begin{equation}
\begin{aligned}
R_f(\pi)&=\frac{1}{|\mathcal L|}\sum_{\ell\in\mathcal L}\frac{1}{|\mathcal P_{f\ell}|}\sum_{p\in\mathcal P_{f\ell}}\frac{1}{K}\sum_{k=1}^{K}y^H_{pk}(\pi),\\
R(\pi)&=\frac{1}{n}\sum_{f=1}^{n}R_f(\pi),\\
\Delta&=R(\pi)-R(\pi')=\frac{1}{n}\sum_{f=1}^{n}d_f,
\end{aligned}
\label{eq:s-family-score}
\end{equation}
where $d_f=R_f(\pi)-R_f(\pi')$ is the paired difference of family $f$ between arms $\pi$ and $\pi'$, and $\mathcal L=\{\mathrm{L1},\mathrm{L2},\mathrm{L3}\}$ with $n$ = \nFamilies for the confirmatory set ($\{\mathrm{L4},\mathrm{L5}\}$ with $n$ = \suppFamDiag for the exploratory one). Because every listed configuration has \nTestPerConfig points, $R(\pi)$ equals the pooled \SRrec{} of the same points.
\begin{description}
\item[Unit.] The task family: each family contributes one paired difference $d_f$ ($n$ = \nFamilies for L1--L3).
\item[Interval.] $B=10{,}000$ bootstrap resamples, each drawing $n$ families with replacement. The BCa (bias-corrected and accelerated) 95\% interval~\cite{efron1987bca}, with its acceleration estimated by a jackknife over families, is primary; the percentile interval is a sensitivity check and the fallback when the BCa calculation is degenerate. A single arm's score $R(\pi)$ receives a 95\% percentile interval from the same family resampling. The two-sided bootstrap tail probability (twice the smaller of the shares of resampled estimates at or below and at or above zero, capped at 1) enters no decision. All intervals condition on the evaluated checkpoints and exclude training-seed variation.
\item[Rule.] Q1 passes if its BCa interval lies above 0 and its estimate is at least $+$\effectThreshold{}~percentage points (pp); Q2 passes if its BCa interval excludes 0 and its estimate is at least \effectThreshold{}~pp in magnitude. A contrast that does not pass is reported as \emph{not~met}.
\item[Order.] Q1 (\TakeoverBC{} minus \DemoCont) is tested first, and Q2 (\IQLR{} minus \TakeoverBC) is confirmatory only if Q1 passes. All other analyses are exploratory; the other L1--L3 contrasts use the same paired family bootstrap without a decision.
\item[Equivalence.] The pre-specified analysis includes no equivalence test: an interval within $\pm$\effectThreshold~pp is not read as equivalence, and a contrast that is not met is not evidence of equivalence.
\item[Power.] With $n$ = \nFamilies families, an observed mean paired difference at the \effectThreshold~pp threshold yields an interval that excludes zero only if the across-family SD of the paired differences is below $5\sqrt{10}/t_{0.975,9}\approx7.0$~pp ($t$ approximation; $\approx8.1$~pp under a normal one), which corresponds to about 50\% power for a true \effectThreshold~pp effect. For 80\% power, the SD must be about $5\sqrt{10}/(1.96+0.84)\approx5.6$~pp or less (normal approximation; smaller under $t$).
\end{description}

\subsection{Interval conventions for nominal changes}
\label{sec:s-nominal-ci}

Nominal changes pool paired start-level differences over the \nNormValid valid starts, so their intervals resample clusters of starts rather than the family scores of \cref{eq:s-family-score}. The main convention, used for the nominal intervals in Sec.~5 and Table~3 of the main paper, resamples the \nConfigs family--level configurations 2,000 times and takes the 95\% percentile interval. Resampling the \nFamilies task families 10,000 times instead serves only as a sensitivity analysis, with the same point estimates and denominators (\cref{tab:s-nominal-clusters}). Under both conventions the interval of \DemoCont contains zero and those of \TakeoverBC, \MixBC and \IQLR lie below zero.

\input{supp/tab_nominal_sensitivity}

%% file: supp/tab_nominal_sensitivity.tex
\begin{table}[t]
\centering\footnotesize\setlength{\tabcolsep}{4pt}\tabstyle
\caption{\textbf{Both interval conventions exclude zero for the same arms.} \dSRnorm: nominal success of the arm minus that of \BaseRestart{} within \horizonMain{} steps, paired over the same \nNormValid{} valid starts, in percentage points (pp); exploratory. Both interval columns are 95\% percentile bootstrap intervals around the same point estimate. Configuration: the \nConfigs{} family--level configurations resampled 2,000 times (main convention). Family: the \nFamilies{} task families resampled 10,000 times (sensitivity analysis only).}
\label{tab:s-nominal-clusters}
\tabzebra
\begin{tabular}{lrcc}
\toprule
\tabhead & & \multicolumn{2}{c}{95\% interval (pp)} \\
\cmidrule(l){3-4}
\tabhead Arm & \dSRnorm{} (pp) & Configuration & Family \\
\midrule
\DemoCont & \dnormDemo{} & [\secFiveNomLoDemo, +\secFiveNomHiDemo{}] & [\dnormLoDemo, \dnormHiDemo{}] \\
\TakeoverBC & \dnormTB{} & [\secFiveNomLoTB, \secFiveNomHiTB{}] & [\dnormLoTB, \dnormHiTB{}] \\
\MixBC & \dnormMix{} & [\secFiveNomLoMix, \secFiveNomHiMix{}] & [\dnormLoMix, \dnormHiMix{}] \\
\IQLR & \dnormIQL{} & [\secFiveNomLoIQL, \secFiveNomHiIQL{}] & [\dnormLoIQL, \dnormHiIQL{}] \\
\bottomrule
\end{tabular}
\end{table}

%% file: supp/app_e.tex
\section{Complete Results}
\label{sec:s-results}
\label{sec:s-pending}

This section reports results for all nine evaluated arms beyond the summary in Table~3 of the main paper: recovery across horizons (\cref{sec:s-res-arms}), the robustness of the Q1 and Q2 decisions (\cref{sec:s-res-robustness}), and results by task family, difficulty level and failure type and on the nominal starts (\cref{sec:s-res-families,sec:s-res-levels,sec:s-res-ftype,sec:s-res-nominal}). Arms are defined in \cref{sec:s-posttrain} and metrics and intervals in \cref{sec:s-evaluation}; all tables are computed from the same evaluation rollouts as the main paper, and values without an interval are point estimates.

\subsection{Recovery across horizons}
\label{sec:s-res-arms}

Table~3 of the main paper gives every arm's L1--L3 recovery at \horizonShort, \horizonMain{} and \horizonLong{} steps, and the main paper plots recovery against the horizon for five arms; \cref{tab:s-res-budget} tabulates these curves for all nine arms at horizons from 100 to \horizonLong{} steps. At every horizon in the table, every takeover-trained arm is above both \BaseRestart{} and \DemoCont. \mbox{IDQL-R2} is ahead of \TakeoverBC{} at 100 and 200 steps but behind it from 400 steps on (\cref{sec:s-rerank}).

\suppressfloats[t]
\begin{table}[t]
\centering\footnotesize\tabstyle
\caption{\textbf{Recovery across horizons (descriptive).} L1--L3 \SRrec{} (\%) within each horizon on the \nTestConfirm{} L1--L3 held-out failures (\samplesPerPoint{} rollouts each), read from the same rollouts as Table~3 of the main paper; the \horizonShort-, \horizonMain- and \horizonLong-step columns repeat its values. Point estimates; $^{\dagger}$exploratory arms.}
\label{tab:s-res-budget}
\input{supp/tab_e_budget}
\end{table}

\subsection{Robustness of the Q1 and Q2 decisions}
\label{sec:s-res-robustness}

The two pre-specified questions are decided on the \nTestConfirm{} L1--L3 held-out failures with the family-paired bootstrap of \cref{sec:s-evaluation}; the checks below vary the interval type, the horizon, the set of families and the handling of faults.
\begin{description}
\item[Primary horizon.] At \horizonMain{} steps, Q1 is +\DeltaOne{}~pp with the 95\% BCa interval [+\DeltaOneLo, +\DeltaOneHi{}] and passes; Q2 is +\DeltaTwo{}~pp [\DeltaTwoLo, +\DeltaTwoHi{}] and is not met. The percentile intervals are [+\DeltaOnePctLo, +\DeltaOnePctHi{}] for Q1 and [\DeltaTwoPctLo, +\DeltaTwoPctHi{}] for Q2, and the two-sided bootstrap tail probabilities are \DeltaOneP{} for Q1 and \DeltaTwoP{} for Q2 (unadjusted; they enter no decision).
\item[Other horizons.] At \horizonShort{} steps, Q1 is \DeltaOneSix{}~pp [\DeltaOneSixLo, \DeltaOneSixHi{}] and Q2 is \DeltaTwoSix{}~pp [\DeltaTwoSixLo, \DeltaTwoSixHi{}]; at \horizonLong{} steps, Q1 is \DeltaOneEighteen{}~pp [\DeltaOneEighteenLo, \DeltaOneEighteenHi{}] and Q2 is \DeltaTwoEighteen{}~pp [\DeltaTwoEighteenLo, \DeltaTwoEighteenHi{}] (95\% BCa). At both horizons Q1 passes and Q2 is not met; at \horizonLong{} steps the Q2 interval lies within $\pm$\effectThreshold{}~pp, which the protocol does not read as equivalence (\cref{sec:s-evaluation}).
\item[Leave one family out.] Rerunning the decision procedure without one family at a time gives Q1 between \lofoOneMin{} and \lofoOneMax{}~pp, with every BCa lower bound at or above \lofoOneLoMin{}~pp, and Q2 between \lofoTwoMin{} and \lofoTwoMax{}~pp, never passing; the Q2 interval lies within $\pm$\effectThreshold{}~pp when any one of families \lofoTwoInsideFams{} is left out. No decision changes.
\item[Per-family differences.] Of the \nFamilies{} per-family Q1 differences, \nFamPosOne{} are positive, \nFamTieOne{} tied and \nFamNegOne{} negative; of the Q2 differences, \nFamPosTwo{} are positive, \nFamTieTwo{} tied and \nFamNegTwo{} negative (\cref{tab:s-res-families}).
\item[Faults.] Physics-guard faults count as failures (counts per arm in \cref{tab:s-res-levels}). Excluding every pair of rollouts (same point and sample) in which either arm ended in a fault changes no pooled paired difference by more than \faultSensMaxShift{}~pp; this covers the Q1 and Q2 pairs, \TakeoverBC{} and \DemoCont{} against \BaseRestart, and every other arm against \TakeoverBC, pooled over all \nRecRowsArm{} rollouts per arm.
\item[Curation and interval flags.] Linking the recorded quality flags (\cref{sec:s-flags}) to the held-out point IDs, \cref{tab:s-curation-sens} repeats the L1--L3 hierarchical BCa procedure (with a separate post-hoc bootstrap stream) after dropping the \nFlagWidthLo{} L1--L3 width-flagged intervals (\suppWidthFlag{} across L1--L5), the eight L1--L3 intervals outside the strict 10--40-frame target, the \suppLateTestConf{} L1--L3 held-out late successes (\suppLateTest{} across L1--L5), the two coordinator-revised held-out intervals, and the union of these flags (\nFlagUnionLo{} L1--L3 points dropped, \nRetainedUnionLo{} retained: Q1 is \qOneUnionEst{}~pp [\qOneUnionLo, \qOneUnionHi{}], Q2 is \qTwoUnionEst{}~pp [\qTwoUnionLo, \qTwoUnionHi{}]). Both primary decisions hold in every subset. These checks audit sensitivity to recorded flags rather than multi-annotator agreement or alternative restart frames inside $[a,b]$.
\item[Base recoverability.] Grouping held-out points by how many of the \samplesPerPoint{} \BaseRestart{} rollouts succeed within \horizonMain{} steps (\cref{tab:s-base-strata}), \nBaseZeroLoPts{} of the \nTestConfirm{} L1--L3 points have $0/3$ \BaseRestart{} successes: \DemoCont{} reaches \srZeroLoDemo{}\% on them, whereas \TakeoverBC{} and \IQLR{} reach \srZeroLoTB{}\% and \srZeroLoIQL{}\% (\srZeroAllTB{}\% for \TakeoverBC{} across the \nBaseZeroAllPts{} $0/3$ points of L1--L5). On the \nBaseAllLoPts{} L1--L3 points where \BaseRestart{} succeeds in $3/3$ rollouts, post-trained arms score 75.0--86.1\%, matching the retention cost seen on nominal starts.
\end{description}

\begin{table*}[t]
\centering\footnotesize\setlength{\tabcolsep}{5pt}\tabstyle
\caption{\textbf{Sensitivity of L1--L3 confirmatory contrasts to recorded held-out curation and interval flags (post hoc).} Hierarchical paired differences (pp) and 95\% family-level BCa intervals (10,000 resamples) when flagged held-out points (\cref{sec:s-flags}) are excluded while keeping the configuration-to-family hierarchy. All ten families and all 30 L1--L3 configurations remain non-empty in every row.}
\label{tab:s-curation-sens}
\input{supp/tab_f_curation}
\end{table*}

\begin{table*}[t]
\centering\footnotesize\setlength{\tabcolsep}{5pt}\tabstyle
\caption{\textbf{Held-out recovery stratified by \BaseRestart{} point-level recoverability (descriptive).} Pooled \SRrec{} (\%) within \horizonMain{} steps across points grouped by how many of the \samplesPerPoint{} \BaseRestart{} rollouts succeed at that point (\samplesPerPoint{} rollouts per point and arm). $^{\dagger}$exploratory arms.}
\label{tab:s-base-strata}
\input{supp/tab_f_strata}
\end{table*}

\subsection{Results by task family}
\label{sec:s-res-families}

\Cref{tab:s-res-families} breaks recovery, the Q1 and Q2 differences and nominal success down by task family for all nine arms; the main paper shows the per-family L1--L3 recovery of five arms and the nominal change of \TakeoverBC. The per-family Q1 differences are largest in families 5 and 14 and near zero in families 9 and 12, the per-family Q2 differences take both signs, and \TakeoverBC{} loses nominal success in every family. These patterns are descriptive: every family is a different task with its own mix of failure types.

\begin{table*}[t]
\centering\footnotesize\tabstyle
\caption{\textbf{Recovery and nominal success by task family (descriptive).} (a)~L1--L3 \SRrec{} (\%) within \horizonMain{} steps, with the per-family differences of Q1 (\TakeoverBC{} $-$ \DemoCont) and Q2 (\IQLR{} $-$ \TakeoverBC) in percentage points (pp); (b)~L4--L5 \SRrec{} (\%); (c)~nominal success within \horizonMain{} steps: \BaseRestart{} in \%, every other arm as the paired change relative to \BaseRestart{} in pp. $n$: rollouts (a, b) or valid starts (c) per arm. All: mean over families, as in Table~3 of the main paper (a, b), or pooled over the valid starts (c). --: no L4--L5 held-out point in family~9. Point estimates; $^{\dagger}$exploratory arms. Task families: 0~black book $\rightarrow$ back caddy; 4~striped book $\rightarrow$ left caddy; 5~two black books upright; 6~yellow book + red mug $\rightarrow$ shelf; 7~black book $\rightarrow$ left caddy; 9~wine bottle $\rightarrow$ rack, close drawer; 12~red mug + yellow book $\rightarrow$ caddy; 14~open middle drawer; 15~striped book + red mug $\rightarrow$ shelf; 17~porcelain mug $\rightarrow$ middle plate.}
\label{tab:s-res-families}
\input{supp/tab_e_families}
\end{table*}

\subsection{Results by difficulty level}
\label{sec:s-res-levels}

\Cref{tab:s-res-levels} gives recovery and nominal success per level for all nine arms, together with the physics-guard faults of each arm and its nominal change relative to \BaseRestart{} within the pre-specified \timeGateScreen{}$T$; the main paper plots five of the arms by level, and its Table~3 gives the pooled \horizonMain-step changes with their intervals. Recovery falls from L1 to L5 for every arm, and every takeover-trained arm loses the most nominal success at L1, where \BaseRestart{} is strongest. L4--L5 recovery has no family-9 points (\cref{tab:s-res-families}).

\begin{table*}[t]
\centering\footnotesize\tabstyle
\caption{\textbf{Recovery and nominal success by level (descriptive).} \SRrec{} (\%) and nominal success (\%) within \horizonMain{} steps per level; $n$: rollouts or valid starts per arm; All: pooled over levels. Faults: recovery rollouts (of the \nRecRowsArm{} per arm) that ended in a physics-guard fault, counted as failures. \dSRnorm{} @\timeGateScreen{}$T$: paired change in nominal success relative to \BaseRestart{} within the pre-specified \timeGateScreen{}$T$ ($T$: mean demonstration length, \cref{sec:s-families-levels}), in percentage points (pp). --: reference arm. Point estimates; L4--L5 and all nominal results are exploratory; $^{\dagger}$exploratory arms.}
\label{tab:s-res-levels}
\input{supp/tab_e_levels}
\end{table*}

\subsection{Results by failure type}
\label{sec:s-res-ftype}

\Cref{tab:s-ftype} splits recovery at \horizonMain{} steps by the failure type of the held-out point, over all levels and separately for L1--L3 and L4--L5. Over all levels, \TakeoverBC{} gains most over \BaseRestart{} and \DemoCont{} on collisions and dropped objects and falls below both only on unclosed drawers, all of which are held out at L1--L3. Failure type is not balanced across task families or levels: most types are concentrated in one or two families, and the mix of types differs by level (\cref{tab:s-ftype-level}). Differences between types therefore also reflect differences between tasks and levels, and the table is descriptive.

\begin{table*}[t]
\centering\footnotesize\tabstyle
\caption{\textbf{Recovery by failure type of the held-out point (descriptive).} \SRrec{} (\%) within \horizonMain{} steps over all levels, on L1--L3 and on L4--L5; $n$: rollouts per arm (held-out points of the type $\times$ \samplesPerPoint). No drawer-not-closed failure is held out at L4--L5. Point estimates; $^{\dagger}$exploratory arms. Failure type is not balanced across families or levels (\cref{sec:s-res-ftype}).}
\label{tab:s-ftype}
\input{supp/tab_e_ftype}
\end{table*}

\subsection{Nominal-start details}
\label{sec:s-res-nominal}

Nominal results are exploratory and follow the definitions of \cref{sec:s-evaluation}.
\begin{description}
\item[Starts.] \nNormValid{} of the \nNominal{} nominal starts per arm are valid, and the valid starts are the same for every arm; all nominal rates and changes use them as the denominator. The \nNormSens-start sensitivity set drops \nNormExcl{} policy-independent unstable starts on which no arm succeeds, so every numerator is unchanged on it.
\item[Intervals.] The main nominal intervals come from the configuration bootstrap over the \nConfigs{} family--level configurations (Table~3 of the main paper); the family bootstrap over the \nFamilies{} task families is a sensitivity analysis only (\cref{tab:s-nominal-clusters}).
\item[Horizon.] Within the stricter \timeGateScreen{}$T$ horizon, \BaseRestart{} succeeds on \normAbsTBase{}\% of the valid starts and no takeover-trained arm exceeds the \normAbsTMix{}\% of \MixBC, so the nominal advantage of \MixBC{} over \TakeoverBC{} shrinks from \secFiveMixGainMain{} to \secFiveMixGainOriginal{}~pp (rates within \timeGateScreen{}$T$ in Table~3 of the main paper, changes in \cref{tab:s-res-levels}).
\item[Pace.] On the \nNormSharedTB{} valid starts that both \TakeoverBC{} and \BaseRestart{} complete within \horizonMain{} steps, \TakeoverBC{} needs more steps on \nNormSlowerTB, and the median paired difference over all \nNormSharedTB{} starts is \medNormSlowerTB{} steps (\nNormSlowerTBDemo{} of \nNormSharedTBDemo{} starts and a median of \medNormSlowerTBDemo{} steps against \DemoCont{} on the same H200 GPUs). The comparison conditions on joint success and does not decompose the nominal loss.
\item[Outcomes, completion curves and same-GPU contrasts.] \Cref{tab:s-nominal-decomp} partitions all \nNormValid{} valid starts into mutually exclusive outcomes, gives cumulative completion at 400, 800 and \horizonMain{} steps and the \nNormSens-start stable-set rate (\nomSensSRBase{}\% for \BaseRestart, \nomSensSRDemo{}\% for \DemoCont, \nomSensSRTB{}\% for \TakeoverBC, \nomSensSRMix{}\% for \MixBC{} and \nomSensSRIQL{}\% for \IQLR), and reports paired nominal changes relative to \DemoCont{} (all evaluated on H200 GPUs). Under \TakeoverBC, \nomLateNTB{} of its \nomSuccNTB{} nominal successes (\nomLatePctTB{}\% of valid starts, \vs \nomFastNTB{} within \timeGateScreen{}$T$) finish in $(\timeGateScreen{}T,\horizonMain]$ with a median completion step of \nomMedRatioTTB{}$T$ (\vs \nomMedRatioTBase{}$T$ for \BaseRestart{} and \nomMedRatioTDemo{}$T$ for \DemoCont), while \nomTimeoutPctTB{}\% of valid starts reach step \horizonMain{} without success or fault (\vs \nomTimeoutPctBase{}\% for \BaseRestart{} and \nomTimeoutPctDemo{}\% for \DemoCont) and faults on the \nNormSens{} stable starts remain rare (\nomFaultStableNTB{} \vs \nomFaultStableNBase). Under \MixBC, \nomLateNMix{} of \nomSuccNMix{} successes (\nomLatePctMix{}\% of valid starts; median \nomMedRatioTMix{}$T$) fall in $(\timeGateScreen{}T,\horizonMain]$, accounting for why its advantage over \TakeoverBC{} is \secFiveMixGainMain{}~pp at \horizonMain{} steps and \secFiveMixGainOriginal{}~pp within \timeGateScreen{}$T$. Relative to H200-evaluated \DemoCont, \horizonMain-step nominal success changes by \dnormVsDemoTB{}~pp [\dnormVsDemoLoTB, \dnormVsDemoHiTB{}] for \TakeoverBC, \dnormVsDemoMix{}~pp [\dnormVsDemoLoMix, \dnormVsDemoHiMix{}] for \MixBC{} and \dnormVsDemoIQL{}~pp [\dnormVsDemoLoIQL, \dnormVsDemoHiIQL{}] for \IQLR, confirming that the retention losses do not stem from the RTX~5090 evaluation of \BaseRestart. Because nominal evaluation stopped at \horizonMain{} steps without logging per-step action traces, \horizonMain-step timeouts combine unfinished slow executions (\cref{sec:s-command-scale}) with trajectories that stall or err before the goal; \dSRnorm{} therefore measures horizon-specific completion under the coupled SpaceMouse scale and training recipe rather than pure skill forgetting.
\item[Families.] \TakeoverBC{} has no nominal success within \horizonMain{} steps in \nFamNormZero{} families (\famNormZero; \cref{tab:s-res-families}).
\end{description}

\begin{table*}[t]
\centering\scriptsize\setlength{\tabcolsep}{3.5pt}\tabstyle
\caption{\textbf{Nominal-start outcome partition, cumulative completion and same-hardware contrasts (exploratory).} Outcomes over the \nNormValid{} valid starts sum to 100\% per arm: success within \timeGateScreen{}$T$, late success in $(\timeGateScreen{}T,\horizonMain]$, \horizonMain-step timeout without fault, and physics-guard fault (of which \nNormExcl{} starts, 11.7\% of \nNormValid, fault in the policy-free 20-step check for every arm). SR @\nNormSens: \horizonMain-step success on the \nNormSens{} stable valid starts. Cumulative SR: success within 400, 800 and \horizonMain{} steps over the \nNormValid{} valid starts. vs.\ \DemoCont{} (H200): paired change relative to \DemoCont{} (both evaluated on H200 GPUs, except RTX~5090 \BaseRestart), with 50-configuration 95\% percentile intervals (2,000 resamples) at \horizonMain{} steps and point estimates within \timeGateScreen{}$T$. $^{\dagger}$exploratory arms.}
\label{tab:s-nominal-decomp}
\input{supp/tab_f_nominal_decomp}
\end{table*}

%% file: supp/tab_e_budget.tex
\setlength{\tabcolsep}{0pt}
\tabzebra
\begin{tabular}{>{\hspace{4pt}}l<{\hspace{4pt}}*{7}{>{\hspace{4pt}}r<{\hspace{4pt}}}}
\toprule
\tabhead & \multicolumn{7}{c}{Horizon (steps)} \\
\cmidrule(l){2-8}
\tabhead Arm & 100 & 200 & 400 & \horizonShort & 800 & \horizonMain & \horizonLong \\
\midrule
\BaseRestart & 11.1 & 13.6 & 16.1 & 19.6 & 20.7 & 22.2 & 23.2 \\
\DemoCont & 10.4 & 13.1 & 17.2 & 19.4 & 21.1 & 22.5 & 23.3 \\
\TakeoverBC & 15.6 & 29.0 & 40.3 & 46.1 & 49.4 & 51.9 & 54.0 \\
Mix-BC$^{\dagger}$ & 16.1 & 25.6 & 32.8 & 38.8 & 42.6 & 45.3 & 46.7 \\
\IQLR & 19.4 & 34.3 & 42.4 & 48.8 & 51.8 & 53.5 & 54.6 \\
\midrule
IQL-R0$^{\dagger}$ & 17.1 & 29.3 & 38.8 & 44.2 & 47.5 & 48.9 & 50.3 \\
AWAC$^{\dagger}$ & 19.7 & 33.5 & 41.4 & 48.8 & 51.1 & 53.8 & 55.8 \\
First-of-32$^{\dagger}$ & 15.4 & 29.3 & 40.3 & 46.4 & 49.9 & 52.1 & 53.3 \\
IDQL-R2$^{\dagger}$ & 26.4 & 30.7 & 37.5 & 38.9 & 39.9 & 40.7 & 41.5 \\
\bottomrule
\end{tabular}

%% file: supp/tab_f_curation.tex
\tabzebra
\begin{tabular}{l r cc c}
\toprule
\tabhead Held-out L1--L3 subset & Points & Q1 (pp) [95\% BCa] & Q2 (pp) [95\% BCa] & Q1 / Q2 \\
\midrule
All L1--L3 confirmatory points (primary) & 240 & +\DeltaOne{} [+\DeltaOneLo, +\DeltaOneHi{}] & +\DeltaTwo{} [\DeltaTwoLo, +\DeltaTwoHi{}] & pass / not met \\
Exclude interval-width flags ($b-a+1<10$ or ${>}41$) & 234 & +29.3 [+16.8, +43.9] & +2.1 [\ensuremath{-}1.9, +6.0] & pass / not met \\
Exclude strict width range ($b-a+1<10$ or ${>}40$) & 232 & +29.2 [+16.7, +43.9] & +1.9 [\ensuremath{-}2.2, +5.9] & pass / not met \\
Exclude held-out late successes (succeeded after $1.5T$) & 231 & +31.2 [+17.9, +45.9] & +2.0 [\ensuremath{-}1.7, +5.4] & pass / not met \\
Exclude coordinator-revised held-out intervals & 238 & +29.6 [+16.9, +44.7] & +1.6 [\ensuremath{-}2.1, +5.1] & pass / not met \\
Exclude union of width, late-success and revised flags & 223 & +31.4 [+18.0, +46.4] & +2.7 [\ensuremath{-}1.3, +6.3] & pass / not met \\
\bottomrule
\end{tabular}

%% file: supp/tab_f_strata.tex
{\renewcommand{\tabgroup}[2]{\rowcolor{TabGroup}\multicolumn{#1}{l}{\rule{0pt}{2.2ex}\textbf{#2}}\\}
\begin{tabular}{l r rrrrrr}
\toprule
\tabhead \BaseRestart{} stratum & Points & \BaseRestart & \DemoCont & \TakeoverBC & Mix-BC$^{\dagger}$ & \IQLR & AWAC$^{\dagger}$ \\
\midrule
\tabgroup{8}{(a) Confirmatory L1--L3 held-out points (\nTestConfirm{} points)}
$0/3$ rollouts succeed & 154 & 0.0 & 3.7 & 46.8 & 37.4 & 47.4 & 48.5 \\
$1\text{--}2/3$ rollouts succeed & 62 & 47.3 & 44.6 & 54.3 & 52.2 & 58.1 & 57.0 \\
$3/3$ rollouts succeed & 24 & 100.0 & 86.1 & 79.2 & 77.8 & 80.6 & 79.2 \\
\midrule
\tabgroup{8}{(b) All L1--L5 held-out points (\nFailTest{} points; pooled)}
$0/3$ rollouts succeed & 276 & 0.0 & 3.0 & 40.5 & 31.3 & 39.7 & 39.4 \\
$1\text{--}2/3$ rollouts succeed & 80 & 45.4 & 38.3 & 55.4 & 48.8 & 56.2 & 53.8 \\
$3/3$ rollouts succeed & 28 & 100.0 & 84.5 & 77.4 & 75.0 & 78.6 & 76.2 \\
\bottomrule
\end{tabular}}

%% file: supp/tab_e_families.tex
\setlength{\tabcolsep}{0pt}
\begin{tabular}{>{\hspace{6pt}}l<{\hspace{6pt}}*{11}{>{\hspace{6pt}}r<{\hspace{6pt}}}}
\toprule
\tabhead & \multicolumn{10}{c}{Task family} & \\
\cmidrule(lr){2-11}
\tabhead Arm & 0 & 4 & 5 & 6 & 7 & 9 & 12 & 14 & 15 & 17 & All \\
\midrule
\tabgroup{12}{(a)~L1--L3 \SRrec{} (\%)}
$n$ & 72 & 72 & 72 & 72 & 72 & 72 & 72 & 72 & 72 & 72 & 720 \\
\BaseRestart & 29.2 & 31.9 & 13.9 & 13.9 & 37.5 & 8.3 & 18.1 & 19.4 & 13.9 & 36.1 & 22.2 \\
\DemoCont & 27.8 & 27.8 & 12.5 & 16.7 & 38.9 & 12.5 & 25.0 & 16.7 & 16.7 & 30.6 & 22.5 \\
\TakeoverBC & 44.4 & 51.4 & 86.1 & 44.4 & 72.2 & 11.1 & 25.0 & 73.6 & 58.3 & 52.8 & 51.9 \\
Mix-BC$^{\dagger}$ & 38.9 & 45.8 & 77.8 & 37.5 & 51.4 & 8.3 & 26.4 & 75.0 & 45.8 & 45.8 & 45.3 \\
\IQLR & 54.2 & 45.8 & 90.3 & 37.5 & 80.6 & 9.7 & 30.6 & 73.6 & 52.8 & 59.7 & 53.5 \\
IQL-R0$^{\dagger}$ & 52.8 & 43.1 & 83.3 & 38.9 & 55.6 & 16.7 & 26.4 & 68.1 & 50.0 & 54.2 & 48.9 \\
AWAC$^{\dagger}$ & 66.7 & 48.6 & 80.6 & 30.6 & 77.8 & 13.9 & 31.9 & 63.9 & 52.8 & 70.8 & 53.8 \\
First-of-32$^{\dagger}$ & 43.1 & 55.6 & 87.5 & 45.8 & 70.8 & 12.5 & 29.2 & 77.8 & 51.4 & 47.2 & 52.1 \\
IDQL-R2$^{\dagger}$ & 73.6 & 34.7 & 94.4 & 20.8 & 55.6 & 9.7 & 8.3 & 54.2 & 27.8 & 27.8 & 40.7 \\
\cmidrule(lr){1-12}
Q1 (pp) & +16.7 & +23.6 & +73.6 & +27.8 & +33.3 & \ensuremath{-}1.4 & 0.0 & +56.9 & +41.7 & +22.2 & +29.4 \\
Q2 (pp) & +9.7 & \ensuremath{-}5.6 & +4.2 & \ensuremath{-}6.9 & +8.3 & \ensuremath{-}1.4 & +5.6 & 0.0 & \ensuremath{-}5.6 & +6.9 & +1.5 \\
\midrule
\tabgroup{12}{(b)~L4--L5 \SRrec{} (\%)}
$n$ & 48 & 48 & 48 & 48 & 48 & -- & 48 & 48 & 48 & 48 & 432 \\
\BaseRestart & 6.2 & 29.2 & 2.1 & 0.0 & 10.4 & -- & 0.0 & 6.2 & 0.0 & 14.6 & 7.6 \\
\DemoCont & 8.3 & 20.8 & 4.2 & 0.0 & 12.5 & -- & 0.0 & 2.1 & 0.0 & 6.2 & 6.0 \\
\TakeoverBC & 14.6 & 35.4 & 47.9 & 27.1 & 25.0 & -- & 20.8 & 85.4 & 33.3 & 41.7 & 36.8 \\
Mix-BC$^{\dagger}$ & 14.6 & 16.7 & 52.1 & 12.5 & 16.7 & -- & 10.4 & 85.4 & 10.4 & 16.7 & 26.2 \\
\IQLR & 14.6 & 25.0 & 58.3 & 18.8 & 27.1 & -- & 16.7 & 83.3 & 22.9 & 35.4 & 33.6 \\
IQL-R0$^{\dagger}$ & 18.8 & 25.0 & 43.8 & 22.9 & 22.9 & -- & 22.9 & 85.4 & 27.1 & 25.0 & 32.6 \\
AWAC$^{\dagger}$ & 20.8 & 25.0 & 37.5 & 27.1 & 27.1 & -- & 12.5 & 64.6 & 27.1 & 33.3 & 30.6 \\
First-of-32$^{\dagger}$ & 14.6 & 31.2 & 52.1 & 27.1 & 25.0 & -- & 25.0 & 85.4 & 29.2 & 37.5 & 36.3 \\
IDQL-R2$^{\dagger}$ & 16.7 & 4.2 & 45.8 & 8.3 & 10.4 & -- & 14.6 & 39.6 & 0.0 & 20.8 & 17.8 \\
\midrule
\tabgroup{12}{(c)~Nominal success: \BaseRestart{} (\%) and paired change of every other arm (pp)}
$n$ & 47 & 49 & 36 & 37 & 26 & 50 & 45 & 49 & 49 & 49 & 437 \\
\BaseRestart{} (\%) & 42.6 & 63.3 & 25.0 & 24.3 & 57.7 & 28.0 & 42.2 & 77.6 & 24.5 & 51.0 & 43.9 \\
\cmidrule(lr){1-12}
\DemoCont & \ensuremath{-}4.3 & 0.0 & +2.8 & 0.0 & +11.5 & +2.0 & \ensuremath{-}2.2 & \ensuremath{-}2.0 & 0.0 & +18.4 & +2.3 \\
\TakeoverBC & \ensuremath{-}42.6 & \ensuremath{-}30.6 & \ensuremath{-}25.0 & \ensuremath{-}8.1 & \ensuremath{-}57.7 & \ensuremath{-}28.0 & \ensuremath{-}40.0 & \ensuremath{-}38.8 & \ensuremath{-}24.5 & \ensuremath{-}12.2 & \ensuremath{-}30.0 \\
Mix-BC$^{\dagger}$ & \ensuremath{-}34.0 & \ensuremath{-}28.6 & \ensuremath{-}8.3 & \ensuremath{-}8.1 & \ensuremath{-}30.8 & \ensuremath{-}10.0 & \ensuremath{-}33.3 & +6.1 & \ensuremath{-}14.3 & \ensuremath{-}8.2 & \ensuremath{-}16.5 \\
\IQLR & \ensuremath{-}42.6 & \ensuremath{-}26.5 & \ensuremath{-}25.0 & \ensuremath{-}5.4 & \ensuremath{-}38.5 & \ensuremath{-}28.0 & \ensuremath{-}35.6 & \ensuremath{-}24.5 & \ensuremath{-}14.3 & \ensuremath{-}6.1 & \ensuremath{-}24.3 \\
IQL-R0$^{\dagger}$ & \ensuremath{-}42.6 & \ensuremath{-}24.5 & \ensuremath{-}25.0 & \ensuremath{-}16.2 & \ensuremath{-}57.7 & \ensuremath{-}28.0 & \ensuremath{-}40.0 & \ensuremath{-}63.3 & \ensuremath{-}22.4 & \ensuremath{-}14.3 & \ensuremath{-}32.7 \\
AWAC$^{\dagger}$ & \ensuremath{-}42.6 & \ensuremath{-}28.6 & \ensuremath{-}22.2 & \ensuremath{-}5.4 & \ensuremath{-}50.0 & \ensuremath{-}28.0 & \ensuremath{-}40.0 & \ensuremath{-}46.9 & \ensuremath{-}24.5 & \ensuremath{-}10.2 & \ensuremath{-}29.5 \\
First-of-32$^{\dagger}$ & \ensuremath{-}42.6 & \ensuremath{-}28.6 & \ensuremath{-}25.0 & \ensuremath{-}8.1 & \ensuremath{-}57.7 & \ensuremath{-}28.0 & \ensuremath{-}40.0 & \ensuremath{-}34.7 & \ensuremath{-}24.5 & \ensuremath{-}12.2 & \ensuremath{-}29.3 \\
IDQL-R2$^{\dagger}$ & \ensuremath{-}42.6 & \ensuremath{-}59.2 & \ensuremath{-}25.0 & \ensuremath{-}24.3 & \ensuremath{-}57.7 & \ensuremath{-}28.0 & \ensuremath{-}42.2 & \ensuremath{-}77.6 & \ensuremath{-}24.5 & \ensuremath{-}30.6 & \ensuremath{-}41.2 \\
\bottomrule
\end{tabular}

%% file: supp/tab_e_levels.tex
\setlength{\tabcolsep}{0pt}
\tabzebra
\begin{tabular}{>{\hspace{6pt}}l<{\hspace{6pt}}*{14}{>{\hspace{6pt}}r<{\hspace{6pt}}}}
\toprule
\tabhead & \multicolumn{6}{c}{\SRrec{} (\%)} & & \multicolumn{6}{c}{Nominal success (\%)} & \dSRnorm{} (pp) \\
\cmidrule(lr){2-7}\cmidrule(lr){9-14}\cmidrule(l){15-15}
\tabhead Arm & L1 & L2 & L3 & L4 & L5 & All & Faults & L1 & L2 & L3 & L4 & L5 & All & @\timeGateScreen{}$T$ \\
\midrule
$n$ & 240 & 240 & 240 & 216 & 216 & 1,152 &  & 88 & 85 & 88 & 88 & 88 & 437 &  \\
\midrule
\BaseRestart & 26.2 & 21.2 & 19.2 & 9.3 & 6.0 & 16.8 & 65 & 77.3 & 51.8 & 44.3 & 27.3 & 19.3 & 43.9 & -- \\
\DemoCont & 30.0 & 20.0 & 17.5 & 6.5 & 5.6 & 16.3 & 72 & 77.3 & 56.5 & 48.9 & 25.0 & 23.9 & 46.2 & +0.5 \\
\TakeoverBC & 62.5 & 50.0 & 43.3 & 40.7 & 32.9 & 46.3 & 45 & 21.6 & 14.1 & 12.5 & 12.5 & 9.1 & 14.0 & \ensuremath{-}33.2 \\
Mix-BC$^{\dagger}$ & 58.8 & 42.5 & 34.6 & 32.9 & 19.4 & 38.1 & 59 & 42.0 & 37.6 & 29.5 & 17.0 & 11.4 & 27.5 & \ensuremath{-}30.9 \\
\IQLR & 65.0 & 47.9 & 47.5 & 38.0 & 29.2 & 46.0 & 51 & 27.3 & 22.4 & 21.6 & 11.4 & 15.9 & 19.7 & \ensuremath{-}31.4 \\
\midrule
IQL-R0$^{\dagger}$ & 59.2 & 49.6 & 37.9 & 36.6 & 28.7 & 42.8 & 47 & 14.8 & 10.6 & 12.5 & 4.5 & 13.6 & 11.2 & \ensuremath{-}33.2 \\
AWAC$^{\dagger}$ & 70.4 & 46.2 & 44.6 & 38.0 & 23.1 & 45.1 & 48 & 18.2 & 14.1 & 14.8 & 11.4 & 13.6 & 14.4 & \ensuremath{-}33.0 \\
First-of-32$^{\dagger}$ & 65.0 & 47.5 & 43.8 & 43.1 & 29.6 & 46.2 & 39 & 23.9 & 12.9 & 13.6 & 11.4 & 11.4 & 14.6 & \ensuremath{-}33.0 \\
IDQL-R2$^{\dagger}$ & 53.8 & 37.1 & 31.2 & 20.4 & 15.3 & 32.1 & 37 & 1.1 & 1.2 & 1.1 & 6.8 & 3.4 & 2.7 & \ensuremath{-}33.6 \\
\bottomrule
\end{tabular}

%% file: supp/tab_e_ftype.tex
\setlength{\tabcolsep}{0pt}
\begin{tabular}{>{\hspace{4pt}}l<{\hspace{4pt}}*{10}{>{\hspace{4pt}}r<{\hspace{4pt}}}}
\toprule
\tabhead Failure type & $n$ & \BaseRestart & \DemoCont & \TakeoverBC & Mix-BC$^{\dagger}$ & \IQLR & IQL-R0$^{\dagger}$ & AWAC$^{\dagger}$ & First-of-32$^{\dagger}$ & IDQL-R2$^{\dagger}$ \\
\midrule
\tabgroup{11}{All levels}
wrong placement & 351 & 22.5 & 23.4 & 43.6 & 36.8 & 49.0 & 42.5 & 53.3 & 43.9 & 37.9 \\
wrong object & 324 & 10.5 & 9.6 & 39.8 & 30.2 & 36.7 & 34.3 & 31.5 & 41.4 & 11.1 \\
collision & 180 & 10.0 & 10.6 & 62.8 & 56.7 & 64.4 & 58.9 & 52.2 & 63.9 & 58.3 \\
missed grasp & 111 & 6.3 & 4.5 & 36.0 & 17.1 & 30.6 & 29.7 & 31.5 & 30.6 & 14.4 \\
stuck & 78 & 32.1 & 33.3 & 47.4 & 35.9 & 37.2 & 33.3 & 48.7 & 44.9 & 30.8 \\
dropped & 75 & 22.7 & 18.7 & 68.0 & 72.0 & 65.3 & 66.7 & 68.0 & 68.0 & 65.3 \\
drawer not closed & 33 & 39.4 & 33.3 & 30.3 & 27.3 & 33.3 & 54.5 & 36.4 & 27.3 & 21.2 \\
\midrule
\tabgroup{11}{L1--L3}
wrong placement & 294 & 24.1 & 24.8 & 45.6 & 38.1 & 52.0 & 42.9 & 57.1 & 45.6 & 40.1 \\
wrong object & 120 & 13.3 & 15.8 & 51.7 & 44.2 & 53.3 & 45.0 & 45.0 & 52.5 & 18.3 \\
collision & 132 & 13.6 & 13.6 & 68.2 & 59.1 & 67.4 & 64.4 & 59.1 & 69.7 & 62.9 \\
missed grasp & 15 & 20.0 & 26.7 & 40.0 & 46.7 & 40.0 & 33.3 & 20.0 & 33.3 & 26.7 \\
stuck & 66 & 36.4 & 36.4 & 51.5 & 40.9 & 37.9 & 39.4 & 54.5 & 50.0 & 36.4 \\
dropped & 60 & 25.0 & 21.7 & 63.3 & 66.7 & 61.7 & 63.3 & 60.0 & 65.0 & 58.3 \\
drawer not closed & 33 & 39.4 & 33.3 & 30.3 & 27.3 & 33.3 & 54.5 & 36.4 & 27.3 & 21.2 \\
\midrule
\tabgroup{11}{L4--L5}
wrong placement & 57 & 14.0 & 15.8 & 33.3 & 29.8 & 33.3 & 40.4 & 33.3 & 35.1 & 26.3 \\
wrong object & 204 & 8.8 & 5.9 & 32.8 & 22.1 & 27.0 & 27.9 & 23.5 & 34.8 & 6.9 \\
collision & 48 & 0.0 & 2.1 & 47.9 & 50.0 & 56.2 & 43.8 & 33.3 & 47.9 & 45.8 \\
missed grasp & 96 & 4.2 & 1.0 & 35.4 & 12.5 & 29.2 & 29.2 & 33.3 & 30.2 & 12.5 \\
stuck & 12 & 8.3 & 16.7 & 25.0 & 8.3 & 33.3 & 0.0 & 16.7 & 16.7 & 0.0 \\
dropped & 15 & 13.3 & 6.7 & 86.7 & 93.3 & 80.0 & 80.0 & 100.0 & 80.0 & 93.3 \\
\bottomrule
\end{tabular}

%% file: supp/tab_f_nominal_decomp.tex
\tabzebra
\begin{tabular}{l rrrr r rrr rr}
\toprule
\tabhead & \multicolumn{4}{c}{Outcome over 437 valid starts (\%)} & & \multicolumn{3}{c}{Cumulative SR (\%)} & \multicolumn{2}{c}{vs.\ \DemoCont{} (H200)} \\
\cmidrule(lr){2-5}\cmidrule(lr){7-9}\cmidrule(lr){10-11}
\tabhead Arm & ${\le}1.1T$ & $(1.1T,1200]$ & Timeout & Fault & \makecell{SR @386\\(\%)} & @400 & @800 & @1,200 & \makecell{$\Delta$@1,200 (pp)\\{}[95\% CI]} & \makecell{$\Delta$@1.1$T$\\(pp)} \\
\midrule
\BaseRestart & 33.6 & 10.3 & 42.8 & 13.3 & 49.7 & 23.8 & 40.5 & 43.9 & \ensuremath{-}2.3 [\ensuremath{-}6.5, +1.8] & \ensuremath{-}0.5 \\
\DemoCont & 34.1 & 12.1 & 40.5 & 13.3 & 52.3 & 26.8 & 42.6 & 46.2 & ref. & ref. \\
\TakeoverBC & 0.5 & 13.5 & 73.5 & 12.6 & 15.8 & 3.7 & 11.2 & 14.0 & \ensuremath{-}32.3 [\ensuremath{-}39.8, \ensuremath{-}25.6] & \ensuremath{-}33.6 \\
Mix-BC$^{\dagger}$ & 2.7 & 24.7 & 59.7 & 12.8 & 31.1 & 6.4 & 20.6 & 27.5 & \ensuremath{-}18.8 [\ensuremath{-}25.3, \ensuremath{-}12.3] & \ensuremath{-}31.4 \\
\IQLR & 2.3 & 17.4 & 66.4 & 14.0 & 22.3 & 5.0 & 16.5 & 19.7 & \ensuremath{-}26.5 [\ensuremath{-}34.5, \ensuremath{-}19.0] & \ensuremath{-}31.8 \\
IQL-R0$^{\dagger}$ & 0.5 & 10.8 & 75.5 & 13.3 & 12.7 & 3.7 & 10.1 & 11.2 & \ensuremath{-}35.0 [\ensuremath{-}43.8, \ensuremath{-}26.7] & \ensuremath{-}33.6 \\
AWAC$^{\dagger}$ & 0.7 & 13.7 & 71.9 & 13.7 & 16.3 & 3.0 & 11.0 & 14.4 & \ensuremath{-}31.8 [\ensuremath{-}39.9, \ensuremath{-}23.7] & \ensuremath{-}33.4 \\
First-of-32$^{\dagger}$ & 0.7 & 14.0 & 73.5 & 11.9 & 16.6 & 3.7 & 11.9 & 14.6 & \ensuremath{-}31.6 [\ensuremath{-}39.2, \ensuremath{-}24.9] & \ensuremath{-}33.4 \\
IDQL-R2$^{\dagger}$ & 0.0 & 2.7 & 82.8 & 14.4 & 3.1 & 1.6 & 2.5 & 2.7 & \ensuremath{-}43.5 [\ensuremath{-}53.2, \ensuremath{-}33.1] & \ensuremath{-}34.1 \\
\bottomrule
\end{tabular}

%% file: supp/app_f.tex
\section{Additional Analyses}
\label{sec:s-analyses}

This section examines the \IQLR critic, which weights the actor data of \IQLR and selects the executed chunk of \mbox{IDQL-R2}: its pre-specified weight diagnostic (\cref{sec:s-critic}), where critic reranking gains and loses against \TakeoverBC (\cref{sec:s-rerank}), what the reranking selects in a post-hoc offline replay (\cref{sec:s-replay}), and how policy command scales compare with demonstrations and takeovers (\cref{sec:s-command-scale}). None of these analyses enters the Q1 or Q2 decision.

\subsection{Critic diagnostics}
\label{sec:s-critic}

\begin{table*}[t]
\centering\footnotesize\setlength{\tabcolsep}{5pt}\tabstyle
\caption{\textbf{Actor chunk counts, sampling weights and effective draw shares for \IQLR{} and AWAC.} Chunks: complete 10-action chunks in the eligible set $\mathcal E$ (\nEligChunksAll{} total: \pctEligUnwRec{}\% in $\mathcal R$, \pctEligUnwFwin{}\% in $\mathcal F_{\mathrm{win}}$). Mean, Median and At cap: summary of fixed actor weights $w_i$ under Eq.~(1) of the main paper (cap 100; --: not logged separately). Draw share: total actor sampling probability $\sum_i p_i$ (\%) over each pool when sampling with replacement. \IQLR{} fails the pre-specified check $\text{Mean}(\mathcal R)>\text{Mean}(\mathcal F_{\mathrm{win}})$, whereas exploratory AWAC$^{\dagger}$ meets it; under both critics over 94\% of actor draws come from $\mathcal R$.}
\label{tab:s-critic}
\input{supp/tab_g_critic}
\end{table*}

\begin{table*}[t]
\centering\footnotesize\setlength{\tabcolsep}{4.5pt}\tabstyle
\caption{\textbf{Command-scale profile across demonstrations, accepted takeovers and \TakeoverBC{} held-out samples (descriptive).} Steps: total recorded control steps ($\mathcal D$ and $\mathcal R$) or predicted chunk steps across the \rpQueries{} held-out first queries ($32\times10$ steps per query from the replay of \cref{sec:s-replay}). $|a_{\mathrm{trans}}|$ $p_{99}$ and $|a_{\mathrm{rot}}|$ $p_{99}$: range of the per-axis 99th percentiles across the three translation or rotation axes. $|a_{\mathrm{trans}}|{>}0.25$ and $|a_{\mathrm{rot}}|{>}0.08$: per-axis range of step shares above the SpaceMouse translation and rotation caps (\cref{sec:s-takeover}). $\lVert p_t\rVert_2$: 50th, 90th and 99th percentiles of the 3-D per-step translation norm.}
\label{tab:s-command-scale}
\input{supp/tab_g_scales}
\end{table*}

The analysis plan pre-specified a diagnostic for the \IQLR weighting: recovery chunks ($\mathcal R$), which lead to task success, should receive a higher mean actor sampling weight than failure-window chunks ($\mathcal F_{\mathrm{win}}$, \cref{sec:s-value-arms}), which lead into a failure. The \IQLR critic does not meet it (\cref{tab:s-critic}): the mean weight is \secFiveRecoveryWeight on recovery chunks and \secFiveFailureWeight on failure-window chunks. The medians are ordered as intended; the inversion of the means comes from a heavy right tail of the failure-window weights, with a larger share at the weight cap, concentrated in families 4 and 14. Because $\mathcal R$ outnumbers $\mathcal F_{\mathrm{win}}$ by \nRecChunksPost{} to \nFwinChunksPost{} complete chunks (\pctEligUnwRec{}\% \vs \pctEligUnwFwin{}\% of $\mathcal E$), \pctDrawRecIQL{}\% of expected actor draws in \IQLR{} still come from $\mathcal R$ and only \pctDrawFwinIQL{}\% from $\mathcal F_{\mathrm{win}}$ (normalized effective sample size \essRecIQLPct{}\% of $|\mathcal R|$). As pre-specified, a failed diagnostic does not remove an arm from the analysis, so \IQLR remains the Q2 arm; the diagnostic characterizes the learned weighting without identifying the cause of the Q2 outcome. The exploratory AWAC critic meets the diagnostic (mean weights \wMeanRecAWAC{} \vs \wMeanFwinAWAC{}, \pctDrawRecAWAC{}\% of draws from $\mathcal R$, ESS \essRecAWACPct{}\% of $|\mathcal R|$); \mbox{IQL-R0} samples recovery chunks only (mean weight \wMeanRecIQLzero{}, \clipRecIQLzeroPct{}\% at cap, ESS \essRecIQLzeroPct{}\%).

\subsection{Critic reranking with IDQL-R2}
\label{sec:s-rerank}

\mbox{IDQL-R2}~\cite{hansenestruch2023idql} draws 32 chunks from \TakeoverBC at every policy query and executes the one with the highest $\min(Q_1,Q_2)$ under the \IQLR critic; \mbox{First-of-32} executes the first of the same 32 samples, so the two arms differ only in the selection rule (\cref{sec:s-inference-variants}). On L1--L3, \SRrec is \recLoIDQL{}\% for \mbox{IDQL-R2}, \recLoTB{}\% for \TakeoverBC and \recLoFirst{}\% for \mbox{First-of-32} (\cref{sec:s-results}). The comparisons below are descriptive, use all \nRecRowsArm rollouts per arm over all \nLevels levels, and pair rollouts across arms by restored point and sampling seed.

\mbox{IDQL-R2} finishes shared successes faster, and most of its losses end at the step limit. On the \idqlBothN rollouts that both \mbox{IDQL-R2} and \TakeoverBC complete within \horizonMain steps, \mbox{IDQL-R2} is faster in \idqlFasterPct{}\% of them (median success step \idqlBothMedStep \vs \tbBothMedStep). Of the \idqlLostN rollouts that \TakeoverBC completes and \mbox{IDQL-R2} does not, \idqlLostTimeoutN reach the \horizonLong-step limit without success, \idqlLostFaultN end in a fault and \idqlLostLateN succeed only after \horizonMain steps. These losses grow with the number of steps that \TakeoverBC needs, whereas \mbox{First-of-32}, which executes the same samples without the critic, keeps most of the \TakeoverBC successes: on the \idqlTBShortN rollouts that \TakeoverBC completes in fewer than 100 steps, \mbox{IDQL-R2} succeeds on \idqlTBShortRate{}\% and \mbox{First-of-32} on \firstTBShortRate{}\%; on the \idqlTBLongN rollouts that take \TakeoverBC 400 steps or more, the two rates are \idqlTBLongRate{}\% and \firstTBLongRate{}\%. By failure type, the largest drops relative to \TakeoverBC occur on wrong-object and missed-grasp failures (\cref{tab:s-ftype}).

\subsection{Offline replay of the reranking (post hoc)}
\label{sec:s-replay}

\paragraph{Setup.} To characterize what the argmax selects, we replayed, after all evaluations, the first policy query of every held-out rollout (\rpQueries queries: \nFailTest points with \samplesPerPoint sampling seeds each) with the evaluated \TakeoverBC checkpoint and \IQLR critic, recording both $Q$ heads, $V$ and all 32 candidates. Because the replay ran on a different GPU type than the evaluation of \mbox{IDQL-R2}, bf16 sampling differs slightly, so individual choices are not reproduced; we report aggregates only. For a 10-step chunk, translation is $\sum_{t=1}^{10}\lVert p_t\rVert_2$ over the position commands $p_t$, rotation is the corresponding sum over the rotation commands, and the gripper value is the mean gripper command. For each property, the rank of the chosen chunk among the 32 candidates (ties share their mean rank) is scaled to 0--100, where 50 means no preference, and averaged over queries.

\paragraph{The argmax favors larger translations.} The chosen chunk has a mean translation rank of \rpSelTrans (95\% bootstrap interval over points [\rpSelTransLo, \rpSelTransHi{}]), compared with \rpSelRot for rotation and \rpSelGrip for the gripper value. Across the candidates of a query, $Q$ and translation have a mean Spearman correlation of \rpRhoTrans, and the chosen chunk's translation is on average \rpTransGain{}\% larger with 32 candidates than with a single sample. The chosen chunk also tends to lie apart from the other candidates: in the normalized 70-dimensional critic action space, its distance to the nearest other candidate has a mean rank of \rpSelNN.

\paragraph{The preference also holds on training takeovers.} Scoring eight pre-drawn \TakeoverBC samples at each of the \rpTrainFrames critic training frames gives a mean Spearman correlation of \rpRhoTransTrain between $Q$ and translation on takeover frames; on failure-prefix frames, the correlation is \rpRhoTransPrefix with translation and \rpRhoRotPrefix with rotation. Scaling the six arm commands of a takeover-frame candidate by factors of 0.5, 0.75, 1, 1.25 and 1.5 raises $Q$ monotonically (by more than $10^{-4}$ per step) for \rpMonoUp{}\% of candidates and lowers it monotonically for \rpMonoDown{}\%.

\paragraph{The two heads often disagree near the top of the ranking.} In \rpGapBelowTwin{}\% of queries, the margin between the best and second-best candidate under $\min(Q_1,Q_2)$ is smaller than the difference between the two heads' margins for the same pair, and the heads pick the same winner in only \rpHeadAgree{}\%. The $Q$ values also spread more at held-out states: with eight candidates per state, the median $Q$ range is \rpQrangeRatio times that on training takeover frames. Because both heads are fitted to the same target, their disagreement is a scale reference, not a calibrated estimate of critic error.

Together, the replay shows an argmax that favors larger translations and often decides by margins smaller than the disagreement between the heads. Whether these choices speed up recoveries already on course and stall long ones (\cref{sec:s-rerank}) remains open, since only the first query of each rollout was replayed.

\subsection{Command-scale profile of demonstrations, takeovers and policy samples}
\label{sec:s-command-scale}

\Cref{tab:s-command-scale} compares the per-step command magnitudes of the VR-collected \liberox{} demonstrations ($\mathcal D$), the SpaceMouse accepted recoveries ($\mathcal R$; \cref{sec:s-takeover}), and all \tbPredStepsAll{} action steps predicted by \TakeoverBC{} across the \rpQueries{} first queries of the held-out replay (\cref{sec:s-replay}). \TakeoverBC{} closely reproduces the SpaceMouse command envelope: its per-axis 99th percentiles are \tbPredPNNTrans{} in translation and \tbPredPNNRot{} in rotation (against \demoPNNTrans{} and \demoPNNRot{} in $\mathcal D$), only \tbPredAboveTrans{}\% and \tbPredAboveRot{}\% of predicted steps exceed 0.25 and 0.08 on any axis, and its median 3-D step translation norm is \tbPredTransNormMed{} (\vs \tkTransNormMed{} in $\mathcal R$ and \demoTransNormMed{} in $\mathcal D$). Although per-step actions were not logged during the full nominal evaluation, this inherited command scale is consistent with the slower completion pace of takeover-trained policies (\nomMedRatioTTB{}$T$ median nominal completion step for \TakeoverBC{} \vs \nomMedRatioTBase{}$T$ for \BaseRestart; \cref{sec:s-res-nominal}) without establishing a causal decomposition of the \horizonMain-step timeouts.

%% file: supp/tab_g_critic.tex
\tabzebra
\begin{tabular}{ll r cccc}
\toprule
\tabhead Arm & Actor pool & Chunks & Mean & Median & At cap (\%) & Draw share (\%) \\
\midrule
\IQLR & Recovery, $\mathcal R$ & 776,507 & \secFiveRecoveryWeight & 0.99 & 0.36 & 94.7 \\
\IQLR & Failure window, $\mathcal F_{\mathrm{win}}$ & 11,926 & \secFiveFailureWeight & 0.52 & 2.2 & 5.3 \\
AWAC$^{\dagger}$ & Recovery, $\mathcal R$ & 776,507 & 1.54 & 1.01 & 0.29 & 98.6 \\
AWAC$^{\dagger}$ & Failure window, $\mathcal F_{\mathrm{win}}$ & 11,926 & 1.41 & 0.69 & -- & 1.4 \\
\bottomrule
\end{tabular}

%% file: supp/tab_g_scales.tex
\tabzebra
\begin{tabular}{l r cccc c}
\toprule
\tabhead Source & Steps & $|a_{\mathrm{trans}}|$ $p_{99}$ & $|a_{\mathrm{rot}}|$ $p_{99}$ & $|a_{\mathrm{trans}}|{>}0.25$ (\%) & $|a_{\mathrm{rot}}|{>}0.08$ (\%) & $\lVert p_t\rVert_2$ ($p_{50}/p_{90}/p_{99}$) \\
\midrule
VR demonstrations ($\mathcal D$) & \nDemoFrames & \demoPNNTrans & \demoPNNRot & \demoAboveCapTrans & \demoAboveCapRot & 0.28 / 0.63 / 1.10 \\
Accepted takeovers ($\mathcal R$) & \nStepsAccepted & 0.25 & 0.07--0.08 & 0.0 & 0.0 & 0.21 / 0.29 / 0.35 \\
\TakeoverBC{} held-out samples & 368,640 & 0.22--0.27 & 0.06--0.07 & 0.2--3.7 & 0.2--0.5 & 0.17 / 0.26 / 0.30 \\
\bottomrule
\end{tabular}

%% file: supp/app_h.tex
\section{Dataset Documentation, Limitations and Ethics}
\label{sec:s-documentation}

This section documents the dataset in the form of a datasheet (\cref{sec:s-datasheet}), states the boundaries of the study design (\cref{sec:s-governance}), describes the human participant (\cref{sec:s-ethics}) and describes the offline data browser (\cref{sec:s-website}).

\subsection{Datasheet}
\label{sec:s-datasheet}

The answers follow the questions of Gebru \etal~\cite{gebru2021datasheets} on motivation, composition, collection, preprocessing, uses, distribution and maintenance, and describe the dataset as frozen at the end of collection; human participants are covered in \cref{sec:s-ethics}.

\paragraph{Motivation.}
\begin{description}[beginpenalty=10000]
\item[Purpose.] To measure, under controlled conditions, what human takeovers at a policy's own failure states add to VLA post-training beyond continued training on demonstrations, for one \pizf checkpoint on \nFamilies \liberox task families. The data are policy-matched by construction: failures, takeover starts and held-out evaluation states all come from rollouts of the same checkpoint.
\item[Creators.] The authors, at Nanyang Technological University, Singapore.
\end{description}

\paragraph{Composition.}
\begin{description}[beginpenalty=10000]
\item[Instances.] (i)~\nFailReleased natural failures of the base checkpoint released into the dataset by the coordinator (\nFailTrain training, \nFailTest held-out), each with an annotated interval and a failure-type label; (ii)~\nAttemptsAll SpaceMouse takeover attempts from restorable simulator states at seeded frames inside training intervals (\nAttemptsActive on active and \nAttemptsArchived on archived sources), among them \nAccepted accepted recoveries, \framesPerFailure for each of \nSourcesComplete complete training failures, with \nStepsAccepted control steps (\hoursAcceptedControl~h at \controlHz~Hz); (iii)~the \nFailTest held-out failure points with their restorable states; (iv)~the screening outcome, annotations and curation decision of each of the \nShipped candidates sent to annotation. Counts are given by level in Table~2 of the main paper and by configuration in \cref{tab:s-configs}.
\item[Sampling.] The released failures are a screened, quota-based subset rather than a random sample of the policy's failures. Candidates come from pre-assigned seed blocks, had not succeeded by \timeGateCollect{}$T$, and passed the near-miss and physics checks and, except in families 0 and 9 (not screened before annotation), the start-validity check; they were then annotated and released under per-configuration quotas. \Cref{sec:s-funnel} gives, per configuration, the rollouts generated, those removed by each check and the resulting candidates.
\item[Content.] Rollouts: two RGB views rendered from the recorded simulator state at every control step, a fixed third-person view and the wrist (eye-in-hand) camera, each $256\times256$ pixels (the policy receives them padded and resized to $224\times224$), proprioceptive state, actions and restorable simulator states. Attempts: source failure, start frame, sampling seed, observations and actions, end reason, steps, control time, discard and review flags, and a pseudonymous operator ID.
\item[Labels.] Annotations (interval, failure type, no-single-event-frame flag and judgment; \cref{sec:s-ftypes}) come from one annotator; the coordinator decides which candidates enter the dataset and finalizes their intervals and labels. Attempt outcomes combine the simulator's task predicate, the operator's end action and the coordinator's acceptance decision, which rests on automatic success and integrity checks (\cref{sec:s-takeover}).
\item[Splits.] Training and held-out failures are seed-disjoint instances of the same configurations; there are no held-out tasks. The \nDevRoots development failures, complete training failures set aside together with their takeovers, serve for implementation checks (\cref{sec:s-recipe}); post-training uses the other \nTrainSrcPost complete training failures (\nTrainAccPost accepted recoveries).
\item[Missing data.] Family~9 L4--L5 (\cref{tab:s-configs}): no accepted L4 recovery, no released L5 training failure and no released L4--L5 held-out failure. Of the candidates sent to annotation, \nUndecided received no curation decision (\cref{sec:s-funnel}); \suppPending attempts have no acceptance decision and are not used for training or evaluation (\cref{tab:s-flags}).
\item[Errors and noise.] Main recorded issues: one annotator without an agreement estimate; \suppWidthFlag width-flagged held-out intervals; \suppOutOfInterval attempts at a frame outside the revised interval of their source; \nPhysics physics-fault attempts on active sources; and late successes retained as failures (\cref{sec:s-configs,sec:s-flags}).
\item[Dependencies.] The \liberox~\cite{wang2026liberox} scenes, task definitions and demonstrations; the MuJoCo and robosuite simulators, which restore the recorded states; and the base checkpoint, fine-tuned with openpi~\cite{openpi} from the \pizf weights (\cref{sec:s-platform-base}).
\item[Sensitive content.] None. The scenes are simulated household layouts and no person appears in the data; the operator data are described in \cref{sec:s-ethics}.
\end{description}

\paragraph{Collection process.}
\begin{description}[beginpenalty=10000]
\item[Acquisition.] Base-policy rollouts were generated in simulation on pre-assigned seeds and screened automatically. One person annotated the candidates and performed the takeovers with a custom collection tool (\cref{sec:s-ethics}); every takeover starts from a restored simulator state.
\item[Takeover sampling.] \framesPerFailure frames per training failure, drawn uniformly without replacement inside the interval with a recorded seed; same-frame redo until success (\cref{sec:s-takeover}).
\item[Timeframe.] Annotation and takeover collection ran from October 20 to November 6, 2026.
\end{description}

\paragraph{Preprocessing, cleaning and labeling.}
Near-miss and physics-fault rollouts were excluded before annotation, as were invalid starts except in families 0 and 9, where the start-validity check was not applied before annotation (\cref{sec:s-funnel}). After annotation, the coordinator released or withheld candidates under the per-configuration quotas, \nSourcesArchived sources were archived, and sampled frames of family-0 L4 and L5 sources were redrawn after physics faults (\cref{tab:s-flags}). All records are kept: archived sources, discarded successes and attempts ending in an operator stop, a time limit or a physics fault are counted separately (\cref{tab:s-configs}).

\paragraph{Uses.}
\begin{description}[beginpenalty=10000]
\item[Intended uses.] (i)~Post-training from takeovers with evaluation at restored held-out failure states; (ii)~same-state comparison of the base policy's continuation with the human recovery at every takeover start; (iii)~learning from failed, abandoned and discarded corrections; (iv)~failure localization from the interval labels; (v)~budget subsampling within the balanced design of \nTrainPerConfig failures $\times$ \framesPerFailure accepted recoveries per configuration (\nConfigsComplete of the \nConfigs configurations complete); (vi)~evaluating any new policy at the same restorable failure states.
\item[Not supported.] Claims about human capability (frame-level success is 100\% by construction; attempt-level rates reflect one operator under the redo rule); claims about the effect of policy matching (one policy), operator variability, when to take over, or sim-to-real transfer.
\item[Held-out use.] The held-out failure states serve evaluation only; methods and hyperparameters should be selected on training data, such as the \nDevRoots development failures.
\end{description}

\paragraph{Distribution.}
\begin{description}[beginpenalty=10000]
\item[Planned public release.] (i)~Data: all \nFailReleased{} released failures and all \nAttemptsAll{} takeover attempts ($\mathcal R$, $\mathcal F$ and $\mathcal N$, plus archived-source and discarded records), distributed both as raw restorable MuJoCo states, fixture layouts, action histories and two-camera RGB streams and as LeRobot~v2.1 datasets~\cite{cadene2026lerobot}; all \nAnnotationRecords{} annotations, curation flags, split manifests and machine-readable task mappings (\cref{tab:s-bddl}) with SHA-256 digests; and the frozen per-rollout evaluation logs (\texttt{rec.jsonl}, \texttt{norm.jsonl}). (ii)~Code: the collection tool, the restore-and-continue and nominal evaluators, the locked \texttt{openpi} post-training patch, the IQL and AWAC critic and action-cache scripts, and the frozen analysis script (\texttt{fpx\_analyze.py}). (iii)~Weights: the fine-tuned \pizf{} base checkpoint and the six post-trained actor and three critic checkpoints (\cref{sec:s-compute}).
\item[Public release and licenses.] The code is released at \fpxcodelink{} under the MIT License. The data, restorable states, annotations and evaluation logs will be released on the Hugging Face Hub under CC~BY~4.0, retaining attribution to \liberox~\cite{wang2026liberox} (demonstrations CC~BY~4.0~\cite{meituan2026liberoxdata}, code MIT~\cite{meituan2026liberoxcode}); and model checkpoints under CC~BY~4.0 subject to the upstream \pizf{} and \texttt{openpi} model terms~\cite{black2025pi05,openpi}.
\item[Project website.] The website \fpxprojectlink{} hosts the data browser (\cref{sec:s-website}) and the project video; the code repository describes how to score the recorded \BaseRestart{} run.
\end{description}

\paragraph{Maintenance.}
\begin{description}[beginpenalty=10000]
\item[Versioning and support.] Snapshot \texttt{v1.0} freezes the dataset as collected through November~6, 2026. Any errata or prospective quality-audit records will be published as new versioned snapshots with a changelog while keeping \texttt{v1.0} available so that all reported results remain reproducible. The authors will maintain the hosted repositories, issue tracker and download mirrors for at least three years following publication; questions and errata reports are handled through the issue tracker of the code repository.
\end{description}

\input{supp/scope_release_ethics}

\subsection{Project website and data browser}
\label{sec:s-website}

\begin{figure*}[p]
\centering
\includegraphics[width=\textwidth]{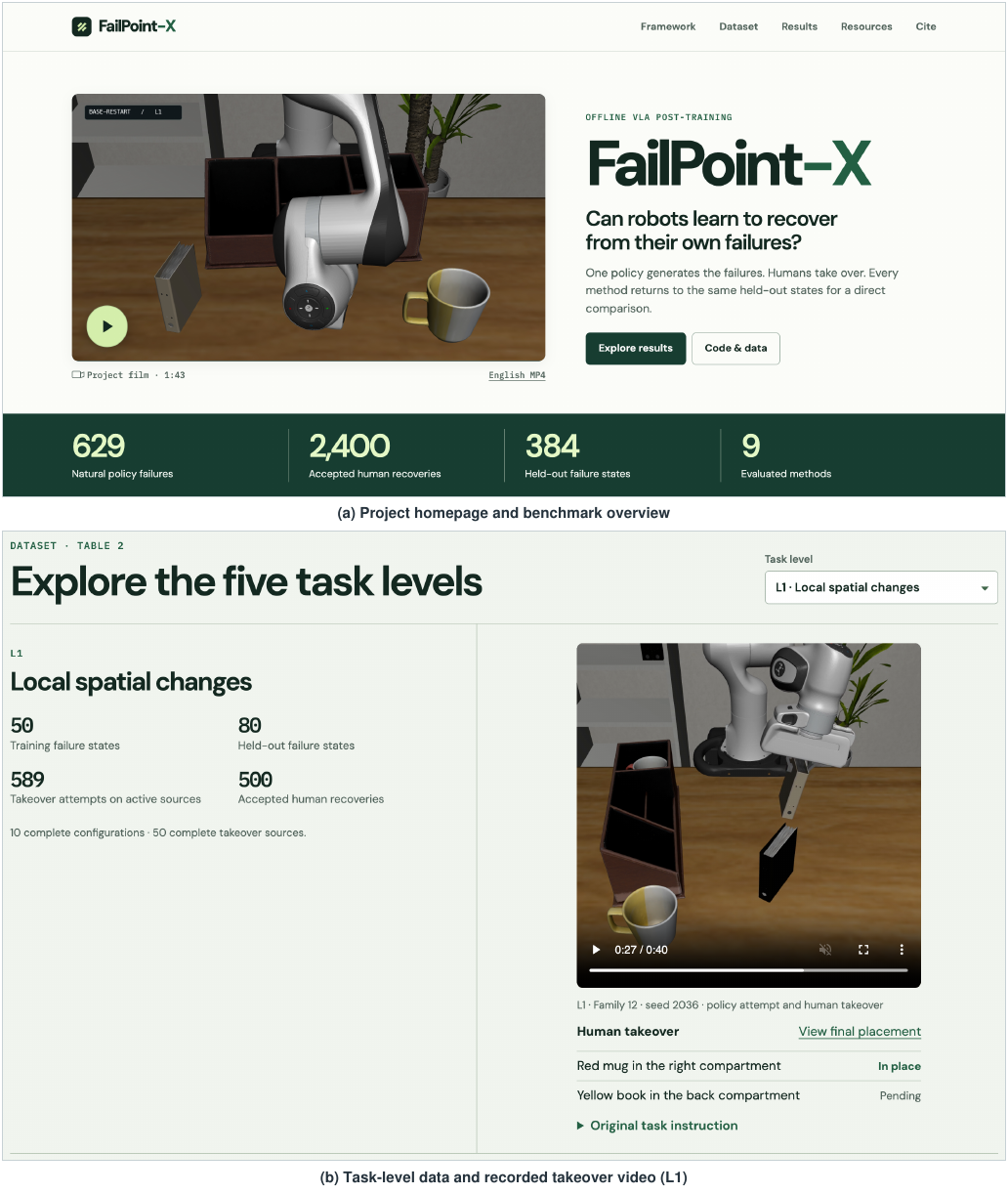}
\caption{\textbf{Project website.} Two views of the homepage. (a)~The opening view includes the project video and dataset counts. (b)~Selecting a task level displays its failure-state and takeover counts alongside a recorded example. The L1 example (family~12, seed~2036) is paused during human takeover: the red mug is in place, and the yellow book has yet to reach its target compartment.}
\label{fig:s-website}
\end{figure*}

The project website (\fpxprojectlink) introduces the benchmark and pairs task-level counts with recorded takeover videos (\cref{fig:s-website}). It also provides interactive results and method rankings. Links on the same page open the recorded failure examples (\cref{fig:s-gallery}), the family~12 task-level scenes, and the policy replays rendered from saved simulator states (\cref{fig:s-cases}).

\clearpage

%% file: supp/scope_release_ethics.tex
\subsection{Design boundaries and limitations}
\label{sec:s-governance}

The study design sets the following boundaries.
\begin{description}
\item[One checkpoint in simulation.] With one \pizf checkpoint in simulation, neither the effect of policy matching itself nor transfer to other architectures or real robots is tested. Recipes initialized from this $\pi_{\mathrm{base}}$ preserve policy matching by construction; evaluating policies trained from another base checkpoint reuses the restorable failure states without policy matching.
\item[One operator and annotator.] One person performed all takeovers and annotations (no blind second annotation, so no agreement estimate) and collected families sequentially, confounding operator practice with family; sensitivity to recorded interval and curation flags is reported in \cref{tab:s-curation-sens}.
\item[Takeover device and pace.] Takeovers use a SpaceMouse at \controlHz{}\,Hz ($\pm0.25$ translation, $\pm0.08$ rotation), whereas the base policy's demonstrations were VR-teleoperated~\cite{wang2026liberox} and use larger commands (\cref{sec:s-takeover,sec:s-command-scale}); post-trained policies inherit this slower command scale, coupling nominal completion at fixed step horizons with execution pace (\cref{sec:s-res-nominal}).
\item[Evaluation hardware.] \BaseRestart{} was evaluated on an RTX~5090 GPU and the other eight arms on H200 GPUs, where bf16 sampling trajectories can differ (\cref{sec:s-cases,sec:s-replay}); same-hardware H200 nominal contrasts against \DemoCont{} yield consistent retention losses (\cref{tab:s-nominal-decomp}).
\item[Curation.] Same-frame redo and \nDiscardedSucc discarded successes curate the accepted data; the dataset therefore keeps every attempt, and we report attempt-level human success (\cref{sec:s-configs}).
\item[Power and coverage.] Confirmatory tests use $n$\,=\,\nFamilies families, which limits power (\cref{sec:s-evaluation}); L4--L5 are exploratory, without family-9 held-out states or accepted recoveries (\cref{tab:s-configs}).
\item[One training seed.] Each of the six actor-trained arms uses one seed (seed~42); the intervals condition on the evaluated checkpoints and resample task families for recovery and configurations for nominal performance, not training seeds.
\item[Restore and faults.] Every restored state matched the recorded simulator state (\cref{sec:s-restore}); physics-guard faults (\nFaultIDQL--\nFaultDemo of \nRecRowsArm rollouts per arm) count as failures, and excluding them shifts no pooled paired difference by more than \faultSensMaxShift~pp (\cref{sec:s-res-robustness}).
\item[Out of scope.] Takeover timing (restarts at frames other than $m$), gains over new demonstrations of equal human time, online or iterative learning and new tasks are not studied.
\end{description}

\subsection{Human participants and ethics}
\label{sec:s-ethics}

The only human participant was one person who served as both takeover operator and annotator under a pseudonymous ID, producing all \nAttemptsAll attempts and all \nAnnotationRecords annotation records, with \hoursControl~h of active control and \hoursAnnotation~h of annotation. Curation decisions (which annotated candidates enter the dataset, frame replacements and takeover acceptance) were made by a coordinator. The task, teleoperating a simulated arm with a 3D mouse, involves no physical robot. The operator and the coordinator are authors; they received no separate compensation and knew the study hypotheses when making curation, discard and frame-replacement decisions. No ethics review was sought. The collection tool records no video or audio of the operator: the only operator data are device actions and timing logs under the pseudonymous ID, and the dataset contains no other personal information.